\documentclass[runningheads]{llncs}

\usepackage{eccv}

\usepackage{eccvabbrv}

\usepackage{graphicx}
\usepackage{booktabs}
\usepackage{algpseudocode}
\usepackage{algorithm}
\usepackage{multirow}
\usepackage[accsupp]{axessibility}  

\usepackage[pagebackref,breaklinks,colorlinks,citecolor=eccvblue]{hyperref}
\usepackage[pagebackref,breaklinks,colorlinks,citecolor=eccvblue]{hyperref}
\usepackage{orcidlink}

\def\eg{\emph{e.g.}} 
\def\ie{\emph{i.e.}}

\DeclareMathOperator*{\argmax}{arg\,max}

\newcommand{\tenc}{g} 
\newcommand{\ienc}{f} 
\newcommand{\D}{d}
\newcommand{\ctok}{*}

\def\pseudowords{custom tokens}

\begin{document}


\titlerunning{Abbreviated paper title}

\title{Descriminative-Generative Custom Tokens\\ for Vision-Language Models}

\author{Pramuditha Perera\\
AWS AI Labs\\
{\tt\small pramudi@amazon.com}
\and
Matthew Trager\\
AWS AI Labs\\
{\tt\small mttrager@amazon.com}
\and
Luca Zancato\\
AWS AI Labs\\
{\tt\small zancato@amazon.it}
\and
Alessandro Achille\\
AWS AI Labs\\
{\tt\small aachille@amazon.com}
\and
Stefano Soatto\\
AWS AI Labs\\
{\tt\small soattos@amazon.com}
}

\author{Pramuditha Perera \and
Matthew Trager \and
Luca Zancato \and
Alessandro Achille \and
Stefano Soatto 
}

\authorrunning{P.~Perera et al.}

\institute{AWS AI Labs}



\maketitle

\begin{abstract}
  This paper explores the possibility of learning \emph{\pseudowords} for representing new concepts in Vision-Language Models (VLMs). Our aim  is to learn tokens that can be effective for both discriminative and generative tasks while composing well with words to form new input queries. The targeted concept is specified in terms of a small set of images and a parent concept described using text. We operate on CLIP text features and propose to use a combination of a textual inversion loss and  a classification loss to ensure that text features of the learned token are aligned with image features of the concept in the CLIP embedding space. We restrict the learned token to a low-dimensional subspace spanned by tokens for attributes that are appropriate for the given super-class. These modifications improve the quality of compositions of the learned token with natural language for generating new scenes. Further, we show that learned \pseudowords{} can be used to form queries for text-to-image retrieval task, and also have the important benefit that composite queries can be \emph{visualized} to ensure that the desired concept is faithfully encoded. Based on this, we introduce the method of Generation Aided Image Retrieval, where the query is modified at inference time to better suit the search intent. On the DeepFashion2 dataset, our method improves Mean Reciprocal Retrieval (MRR) over relevant baselines by 7\%.
\end{abstract}

\section{Introduction}

Humans use language to represent and communicate concepts. They have the ability to easily associate new meanings to words or phrases, for example to describe refinements of known concepts---\emph{the neighbor's dog} is a particular \emph{dog} that one can quickly become familiar with. A phrase with a new meaning can immediately be used together with other words to compose new concepts---\emph{the neighbor's dog, covered in mud, running in the living room} has a clear meaning, even if this scene has never happened. Finally, language is flexible since a phrase can be used to imagine or visualize an associated concept (``generation''), and also to determine whether a particular scene belongs to that concept (``discrimination''). 

\begin{figure}[!h]
\centering
	\resizebox{0.7\linewidth}{!}{
\includegraphics[width=1\linewidth]{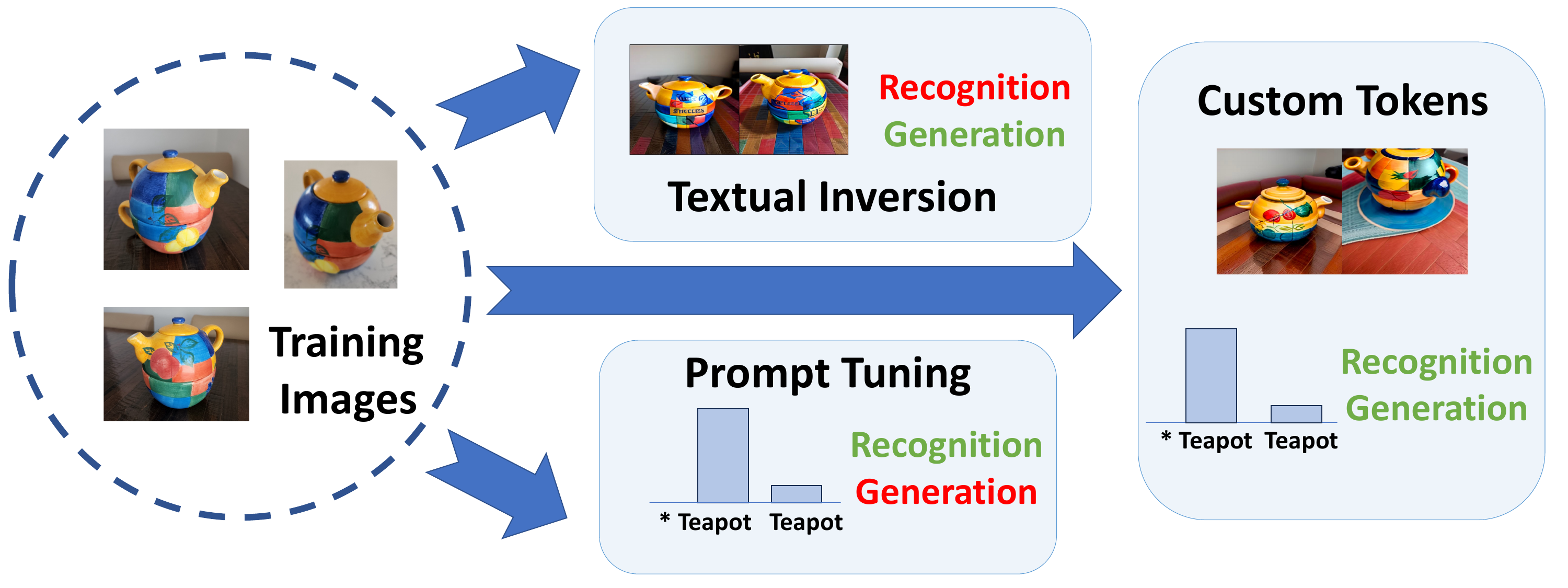}}
    \caption{ Learning custom tokens to represent a given set of images. Prompt tuning and textual inversion learn representations that are optimized for image recognition and generation, respectively. In this paper we examine the possibility of learning {\pseudowords} that can be used for both generation and recognition and also compose with natural language. \vspace{-.3cm}}%
    \label{fig:overview}
\end{figure}

In this work, we focus on the task of learning new tokens to represent custom concepts, for pre-trained VLM~\cite{radfordLearningTransferableVisual2021}.  We assume that each concept is defined in terms of its parent concept and a small number of example images. We refer to these tokens as \emph{\pseudowords} for the remainder of the paper. Learned \emph{\pseudowords}  should behave similarly to words in all of the aspects discussed above: i) exhibit correspondence with images of the same concept in a joint text-image feature space, ii) generate visual samples from the distribution of the underlying concept, and iii) compose well with natural language as shown in Figure~\ref{fig:overview}.

Large-scale Vision Language Models (VLMs) can express complex associations between language and images, but they currently lack this level of flexibility. In particular, while many recent works propose methods for learning concepts from small sets of images~\cite{zhou2022coop,saitoPic2WordMappingPictures2023,cohenThisMyUnicorn2022a,textinversion,ruizDreamBoothFineTuning2022}, each one of these approaches has only focused on a particular aspect of the broader problem. Prompt tuning methods learn a token for discriminative tasks, such as classification~\cite{zhou2022coop} or retrieval~\cite{saitoPic2WordMappingPictures2023,cohenThisMyUnicorn2022a}; on the other hand, text inversion methods learn tokens specialized for image generation~\cite{textinversion,ruizDreamBoothFineTuning2022}. While most of these works also aim to learn tokens that compose well with natural language, the more general question of whether analysis and synthesis tasks can be balanced together simultaneously has not been addressed.

In this work, we aim to train {custom tokens} that compose with natural language and yield prompts that are effective for both image generation and for retrieval/discriminative tasks. Since any prompt constructed compositionally for retrieval is also applicable to image generation, our approach opens the possibility of \emph{visualizing} queries to better understand the retrieval process. In particular, if the visualization of the query does not represent the search intent, a user could apply prompt engineering to change the query until the visualization is consistent with their intent. In this paper, we also introduce an algorithm for \emph{Generation Augmented Image Retrieval (GAIR)}, which can be used to automate this process and improve retrieval results based on a CLIP similarity measure.

In order to learn {\pseudowords} we consider an architecture that can handle discriminative and generative tasks jointly. In particular, we leverage a generative model (Stable Diffusion \cite{stable_diffusion_Rombach_21}) conditioned on CLIP text features \cite{radfordLearningTransferableVisual2021}  and learn {\pseudowords} with a new loss function which combines both discriminative and generative objectives. We show that restricting {\pseudowords} to lie in a lower dimensional subspace spanned by relevant English tokens improves the compositionality of the learned token, \ie, the token can be composed with English language to represent new scenes. Similarly, we show that composing new search queries with the learned token improves text-to-image retrieval performance. Since the composed search query can be visualized, it provides a qualitative measure for the user to determine whether the search query is consistent with the intended search. In summary, our main contributions are as follows:

\begin{itemize} 
    \item We explore a unified framework for learning custom tokens for both generative and discriminative tasks while preserving compositionality with natural language.
    \item We demonstrate that learned custom tokens can generate images of the target concept and produce classifiers that generalize better compared to prompt tuning.
    \item We introduce a Generation Augmented Image Retrieval algorithm that can be used to automatically modify a test query to improve text-to-image retrieval performance.
\end{itemize}

\begin{figure}[!t] \centering
\includegraphics[width=.8\linewidth]{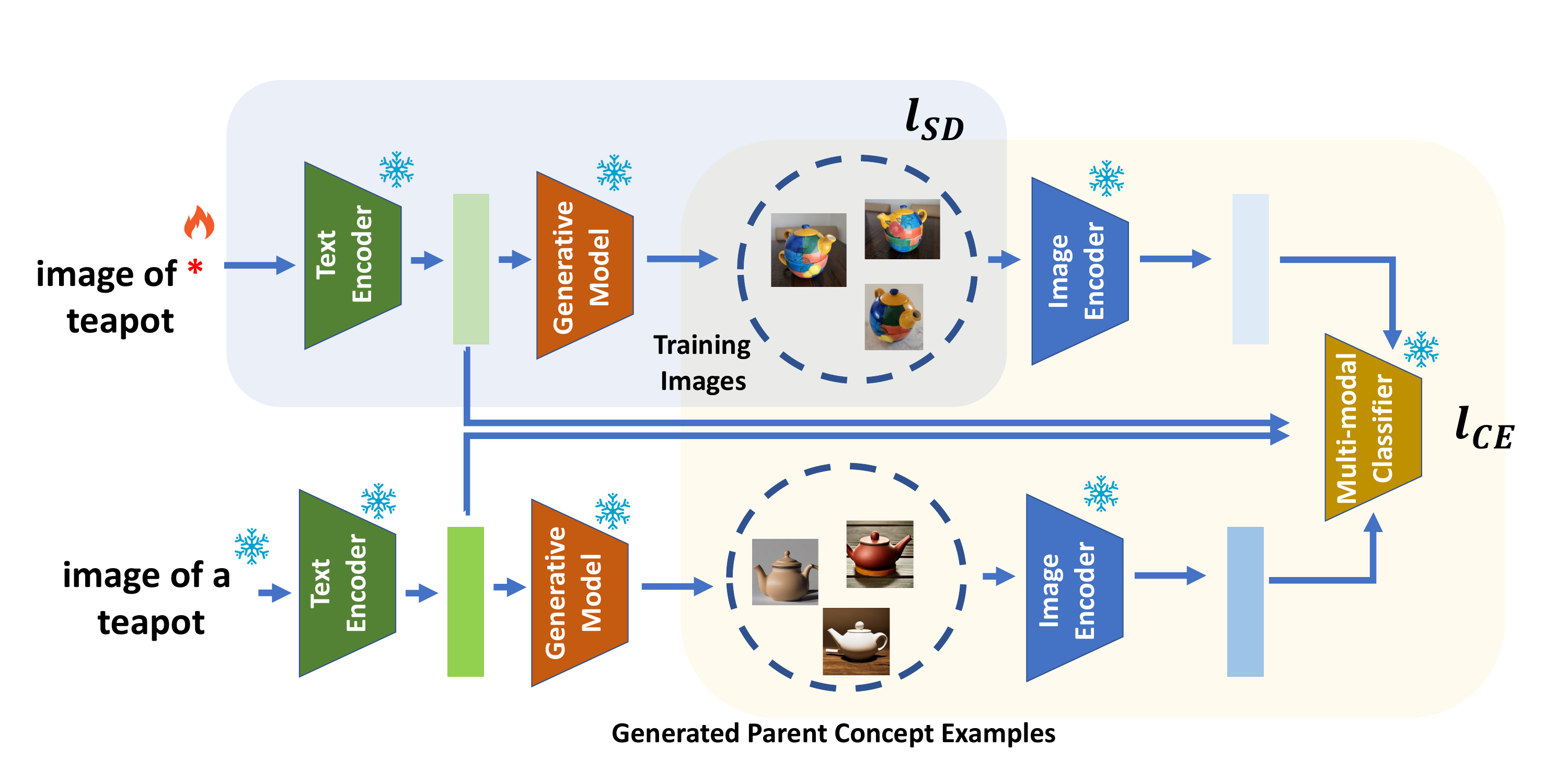}
    \caption{ Overview of the proposed method for learning a custom token for the teapot class. We learn custom tokens that can be used to generate images of the targeted concept and produce discrimination between other concepts. 
     \vspace{-.3cm}}%
    \label{fig:method}
\end{figure}

\section{Related Work}

Diffusion Probabilistic Models (DPMs)~\cite{ddmps-sohl-dickstein15,ddpm} have recently been shown to be remarkably effective for image generation. In particular, models such as DALLE-2~\cite{rameshHierarchicalTextConditionalImage2022}, Imagen~\cite{sahariaPhotorealisticTexttoImageDiffusion2022}, Stable diffusion~\cite{stable_diffusion_Rombach_21} are able to generate high-quality images conditioned on arbitrary textual prompts (text-to-image synthesis). A key component for this success is the availability of vision-language models such as CLIP~\cite{radfordLearningTransferableVisual2021}, that were trained on massive amounts of data and provide strong text-image priors. However, generating peculiar objects that were not observed during training requires \emph{inversion}, i.e., associating example images to the conditioning variables that are able to generate those same images. This problem has been studied extensively in the GAN literature~\cite{creswellInvertingGeneratorGenerative2016,liptonPreciseRecoveryLatent2017,abdalImage2StyleGANHowEmbed2019}. Using diffusion models, \emph{textual inversion}~\cite{textinversion} has been proposed as an approach for inverting 3-5 images into a token (or ``pseudo-word'') for the frozen language model. Although the learned tokens can be used in generic sentences to generate images in new contexts and styles, the compositional abilities are still limited (see Section~\ref{sec:results}). \emph{Dreambooth}~\cite{ruizDreamBoothFineTuning2022} is an alternative inversion method for diffusion models, however it requires fine-tuning the parameters of the diffusion model for each new concept. We also mention \emph{Null-Text Inversion}~\cite{mokadyNulltextInversionEditing2022} that modifies the unconditional textual embedding that is used for classifier-free guidance, rather than the input text embedding. This approach, however, does not easily allow to generate compositions of learned concepts. Finally, several recent methods use inversion-style techniques for flexible image editing~\cite{mokadyNulltextInversionEditing2022, dongPromptTuningInversion2023}.

Prompt tuning is a parameter-efficient fine-tuning method originally proposed in NLP literature~\cite{liuPretrainPromptPredict2021} that optimizes a restricted set of token embeddings in a transformer-based model. The adoption of transformers in Computer Vision has allowed for similar techniques to be used for vision tasks as well \cite{jiaVisualPromptTuning2022, zhou2022coop, wang2022learning}. In particular, in the vision domain, prompt tuning can be performed by optimizing image tokens (VPT~\cite{jiaVisualPromptTuning2022}) or, in the case of vision-language models, text tokens (CoOP~\cite{zhou2022coop}). Learning language tokens opens the door to do compositional manipulations, for example compositions of multiple learned tokens~\cite{nayakLearningComposeSoft2022} or compositions with natural language~\cite{saitoPic2WordMappingPictures2023, cohenThisMyUnicorn2022a}. These methods, however, train tokens to perform classification or retrieval tasks. To our knowledge, the problem of learning tokens that are effective for both generation and classification has not been investigated.

\section{Preliminaries}
\noindent \textbf{Prompt Tuning for Classification}. Prompt tuning \cite{lester-etal-2021-power, wang2022learning} allows the output of a Transformer-based model to be influenced by a learnable token embedding. \cite{zhou2022coop} adopted prompt tuning for image recognition tasks by forming input prompts of the form $\tau_i = [* , CLASS_i ]$, where `$\ast$' is a nominal token corresponding to the learnable token embedding $e_*$. However, this style of prompting can only be used with known-concepts, since concepts are specified in terms of natural-language. \cite{nayakLearningComposeSoft2022} introduced an alternative form of prompting that can be used to learn tokens for new concepts. In their version, the text prompt is formed as $\tau_i = [t, *_i]$ for each concept $i$, where $t$ was a nominal text of the form `image of a'.  During training, embedding vectors $e_{*_i}$ are optimized with a CLIP-style contrastive loss by considering the similarity between the text feature $\tenc(\tau_i)$ and the corresponding image feature,  where $\tenc$ is the text encoder.

\noindent \textbf{Diffusion Based Generation}.
Latent Diffusion model operates on principles of Denoising Diffusion Probabilistic Models (DDPMs)~\cite{ddpm} where images are first projected to a latent space through an autoencoder. The autoencoder consists of an encoder $\epsilon$ and decoder $\delta$. A diffusion model is trained to produce learned latent codes and can be conditioned on an external variable $\tau$. During the reverse diffusion process, at time step $t$, a noise vector $\eta$ is  sampled from the standard normal distribution. Parameters $\theta$ of the denoising network $\nu_\theta$ are optimized by minimizing the loss $l_{\text{DM}} = E_{\eta \sim N(0,1), z \sim \epsilon(x) ,t} \big[|| \eta - \nu_\theta (z, t, \tau) ||_2^2\big].$ With this approach, the network learns to remove the noise added to the image during the forward diffusion process. Then, at inference time, a sample from the trained model is produced drawing a noise vector from $N(0,1)$ which is sequentially denoised for $T$ steps using the denoising network $ \nu_\theta $ conditioned with the input $\tau$ to produce a latent vector $z_0$. The output image $\hat{x}$ is produced  by passing the latent vector $z_0$ through the decoder network $\delta$. We conceptualize the entire image generation pipeline by the process $d(\tau)$, where    $\hat{x} = d(\tau) = \delta(z_0)$. The latent diffusion was initially proposed for text-to-image generation where the generator was conditioned on BERT features. More recently, Stable Diffusion has been introduced where the text conditioning uses a CLIP text encoder \cite{stable_diffusion_Rombach_21}.

\noindent \textbf{Textual Inversion}.
Textual inversion \cite{textinversion} is a method based on diffusion models that learns a new token with the objective of generating images of a novel concept. Textual inversion originally used latent diffusion in their architecture and thus conditioned the generation process using a BERT text encoding.  The model is trained by feeding training images $\mathbf{x}$ to the network while using $\tau = g([t, *])$ as conditioning, where $\ast$ is a new token added to the vocabulary. To avoid overfitting, a set of different paraphrasing of $\tau = \text{`image of a *'}$ is used. During training, the parameters of the diffusion model $\theta$ and text encoder $\tenc$ are kept frozen while the diffusion loss $l_{DM}$ is minimized by optimizing the embedding $e_*$, where $e_*$ is the embedding of the token $\ast$.

\section{Proposed Method}

With multi-modal models, it is possible to use natural language to specify concepts for visual tasks (\eg, in classification and generation). Despite the flexibility and convenience of these models, the quality of the results is limited by 1) the ability of the user to express their exact intent in a textual prompt 2) the capability of the model to interpret such prompt correctly. Both of these issues are especially significant in the case of fine-grained image classification or generation and/or in the case of customized concepts, since textual prompts may be unsuited for conveying visual details and the model may have never seen a personalized concept during training (\eg, “my dog”). To address both issues, we propose to define new custom tokens that can be composed with plain text tokens to query VLMs both for image generation and classification. Our learned tokens are multi-modal in that they are defined through the use of positive images (different images of the desired concept) and coarse textual description. 

In particular, we assume we have access to a set of $n$ positive images $ \mathbf{x} = [x_1, x_2, \dots ,x_n]$ (where $n$ is generally 3-5) of the desired concept drawn from an image distribution $ \mathbf{\mathcal{X}}$ and an associated textual description of it given by a super-concept label $c$. 
Given a text encoder $\tenc$, image encoder $\ienc$ and a diffusion model $\D$, our task is to learn a customized word-token (that we denote with $\ctok$) expressed as the token embedding $e_{\ctok}$ that (i) can be used both for image analysis and synthesis tasks and (ii) that is composable with other English tokens in the pre-trained vocabulary of a given VLM. 
In particular, our customized word-tokens are designed so that they can be used for:

\begin{itemize}
    \item \textbf{Generation}: Given a context text $t$, our custom token $\ctok$ can be composed with the textual query and can be passed as input to a text to image generative model: 
    $\D \big(\tenc([t, c, \ctok])\big) \longrightarrow x_i \in \mathbf{\mathcal{X}} $, where $t$ is the text context and we use the square brackets to denote the tokens concatenation. For example, it is common to use different paraphrasing of ``an image of a'' for $t$.
    \item \textbf{Recognition}: Given a context text $t$, negative class labels $y_i$, our custom token $\ctok$ can be used to discriminate the custom concept from negative classes: $\tenc([t, c, \ctok])^T \ienc (x_j) >  \tenc([t, c, y_i])^T \ienc(x_j) \quad \forall i,j$, where we parametrize the scoring function for classification as the product of text/image embedding vectors (following CLIP dual encoder parametrization \cite{radfordLearningTransferableVisual2021}).
\end{itemize}

\begin{figure} 
      \centering
	\resizebox{0.8\linewidth}{!}{

      \begin{tabular}{@{}c|ccc|ccc|ccc@{}}
      \toprule
      &
    \multicolumn{3}{c}{Image of a $\ast$ teapot on a table} &
    \multicolumn{3}{c}{ Image of a $\ast$ teapot on a sink} &
    \multicolumn{3}{c}{ Image of a $\ast$ teapot on a beach} \\
    \hline
    
            TI(BERT) &
        \includegraphics[width=0.11\linewidth]{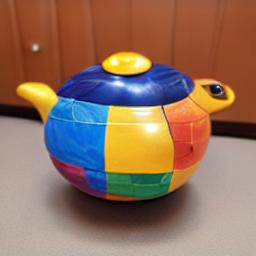} & \includegraphics[width=0.11\linewidth]{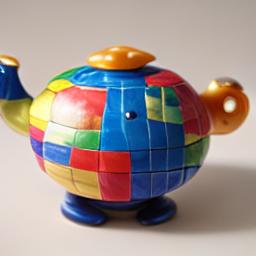} & \includegraphics[width=0.11\linewidth]{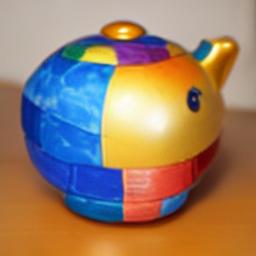} &
        \includegraphics[width=0.11\linewidth]{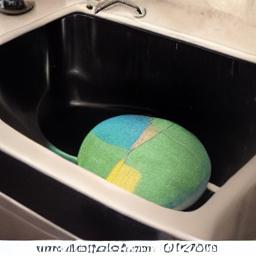} & \includegraphics[width=0.11\linewidth]{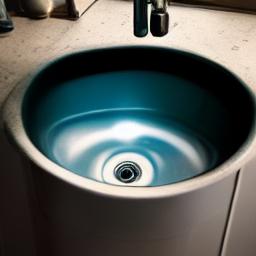} & \includegraphics[width=0.11\linewidth]{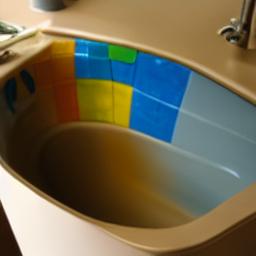} &
        \includegraphics[width=0.11\linewidth]{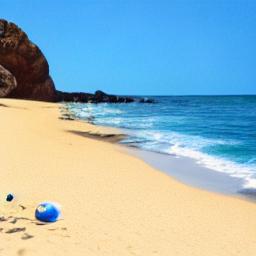} & \includegraphics[width=0.11\linewidth]{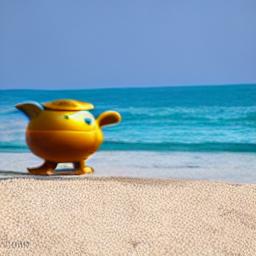} & \includegraphics[width=0.11\linewidth]{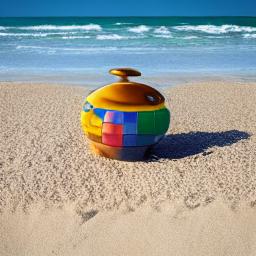} \\
        
      TI(CLIP) &
      \includegraphics[width=0.11\linewidth]{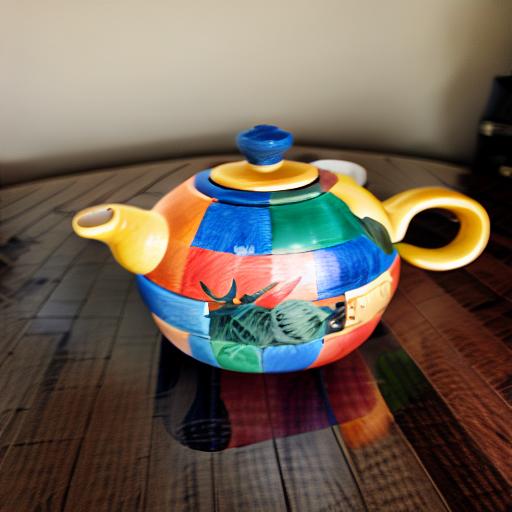} & \includegraphics[width=0.11\linewidth]{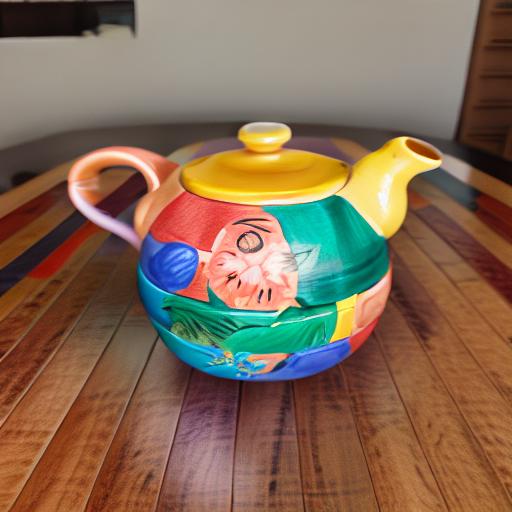} & \includegraphics[width=0.11\linewidth]{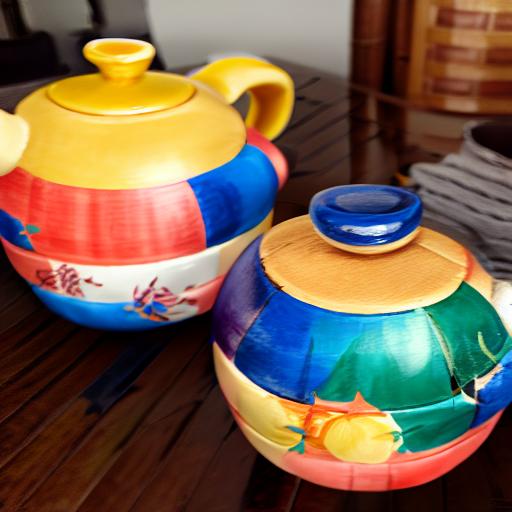} &
        \includegraphics[width=0.11\linewidth]{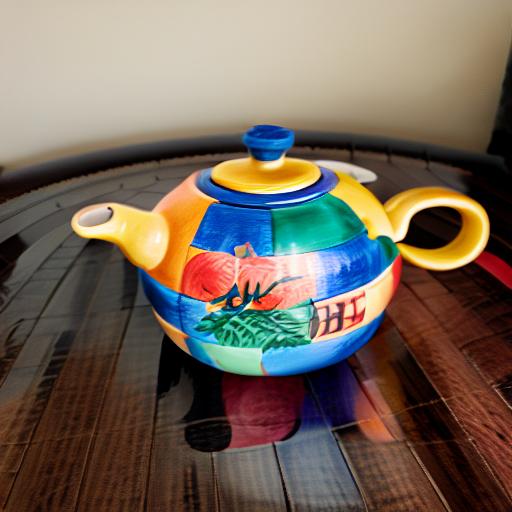} & \includegraphics[width=0.11\linewidth]{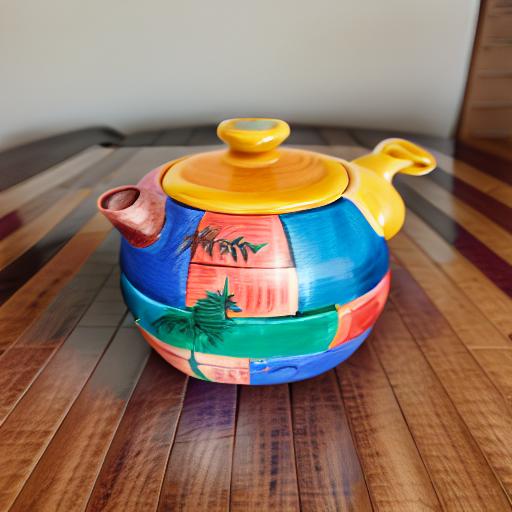} & \includegraphics[width=0.11\linewidth]{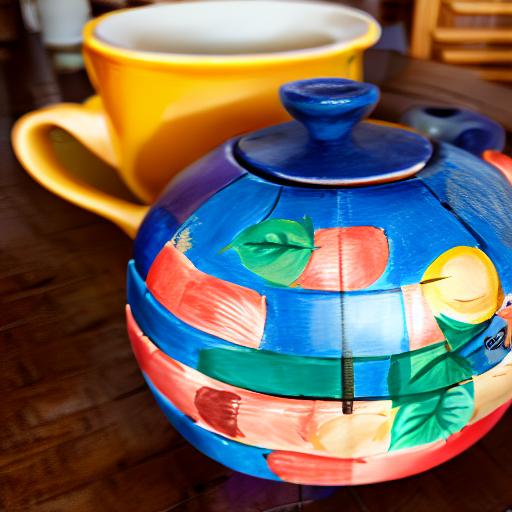} &
        \includegraphics[width=0.11\linewidth]{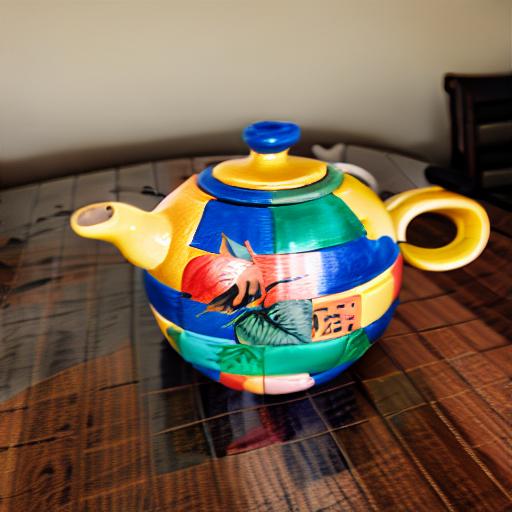} & \includegraphics[width=0.11\linewidth]{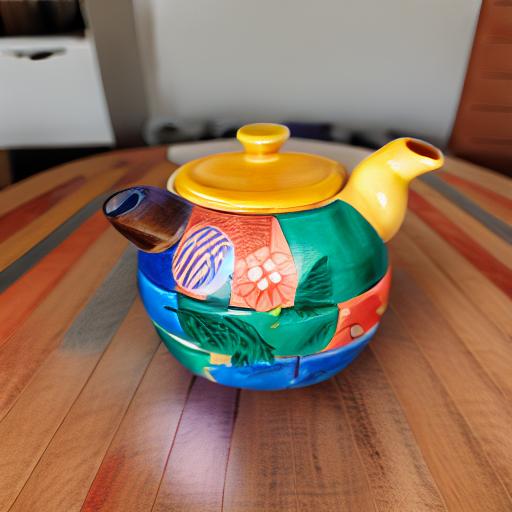} & \includegraphics[width=0.11\linewidth]{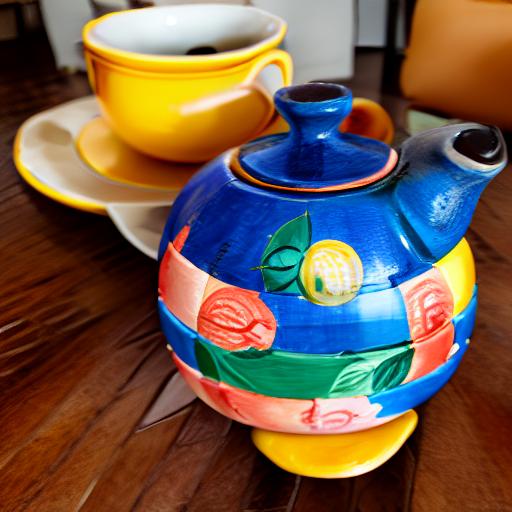} \\

        Ours(CLIP) &
        \includegraphics[width=0.11\linewidth]{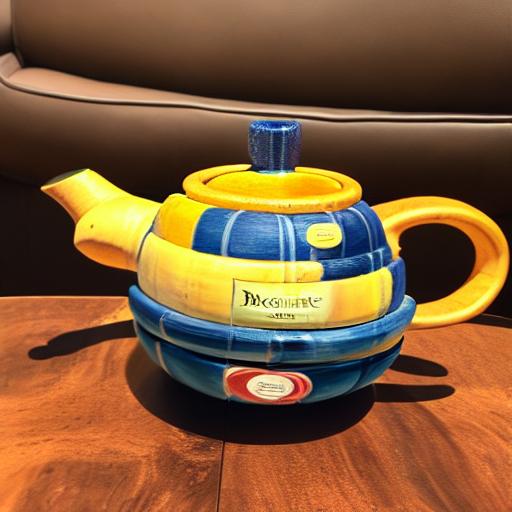} & \includegraphics[width=0.11\linewidth]{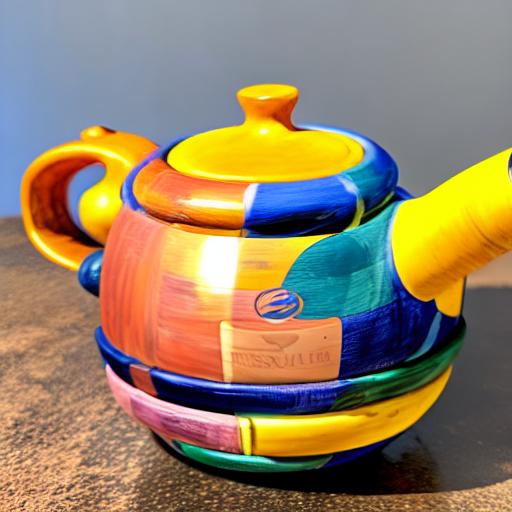} & \includegraphics[width=0.11\linewidth]{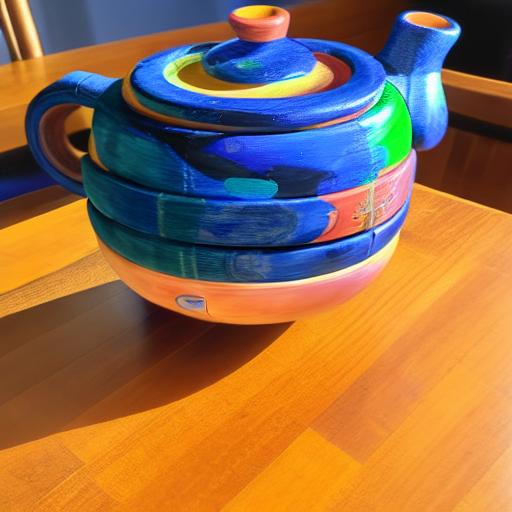} &
        \includegraphics[width=0.11\linewidth]{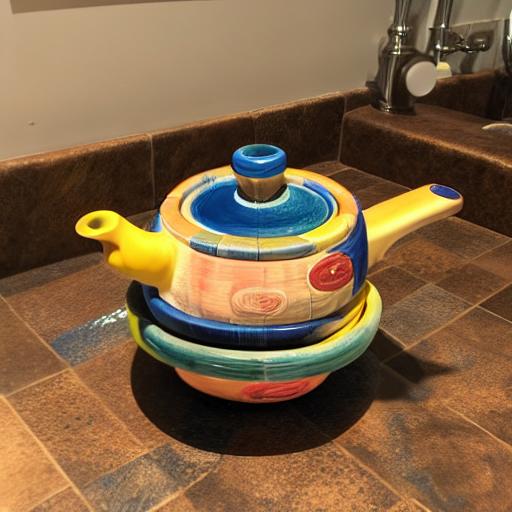} & \includegraphics[width=0.11\linewidth]{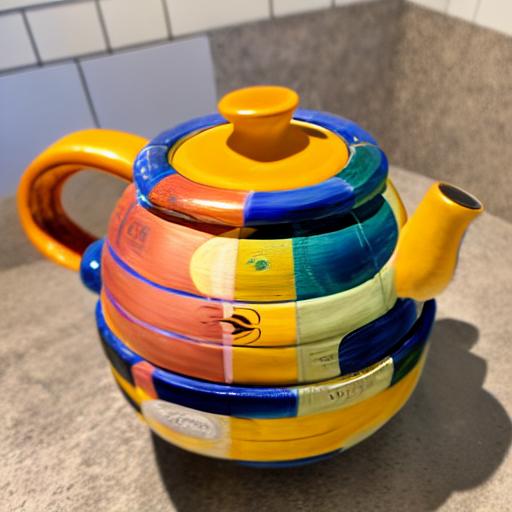} & \includegraphics[width=0.11\linewidth]{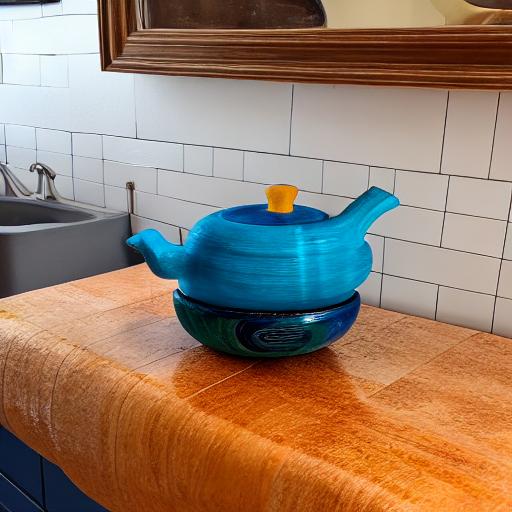} &
        \includegraphics[width=0.11\linewidth]{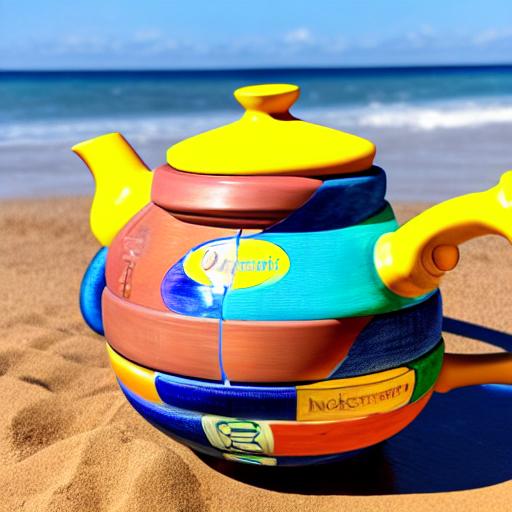} & \includegraphics[width=0.11\linewidth]{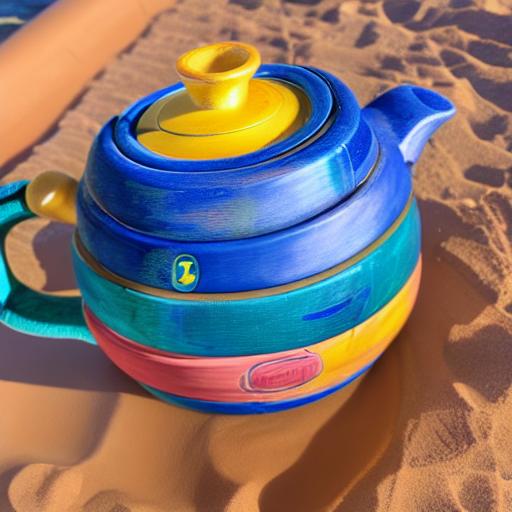} & \includegraphics[width=0.11\linewidth]{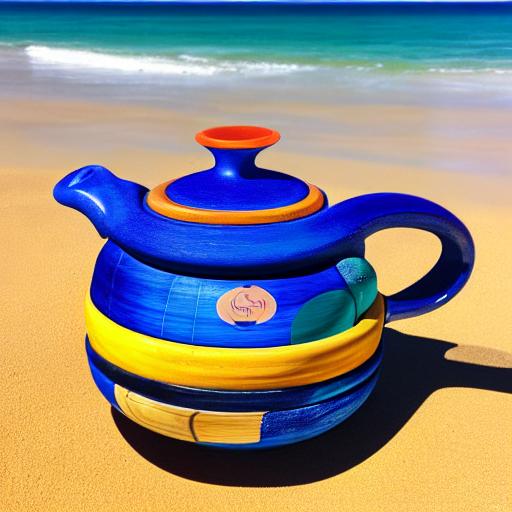} \\
        \hline
        
      \end{tabular}} \caption{Textual Inversion conditioned on CLIP does not compose well with natural language. Our \pseudowords{} compose better with other text after subspace projection is performed.} \label{fig:ti_comp}
\end{figure} 

\begin{figure}[ht] 
\centering
\resizebox{.9\linewidth}{!}{ 
    \qquad \hspace*{-0.7cm}
    \subfloat[\centering]{{\includegraphics[width=.24\linewidth ]{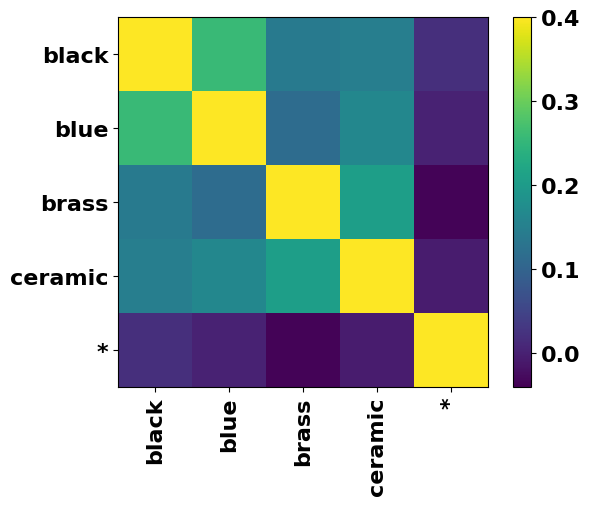} } \label{fig:np_mtx}} 
    \subfloat[\centering]{{\includegraphics[width=.24\linewidth ]{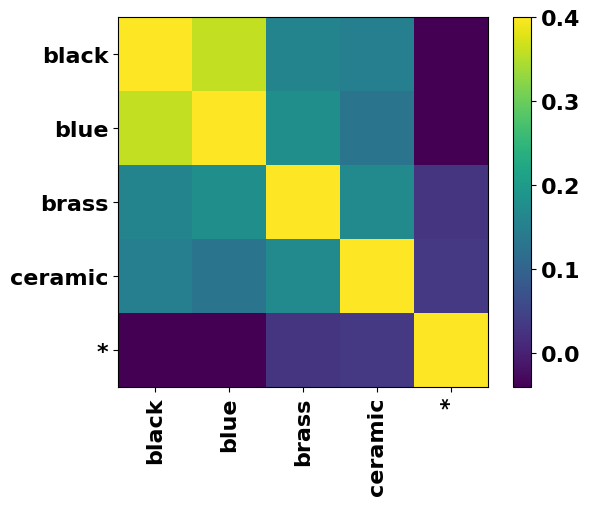}} \label{fig:ce_mtx}}
    \subfloat[\centering]{{\includegraphics[width=.24\linewidth ]{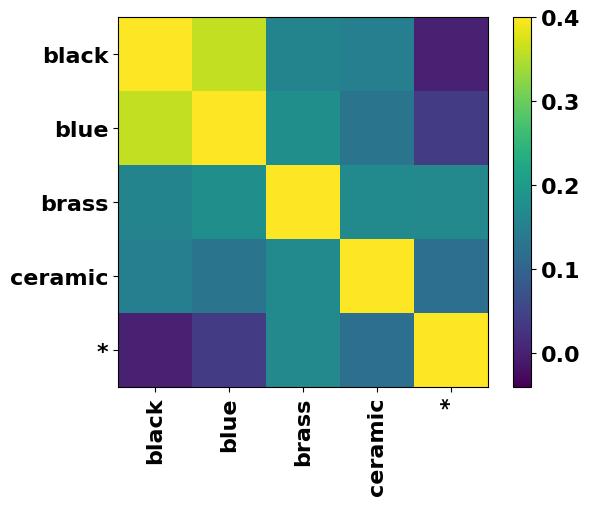} }\label{fig:pw_mtx}}%
    \subfloat[\centering]{{\includegraphics[width=.20\linewidth ]{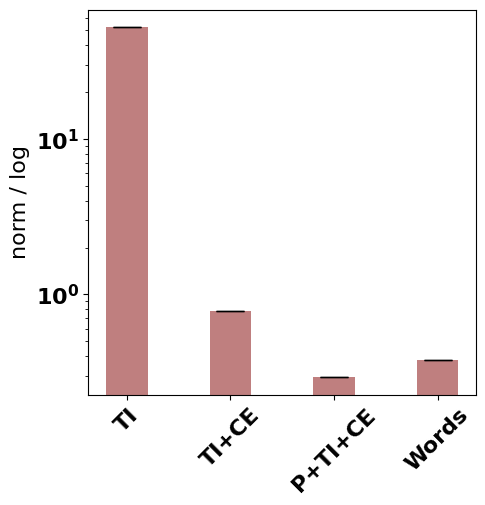} }\label{fig:normplot}}%
     }
    \caption{ Cosine similarity between normalized text embeddings and the learned token embedding (last column) for  (i) textual inversion: TI (ii) textual inversion + cross-entropy loss: TI+CE (iii) subspace projection + cross-entropy loss: P+TI+CE (iv) norm of learned word embedding with different methods. Image intensity clipped at 0.4 for visualization. }%

\end{figure}

\begin{table} 
        \centering
\caption{Classification performance on Caltech256-teapot class and generated images when the weights of the loss is varied. \vspace{-0cm}}
	\resizebox{0.7\linewidth}{!}{ 
      \begin{tabular}{@{}lccccc@{}}
      \toprule
      Weights $(\lambda_{SD}, \lambda_{CE})$ & (0,1) & (1, 0.1) & (1, 0.001) & (1, 0) \\ \midrule
      Accuracy (teapot) & 98.5\% &  99.2\% & 93.3\% & 100\% \\ 
      Accuracy ($\ast$ teapot) & 100\% &  100\% & 100\% & 0.0\% \\[.1cm] 
        \raisebox{.5cm}{\multirow{1}{*}{Generated Images}}
         &          
        \includegraphics[width=0.2\linewidth]{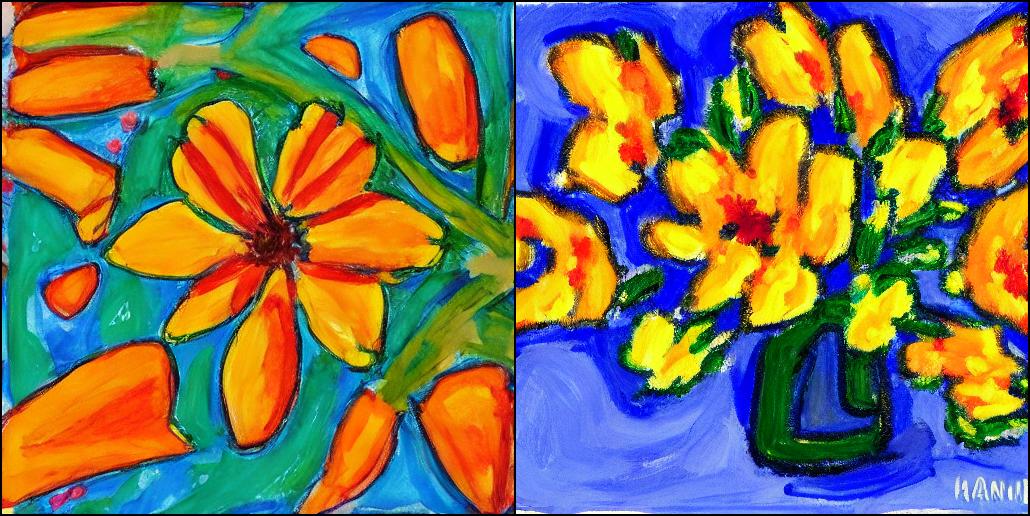} & \includegraphics[width=0.2\linewidth]{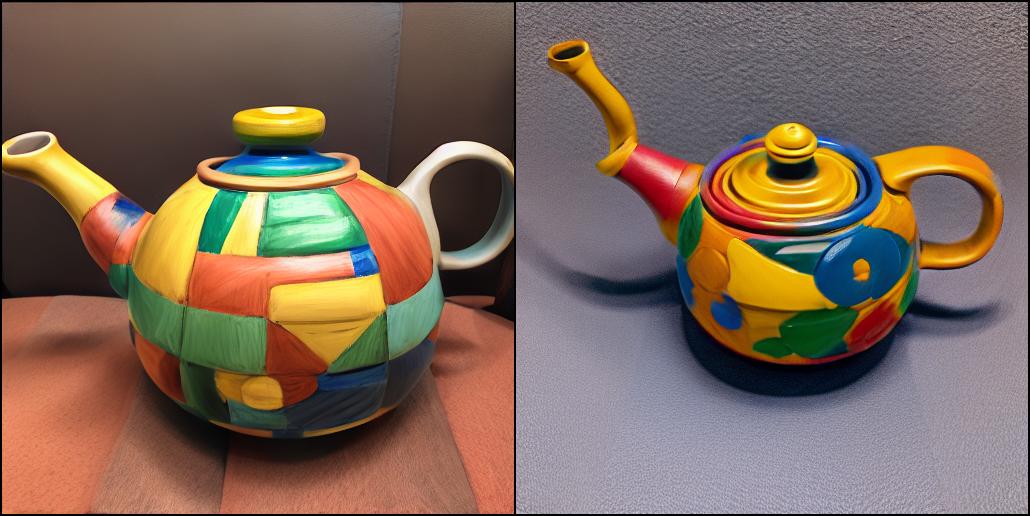} & \includegraphics[width=0.2\linewidth]{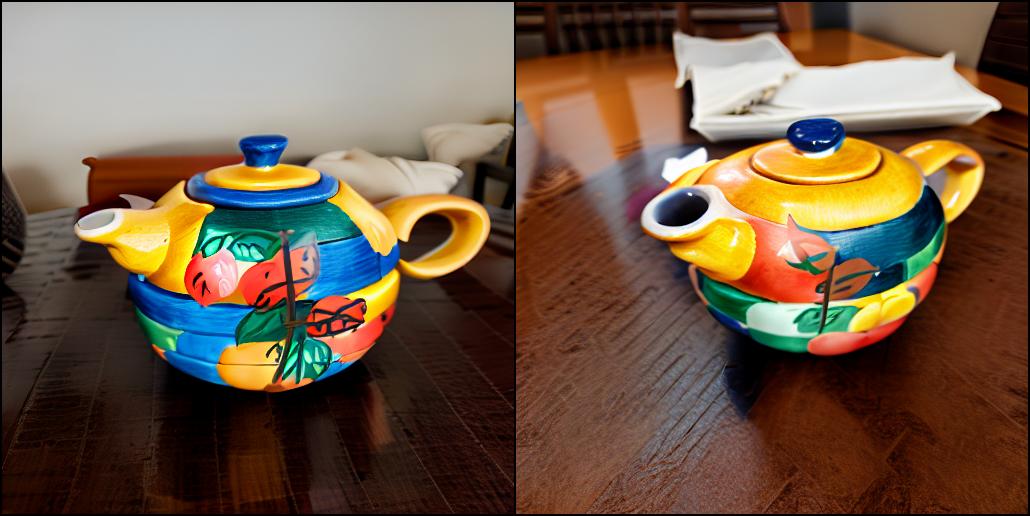} & \includegraphics[width=0.2\linewidth]{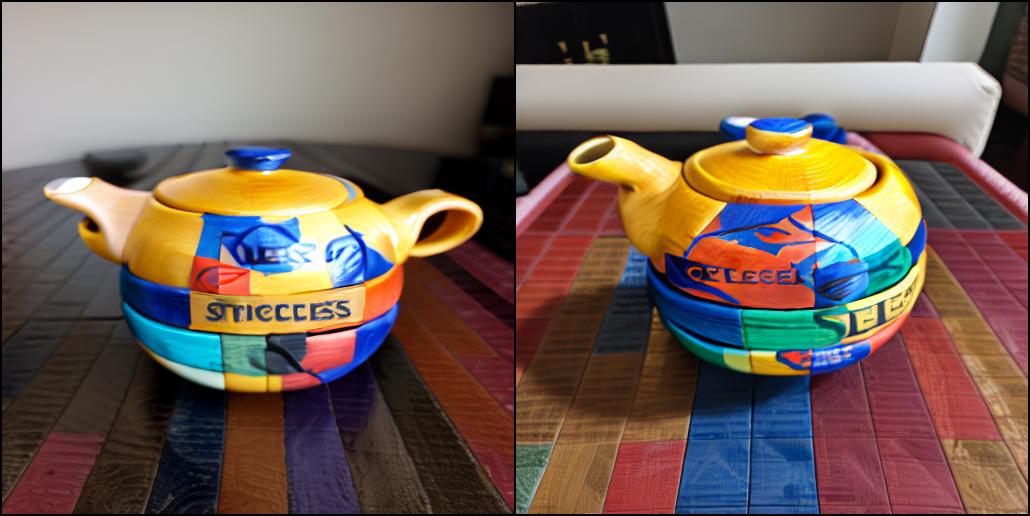}  \\
        \bottomrule
      \end{tabular}} 
      \label{fig:weight_ablation}
\end{table}


The outline of the proposed solution is illustrated in Figure~\ref{fig:method}. Our method has two main components: a textual inversion network and a classification network. The token embedding $e_*$ is trained using textual inversion network (shaded in blue) such that suitable prompts containing the token generate images of the targeted concept. When CLIP's text encoder is used as $\tenc$, one would expect that its latent feature may be useful for classification. However, its recognition performance tends to be poor as we later show in experimental results (Section~\ref{sec:results}). We therefore use losses of classification network (yellow) jointly with the textual inversion network to learn token embeddings that provides stronger joint classification/generation performance. Furthermore, we show that our joint training helps regularize discriminative tokens learned with prompt tuning and leads to a classifier that generalizes better than the prompt-tuning paragon \cite{zhou2022coop}.

\subsection{Textual Inversion Network} \label{sec:clip}

We consider the network proposed in \cite{textinversion} as the blueprint to our textual-inversion network. Although this setup works well for learning image-tokens for generation, since text features are encoded with BERT, there is no direct way of integrating the model with the image-to-text classifier as shown in Figure~\ref{fig:method}. To facilitate this operation, we modify the textual inversion model by replacing its backbone with a Stable Diffusion Model \cite{stable_diffusion_Rombach_21} that uses a frozen CLIP ViT-L/14 text encoder for conditioning. By making this choice, we are able to obtain text-to-image scores by comparing CLIP text features with image feature of a given image. In this setup, $\tenc$ and $f$ are CLIP text and image encoders, $d$ is a stable-diffusion model.

With this change, the generated \pseudowords{}  are able to faithfully generate images similar to the training set. However, the learned model has several issues. In Figure~\ref{fig:ti_comp} we visualize images sampled from a  model trained on the concept shown in Figure~\ref{fig:overview} with 5 tokens. Although the model appears to have learned structure and textures of the target concept,  Figure~\ref{fig:ti_comp} suggests that the learned token (TI-CLIP) doesn't compose well with natural language. For example, the image generated for the caption `image of a $\ast$ teapot on a beach' looks very similar to the image generated for `image of a $\ast$ teapot on sink'. On the other hand, we observed that the learned token cannot be used to differentiate images of the target concept from images of its parent-concept (see Section~\ref{sec:align} for more details). 

In order to understand the reason behind these shortcomings, we investigated the embedding of the learned token. First, we considered a set of attributes that could be used to describe the parent concept. For example, black, blue, ceramic \emph{etc.} can be regarded as valid attributes of the parent-concept teapot. We considered average token embeddings for each attribute and compared it with the average learned embedding in terms of cosine similarity. In Figure~\ref{fig:np_mtx} we visualize an extraction of the obtained affinity matrix. According to this figure (final row and column), it seems that the learned embedding is significantly different from other attribute embeddings. In Figure~\ref{fig:normplot} we compare the distribution of norms of each attribute embedding considered  against the average norm of the learned embedding. This figure suggests that the the learned embedding has a norm $> 50 \times$ than the average attribute embedding norm. 

These findings suggest two potential issues with learned embedding: i) the learned embedding has a significantly larger norm compared to other word embeddings. Since word embeddings are not normalized before passing to the encoder, the learned embeddings could negate the contribution from the context. 
ii) Even when embeddings are normalized, the learned token has low similarity to other adjectives/attributes that are known to compose well with the parent concept.

\subsection{Classification Network} \label{sec:align}



To improve the classification performance of the learned token, we propose to add a regularization term to the loss function through a classification loss. First, we sample $k$ images $\mathbf{x'} = [x'_1, x'_2, \dots , x'_k]$ from the diffusion model with the prompt `image of a $c$', where $c$ is is the parent concept provided. We treat these images as negative images during training. We form a secondary training set by considering images $\mathbf{x''} = [\mathbf{x}, \mathbf{x'} ]$ with corresponding ground truth labels $\mathbf{y} = [\mathbf(1)\in \mathcal(R)^n, \mathbf(0)\mathcal(R)^k]$. During training we consider two prompts of the form `image of a * c' and `image of a $c$' and obtain their text features $\tau_*$ and $\tau_{parent}$ respectively. We then compute a class scores  $\ienc(\mathbf{x''}) \cdot \tau_*$ and  $\ienc(\mathbf{x''}) \cdot \tau_{parent}$ and evaluate a classification loss with balanced binary cross entropy $l_{CE} = \text{BCE}(\mathbf{x''} ,\mathbf{y} )$. Our final training loss becomes a linear combination of stable-diffusion loss $l_{SD}$ and classification loss $l_{CE}$ in the form $\lambda_{SD} l_{SD}+ \lambda_{CE} l_{CE}$. 

In Figure~\ref{fig:weight_ablation}, we illustrate samples generated with the prompt `image of a $\ast$ teapot' when weights $\lambda_{SD}$ and $\lambda_{CE}$ are varied. Here, we report how well the learned embedding is able to recognize the target concept from its parent-concept (teapot from the Caltech256 dataset). Note that the cases at either ends with weight (0,1) and (1,0) are equivalent to prompt tuning and textual inversion respectively. For pure textual inversion case, it appears that balanced-classification accuracy is merely 50\%. According to Figure~\ref{fig:weight_ablation}, increasing weight $\lambda_{CE}$ increases overall classification accuracy. In addition to improving recognition performance, adding regularization adds more structure to the learned embedding. In Figure~\ref{fig:normplot}, we plot the norm of the embedding (TI+CE). According to the plot, adding the regularization has  caused norm of the embedding to drop significantly. However, the correlation between the learned embedding with other attribute embeddings remains poor as shown in Figure~\ref{fig:ce_mtx}.

\subsection{Improving Compositionality}

In Section~\ref{sec:clip}, we considered possible adjectives/attributes of the parent-concept. It should be noted that the parent-concept composes well with these adjectives to form more complex concepts (\eg, ceramic teapot, red teapot, red ceramic teapot). We hypothesize that the embeddings of these adjectives lie in a lower-dimensional sub-space where each embedding vector meaningfully composes well with the parent concept. If this is true, we argue that by enforcing the learned embedding to lie in the same subspace we may increase the possibility of it composing well with natural language. In order to enforce this constraint, we first form a matrix $\mathbf{e}$ with average embeddings of all the selected adjectives. We calculate the mean vector $\mu_\mathbf{e}$ of the matrix $\mathbf{e}$  and perform Principal Component Analysis to find a projection matrix $P$. During training we transform the token $e_*$ using the linear transformation $P(e_*-\mu_\mathbf{e})+\mu_\mathbf{e}$. 

In Figure~\ref{fig:normplot}, we plot the norm of the learned embedding under TI+CE+P. We see that the linear projection has further reduced the norm of the embedding from before - and now lies withing one standard deviation of the word embeddings. In Figure~\ref{fig:pw_mtx}, we illustrate the correlation between the learned embedding and other attribute embeddings. Compared to other alternatives, it can be seen that learned embedding is now much more similar to other word embeddings. In Figure~\ref{fig:ti_comp} (Ours-CLIP), we visualize some of the images generated when subspace projection is employed. According to the Figure, backgrounds of generated images have changed with respect to the prompt. However, it can be seen that preserving fidelity to original concept is challenging in some cases, particularly when composing with other words. We study this phenomenon in detail in the experimental results section.

\subsection{Generation Aided Image Retrieval(GAIR)}
Our \pseudowords{} can be composed with natural language and be used for both discriminative and generative tasks. These properties can be particularly useful for image retrieval. Indeed, a user can construct a query $[t,*,c]$ in natural language which includes the learned token, and also then \emph{visualize} the query and verify whether it is consistent with the intended search. This kind of verification is useful to monitor two kinds of potential issues: 1) the token may have not encoded the target concept correctly (for example, spurious common features of the training images may have been incorporated), 2) the token might not compose well with the textual query. Observing sample images generated from the query allows the user to easily identify these issues and correct for them. 

Once an issue has been identified, a simple approach to improve the query would be to try paraphrasing the text via prompt engineering. An alternative method that is more easily controllable is to manually amplify contribution from certain words to alter the generated image. \cite{Trager2023} showed that CLIP textual features of a sentence can be expressed as linear feature combination of features associated with its constituent words. Drawing inspiration from this observation, we express a query $q$ as a linear combination of a query with and without the learned token, that is, $q = w g([t, * , c]) + (1-w) \sum_i \frac{1}{|A|}g([t, a_i, c ]) $, where $a_i \in A$ is a set of related attributes. Note that when $w=1$, the query $q$ is equal to the original search query $g([t,*,c])$.

This formulation allows the user to give more importance to some part of the context by reducing the weight $w$ as needed. An optimal value for $w$ for a given query can be chosen by visualizing queries for different choices of $w$. When human supervision is not feasible, this process can be automated based on CLIP score using the Algorithm~\ref{alg:cap}.  Here, for different values of $w$, we formulate the conditioning vector $q$ and use it to generate images. In each case we check whether the generated image contains i) the object under consideration,  by comparing it to original training images $\mathbf{x}$ in CLIP space and ii) the context described by the caption, by comparing the image to context images $\mathbf{x_c}$ in the CLIP space. Context images are generated with the same query where  custom tokens are replaced with a related attribute as in $d(\sum_{a_i \in A} \frac{1}{|A|} g([t, a_i, c ]) $. We select the value of $w$ that had produced the largest value of the minimum CLIP score for context and object. Note that taking the minimum of $f(I)^T f(\mathbf{x}) $ and $f(I)^T f(\mathbf{x_c}) $ ensures that the chosen parameter produces images that has reasonably good CLIP score for both objects and contexts.




\begin{algorithm}
\centering
\caption{Generation Aided Image Retrieval }\label{alg:cap}
\begin{algorithmic}
\Require training images $\mathbf{x}$, parent concept images $\mathbf{x_c}$, caption $t$
\State $S \gets []$ \Comment{Scores list}
\State $W\gets [w_0, \dots, w_N]$ \Comment{Weights list}
\For{$ w \in W$}
    \State $q \gets w g[t,*,c]+ (1-w) \frac{1}{|A|} \sum_{i} g[t,a_i,c]$
    \State Generate image $I \gets d(g[q])$
    \State $score \gets \min(f(I)^T f(\mathbf{x_c}), f(I)^T f(\mathbf{x}))$
    \State $S.\text{append}(score)$ \Comment{Update scores list}
\EndFor
\State $w^* \gets W[\argmax(S)]$ \Comment{Optimal weight}
\end{algorithmic}
\end{algorithm}

   

   

   


\section{Experimental Results} \label{sec:results}
In this section we present  qualitative results in terms of generation quality, quantitative results with respect to recognition performance on the Textual Inversion dataset \cite{textinversion} and text-to-image retrival performance on DeepFashion2\cite{ge2019deepfashion2} dataset. As baseline comparisons we consider i) textual inversion with a BERT encoder \cite{textinversion}, ii) textual inversion with a CLIP text encoder,  and iii) CLIP based prompt tuning \cite{nayakLearningComposeSoft2022}.
To make our comparison fair we add to the prompt tuning objective negative images sampled with the Stable Diffusion model using the superclass textual information as input conditioning.\\


\noindent \textbf{Model details.} For all the experiments we used 10 trainable tokens. For main results section, we considered  100 attributes from CQGA dataset \cite{Naeem_2021_CVPR} that correlated most with the given concept in the CLIP space to define the lower dimensional sub-space and set $\lambda_{CE}=10^{-5}$, $\lambda_{SD}=1$. We used an effective batch size of 4 with a learning rate of $5 \cdot 10^{-4}$. We trained each model for 20000 iterations. We used the publicly available Stable Diffusion 1.4 trained on LAION dataset as the backbone model.\\

\noindent \textbf{Datasets.} The Textual Inversion dataset contain multiple classes with 5-6 images per class. In our experiments we trained models using 3 images for training and used remaining images for inference. We carried out quantitative experiments on classes that had parent classes in common with Caltech256 dataset. We used the Caltech256 dataset to simulate negative classes since it contains all superclass concepts for all the classes present in the Textual Inversion dataset. In addition, we experimented on DeepFashion2 dataset\cite{ge2019deepfashion2} - which consists of clothing products where product identity is available. We used the PerVL benchmark\cite{cohenThisMyUnicorn2022a} to obtain train-test splits and image captions for the dataset. We used all available training images per class when training on this dataset.\\

\begin{figure*} 
  
    \centering
	\resizebox{0.9\linewidth}{!}{
      \begin{tabular}{@{}c|ccc|ccc|ccc||c@{}}
      \toprule 
     Method &
    \multicolumn{3}{c}{Image of a $\ast$} &
    \multicolumn{3}{c}{ Image of a $\ast$ on a sink} &
    \multicolumn{3}{c}{ Image of a $\ast$ on a beach} & Class Samples \\
    \hline
        TI (BERT) &
        \includegraphics[width=0.11\linewidth]{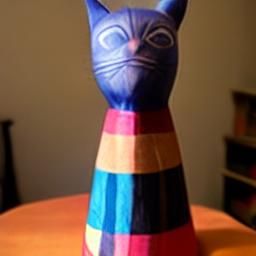} & \includegraphics[width=0.11\linewidth]{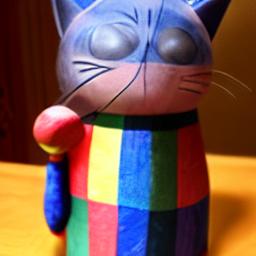} & \includegraphics[width=0.11\linewidth]{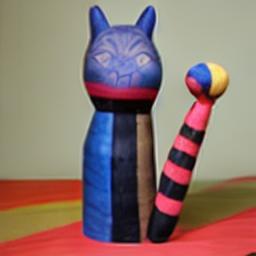} &
        \includegraphics[width=0.11\linewidth]{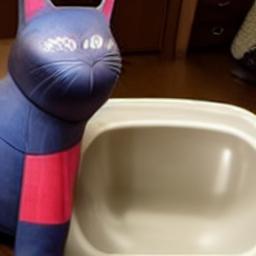} & \includegraphics[width=0.11\linewidth]{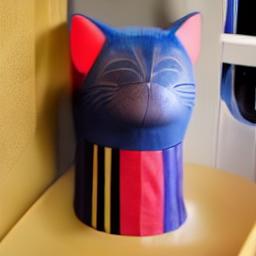} & \includegraphics[width=0.11\linewidth]{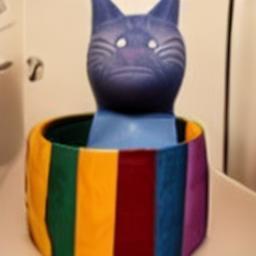} &
        \includegraphics[width=0.11\linewidth]{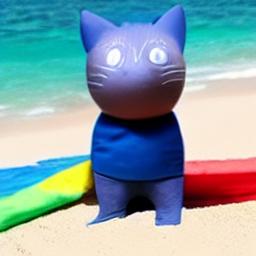} & \includegraphics[width=0.11\linewidth]{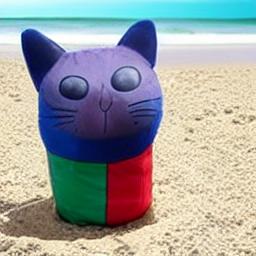} &
        \includegraphics[width=0.11\linewidth]{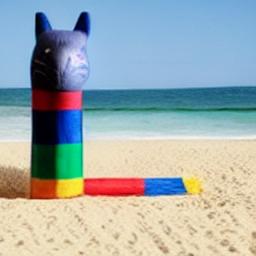} & \includegraphics[width=0.11\linewidth]{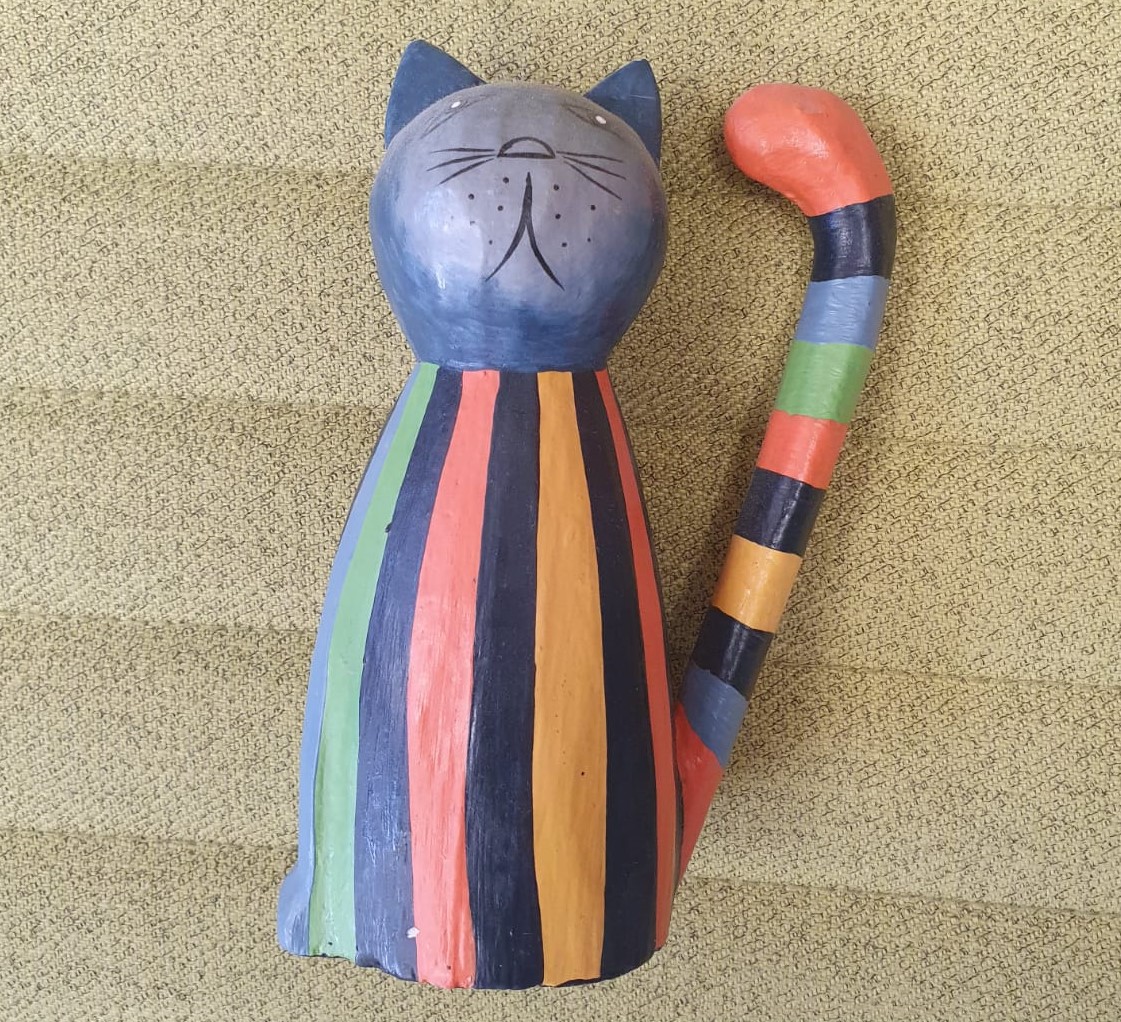} \\
        TI (CLIP) &
        \includegraphics[width=0.11\linewidth]{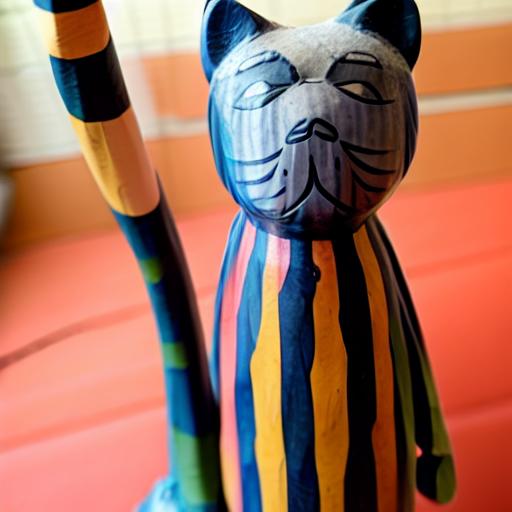} & \includegraphics[width=0.11\linewidth]{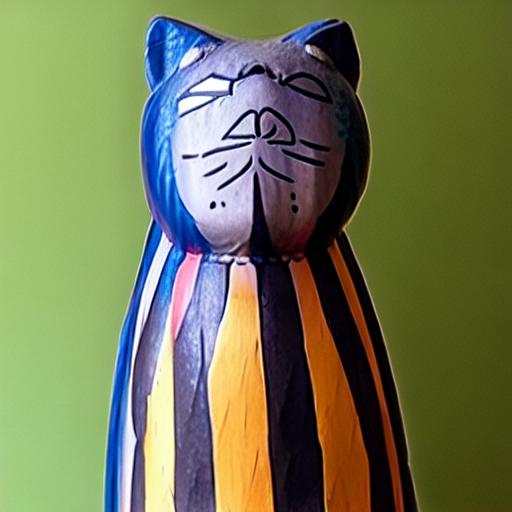} & \includegraphics[width=0.11\linewidth]{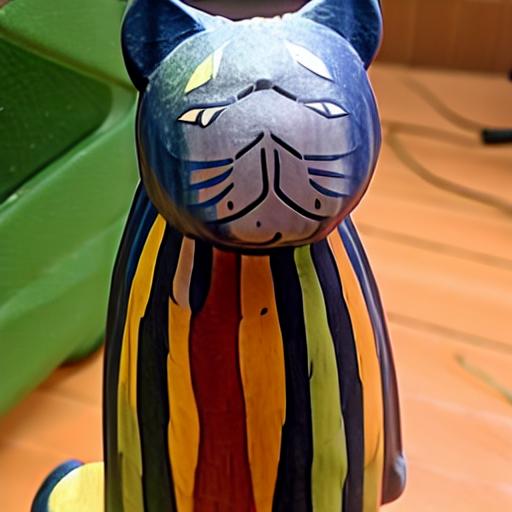} &
        \includegraphics[width=0.11\linewidth]{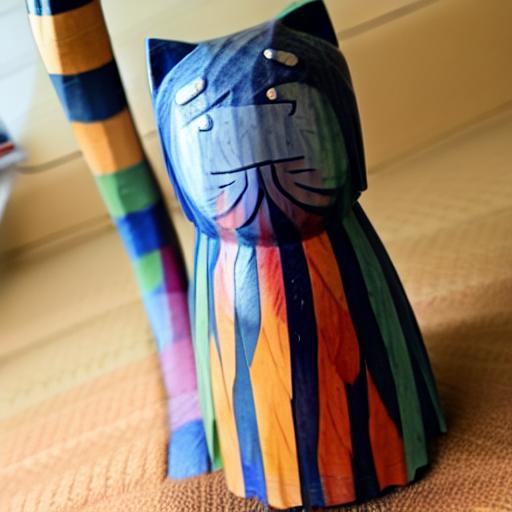} & \includegraphics[width=0.11\linewidth]{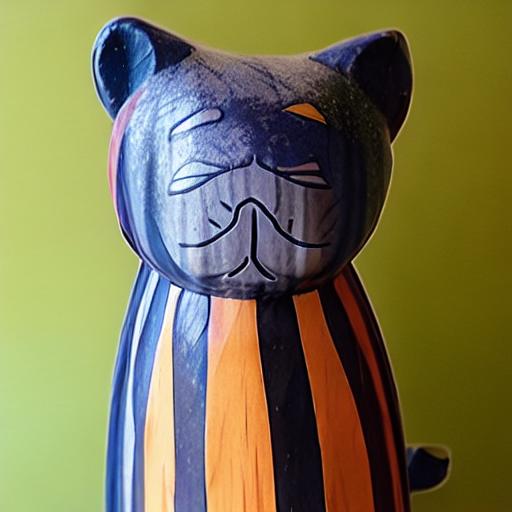} & \includegraphics[width=0.11\linewidth]{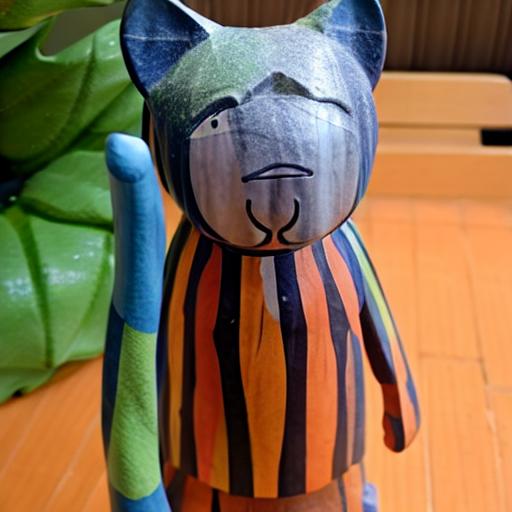} &
        \includegraphics[width=0.11\linewidth]{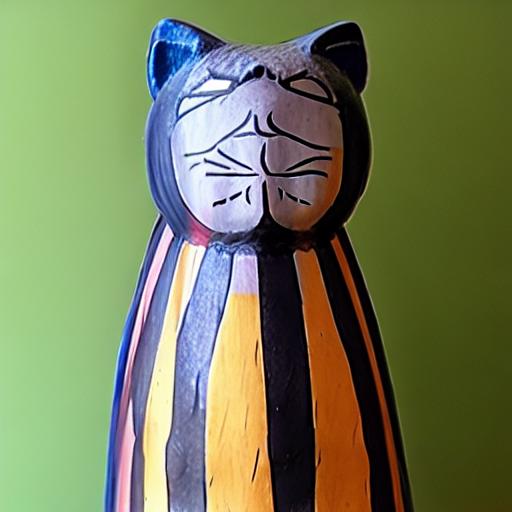} & \includegraphics[width=0.11\linewidth]{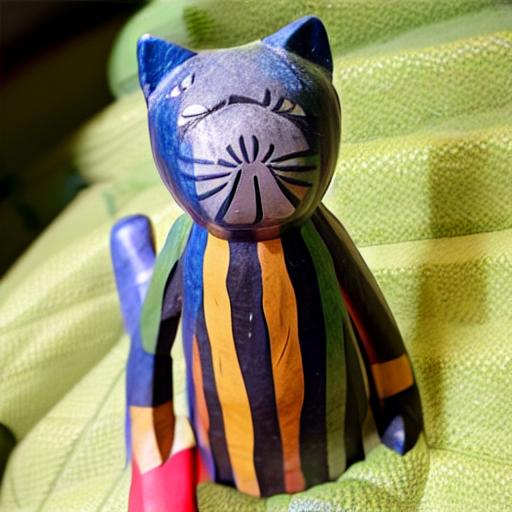} & \includegraphics[width=0.11\linewidth]{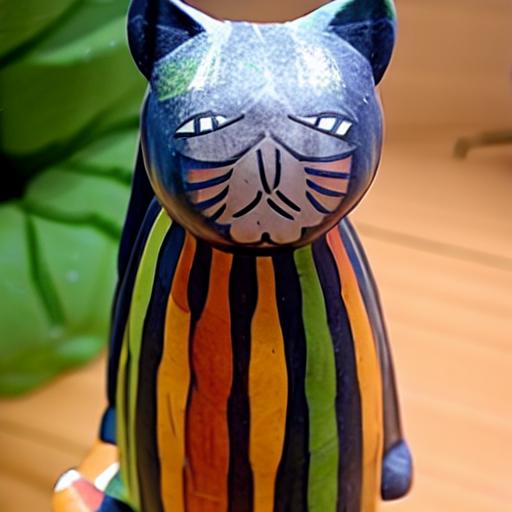}& \includegraphics[width=0.11\linewidth]{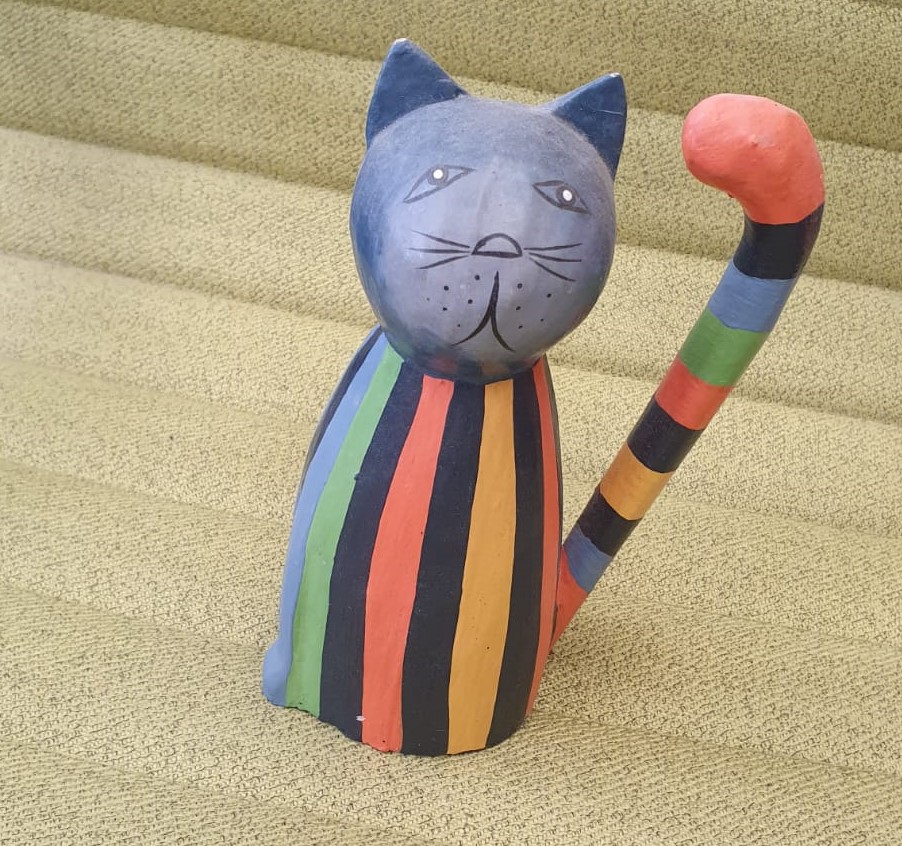} \\
        Ours (CLIP) &
        \includegraphics[width=0.11\linewidth]{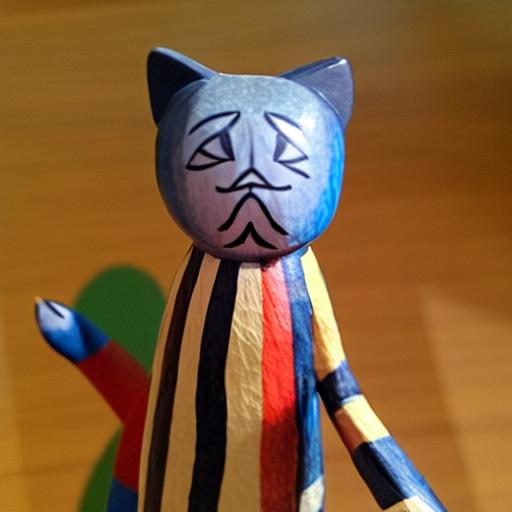} & \includegraphics[width=0.11\linewidth]{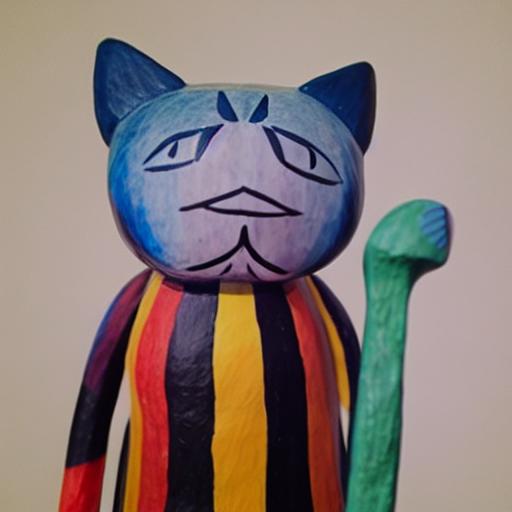} & \includegraphics[width=0.11\linewidth]{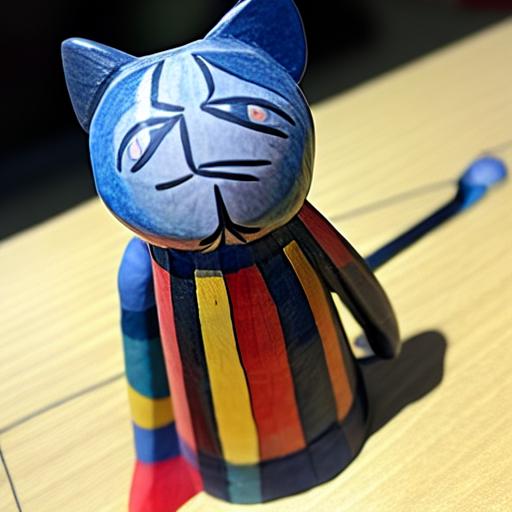} &
        \includegraphics[width=0.11\linewidth]{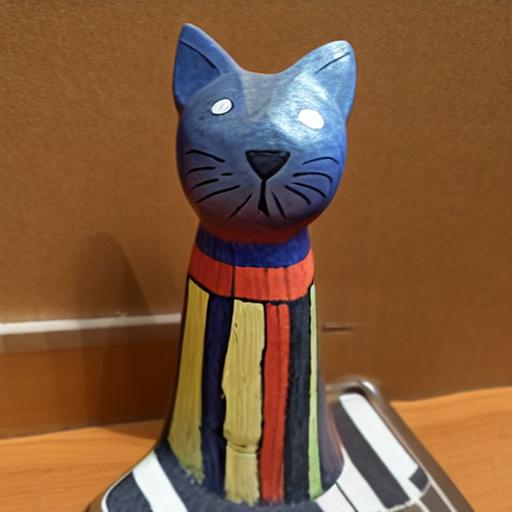} & \includegraphics[width=0.11\linewidth]{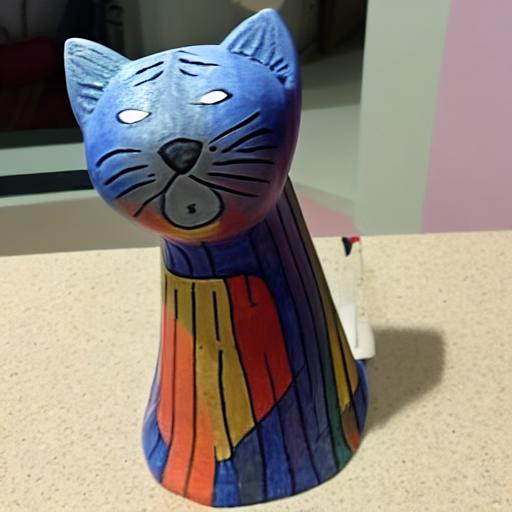} & \includegraphics[width=0.11\linewidth]{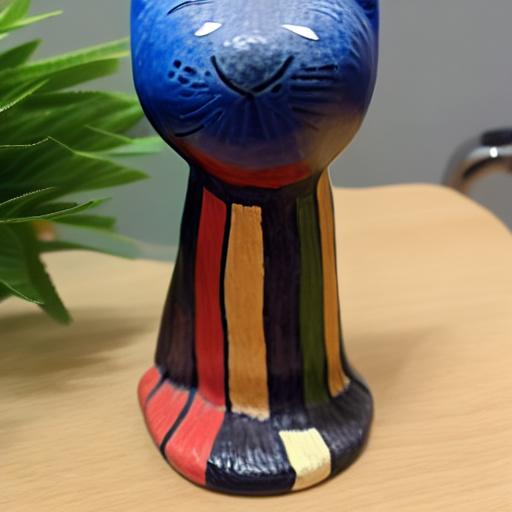} &
        \includegraphics[width=0.11\linewidth]{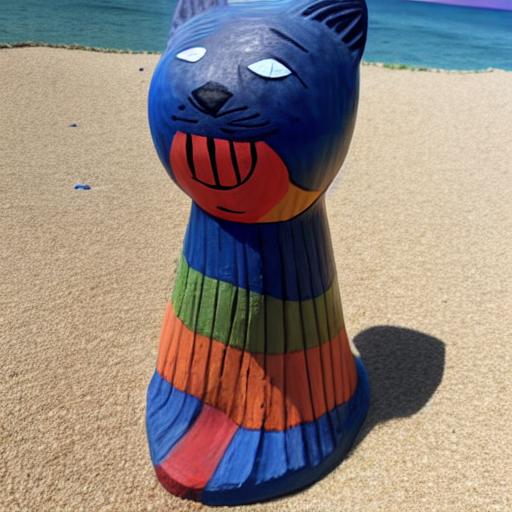} & \includegraphics[width=0.11\linewidth]{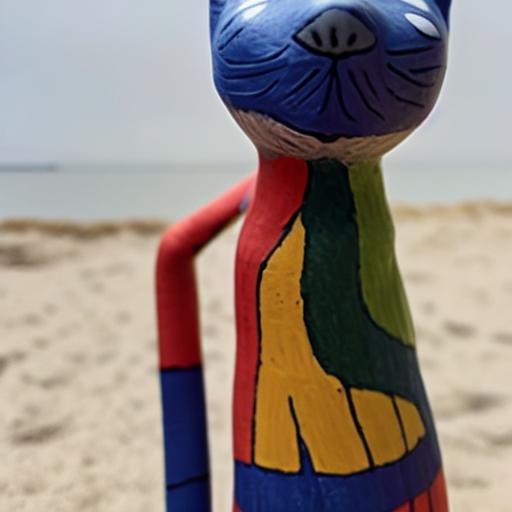} & \includegraphics[width=0.11\linewidth]{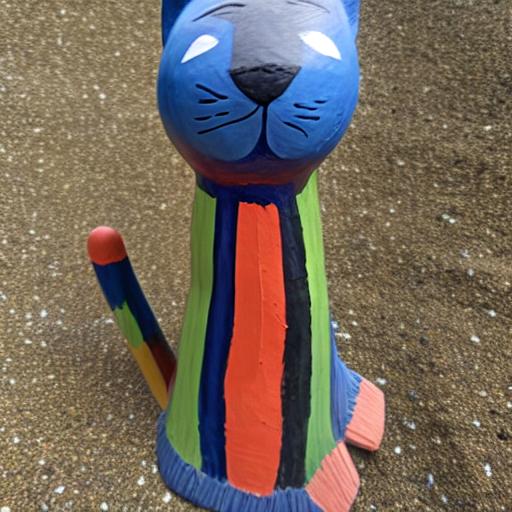} & \includegraphics[width=0.11\linewidth]{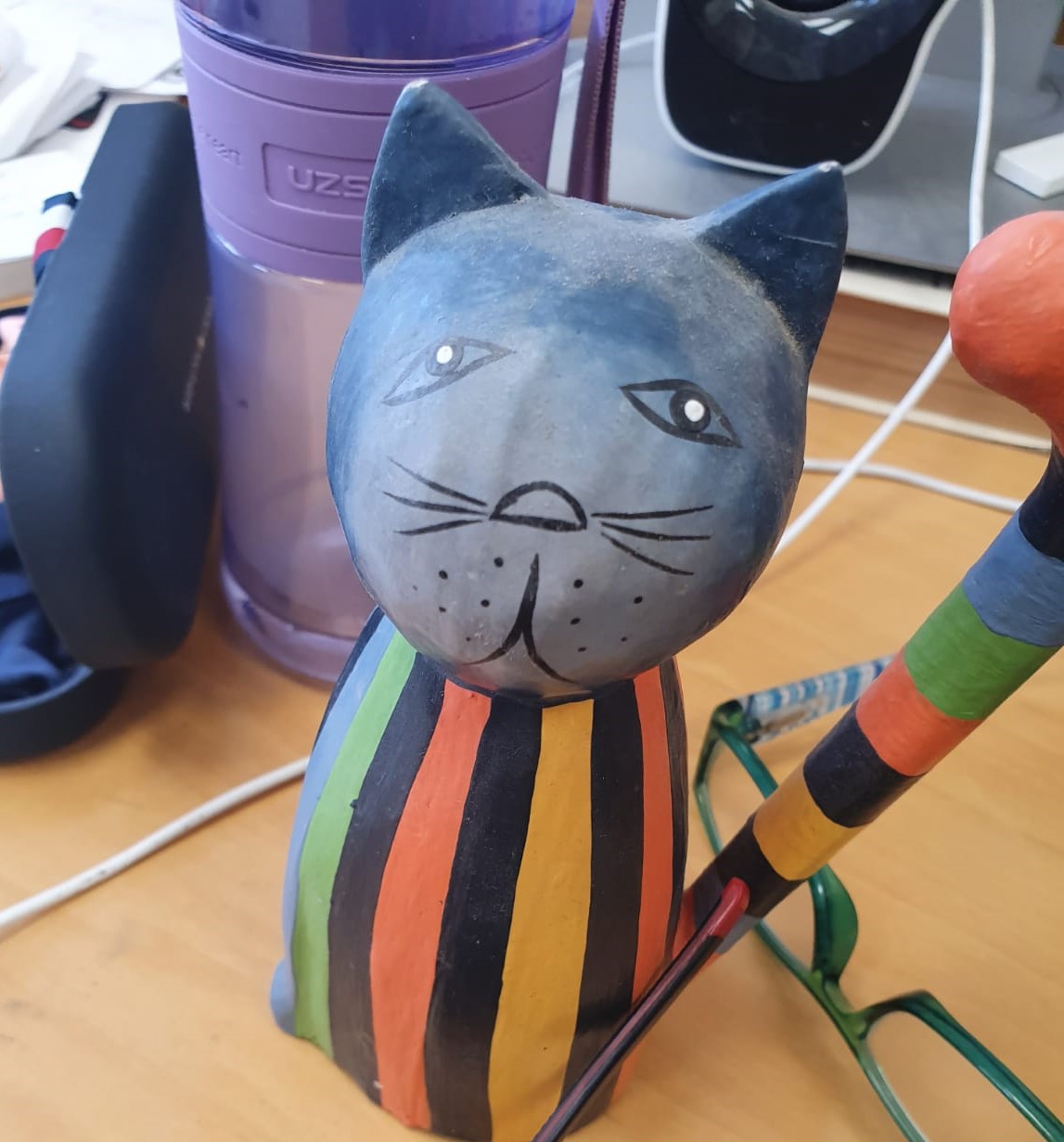} \\
        \hline
        
        TI (BERT) &
        \includegraphics[width=0.11\linewidth]{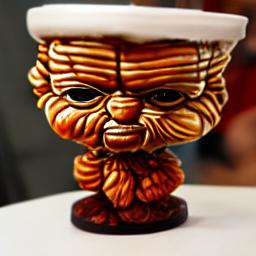} & \includegraphics[width=0.11\linewidth]{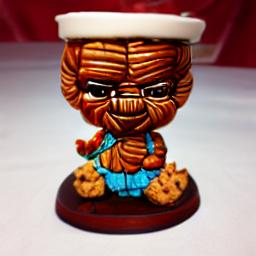} & \includegraphics[width=0.11\linewidth]{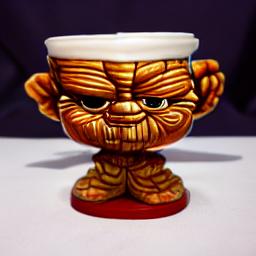} &
        \includegraphics[width=0.11\linewidth]{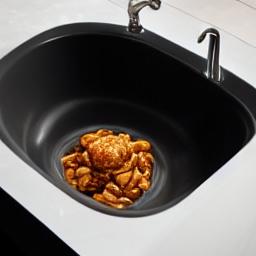} & \includegraphics[width=0.11\linewidth]{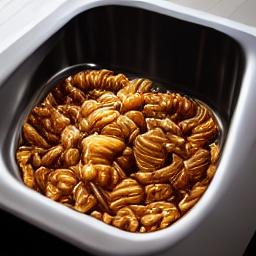} & \includegraphics[width=0.11\linewidth]{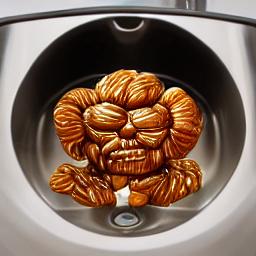} &
        \includegraphics[width=0.11\linewidth]{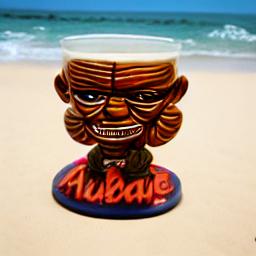} & \includegraphics[width=0.11\linewidth]{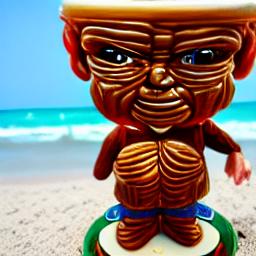} & \includegraphics[width=0.11\linewidth]{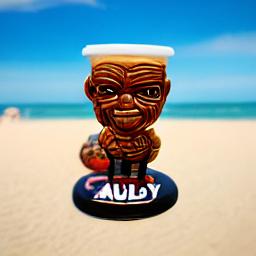}& \includegraphics[width=0.11\linewidth]{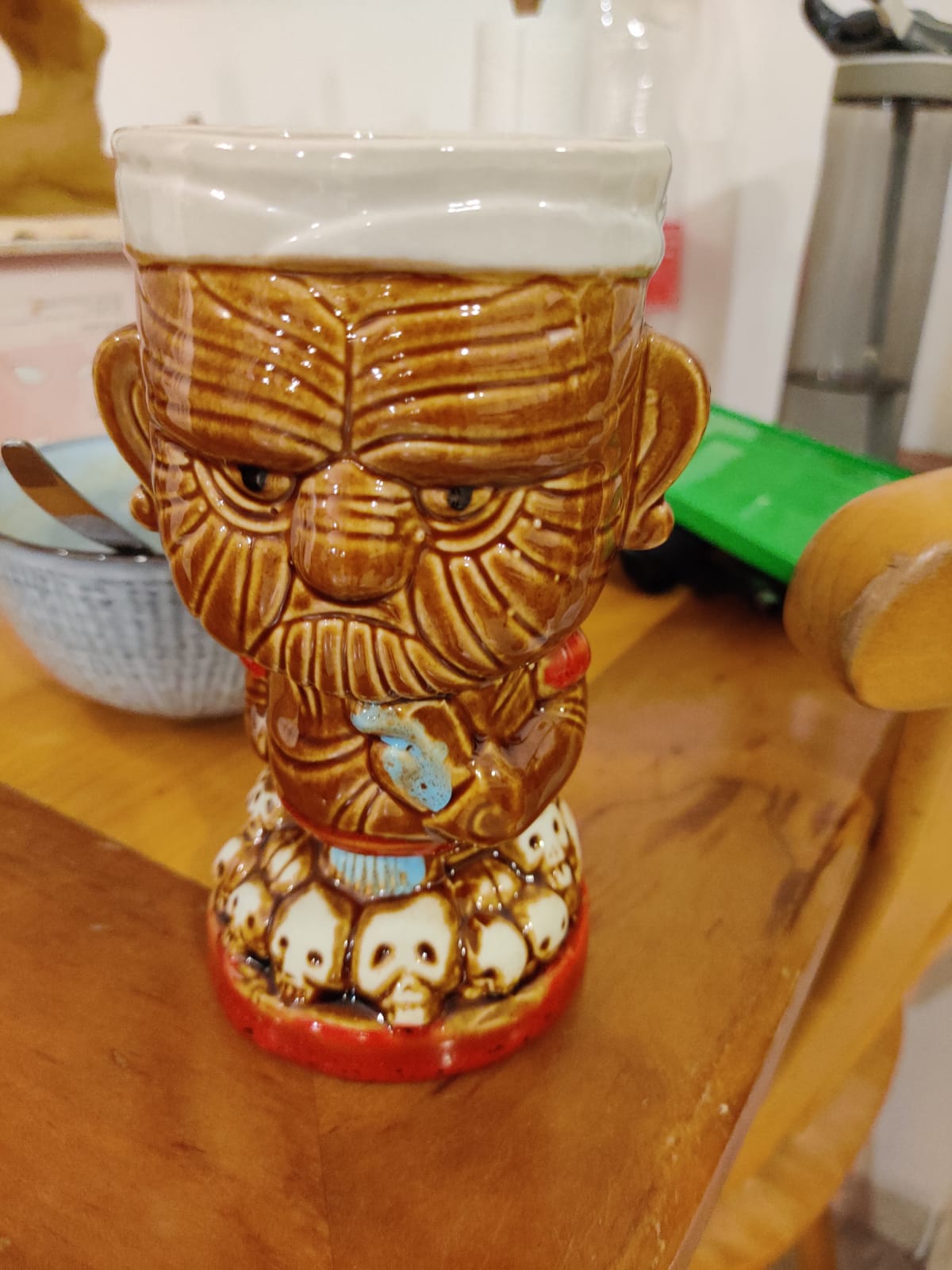} \\
        TI (CLIP) &
        \includegraphics[width=0.11\linewidth]{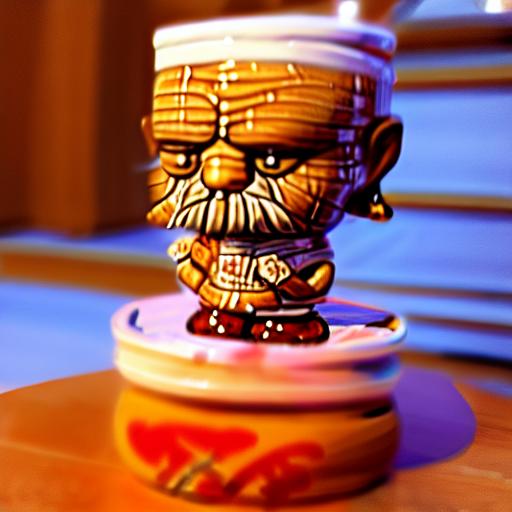} & \includegraphics[width=0.11\linewidth]{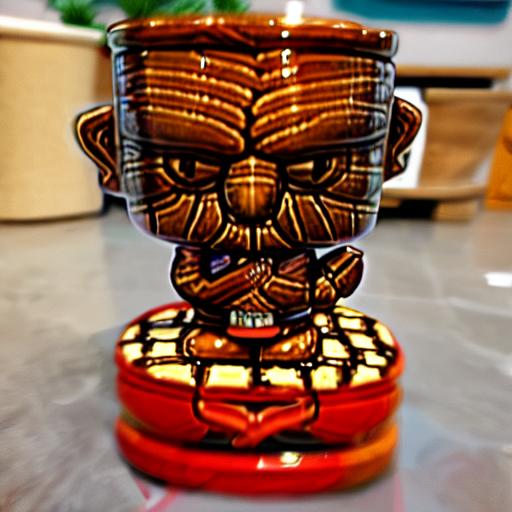} & \includegraphics[width=0.11\linewidth]{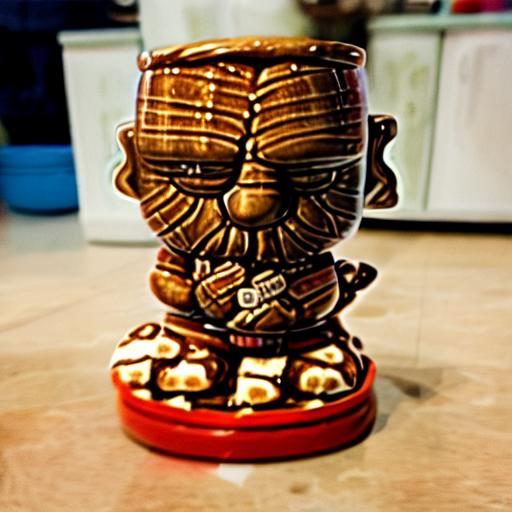} &
        \includegraphics[width=0.11\linewidth]{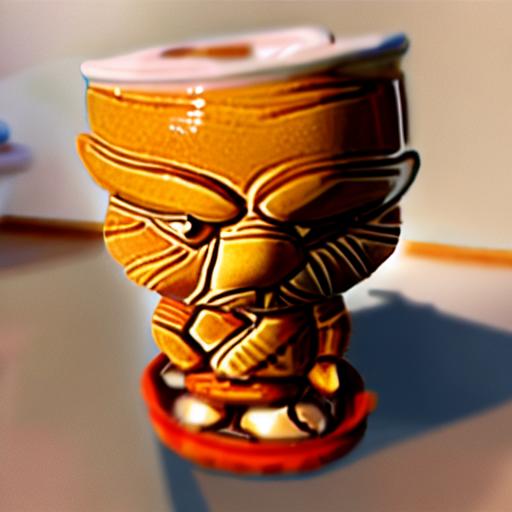} & \includegraphics[width=0.11\linewidth]{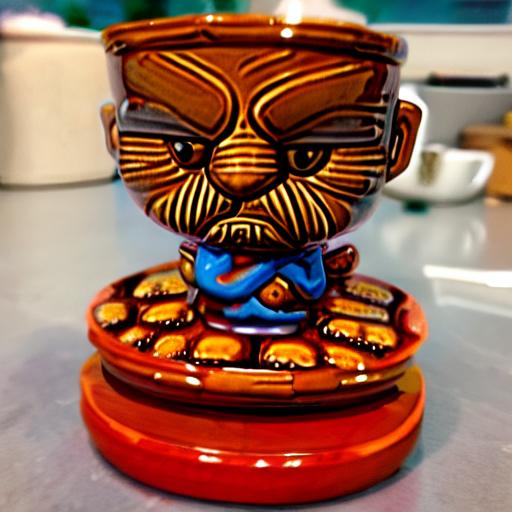} & \includegraphics[width=0.11\linewidth]{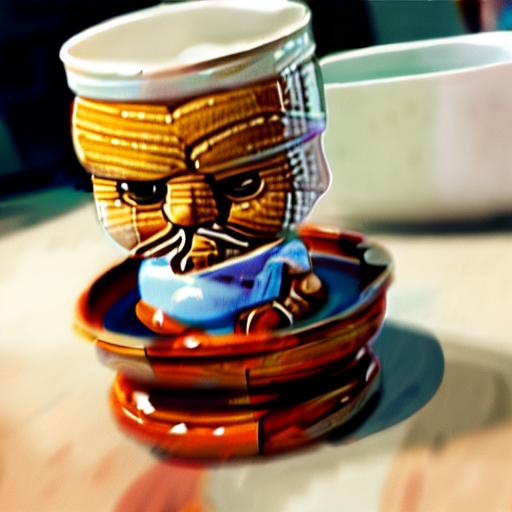} &
        \includegraphics[width=0.11\linewidth]{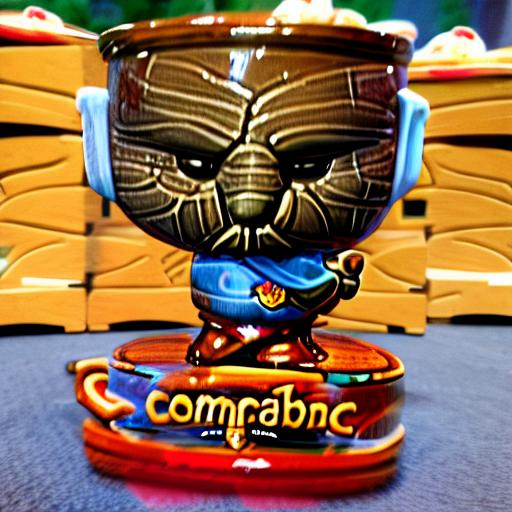} & \includegraphics[width=0.11\linewidth]{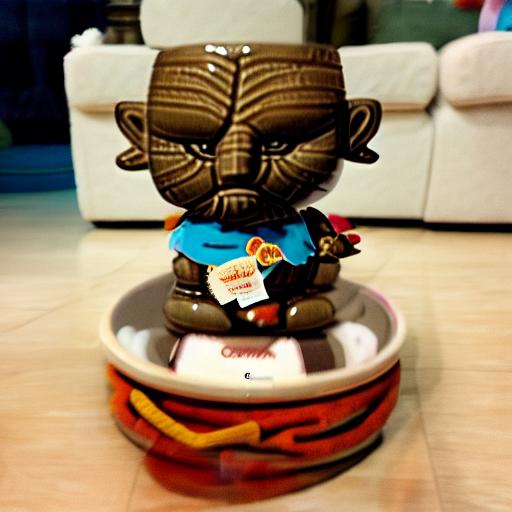} & \includegraphics[width=0.11\linewidth]{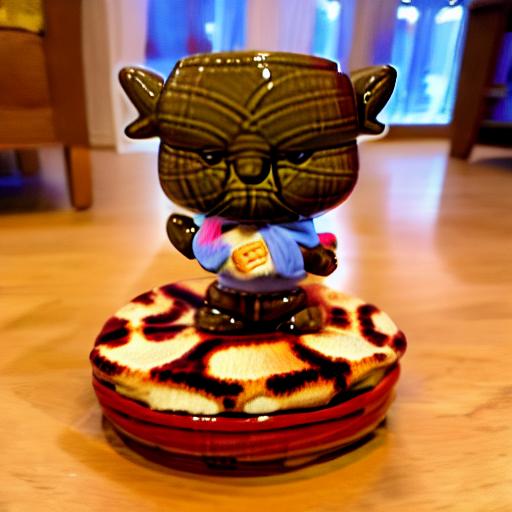}& \includegraphics[width=0.11\linewidth]{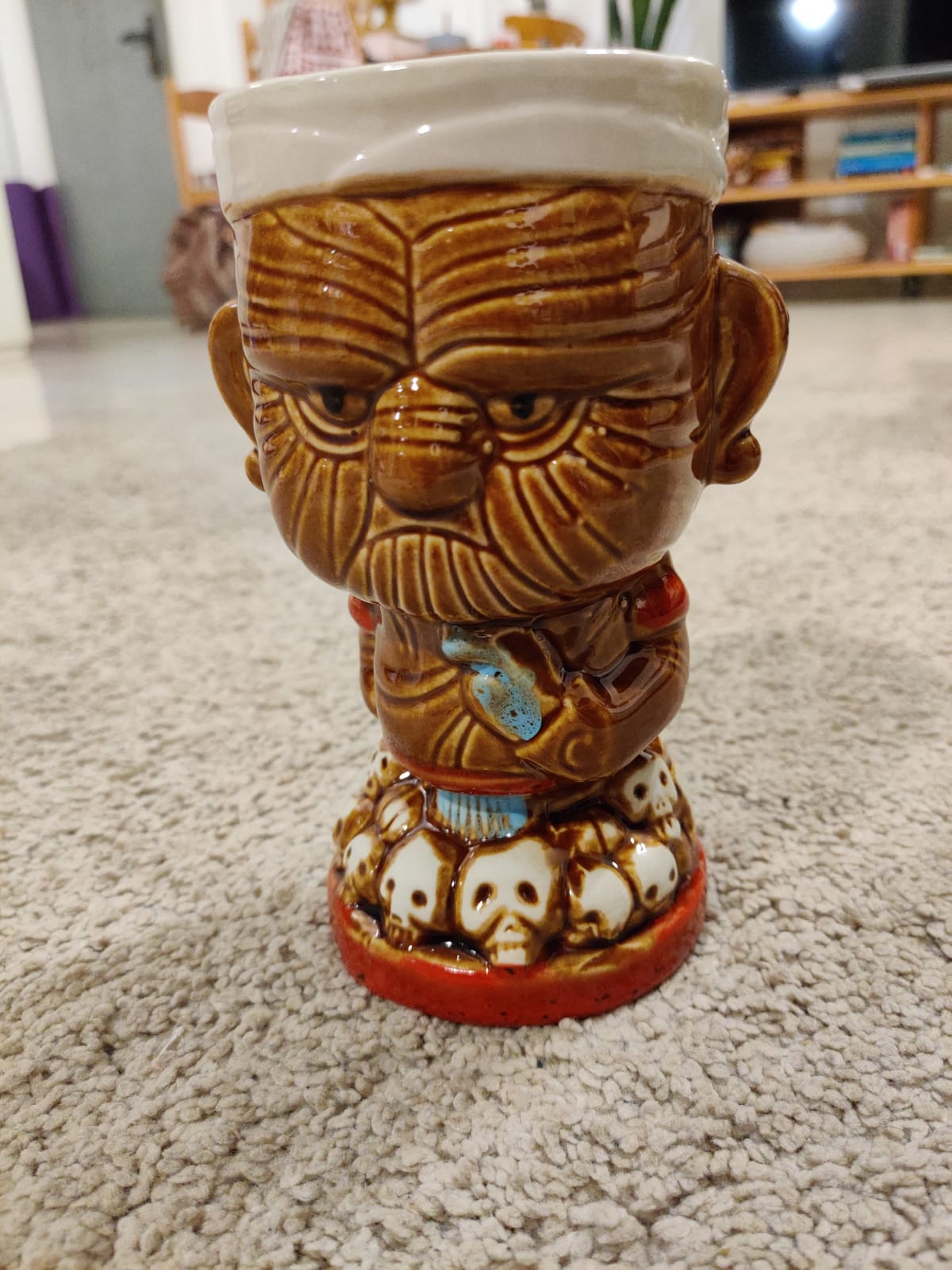} \\
        Ours (CLIP) &
        \includegraphics[width=0.11\linewidth]{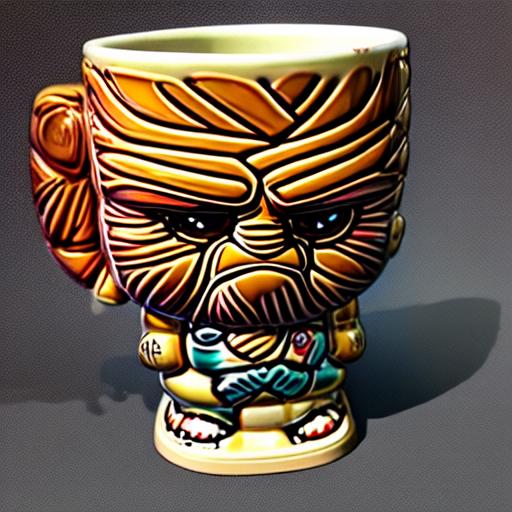} & \includegraphics[width=0.11\linewidth]{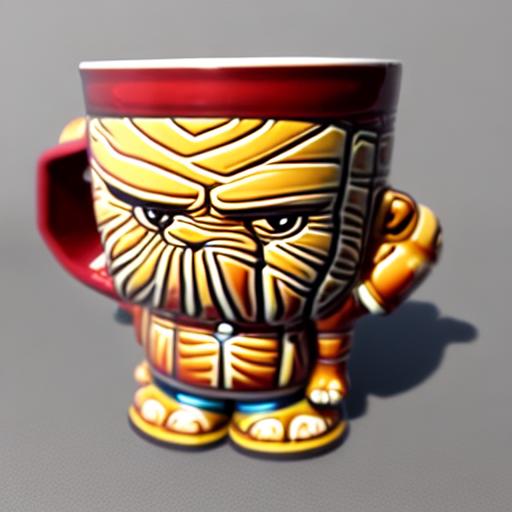} & \includegraphics[width=0.11\linewidth]{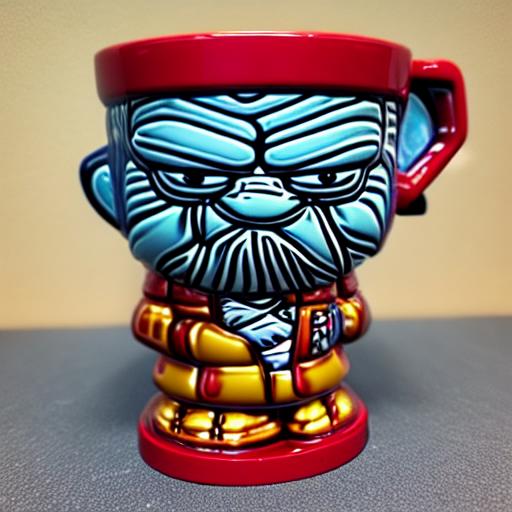} &
        \includegraphics[width=0.11\linewidth]{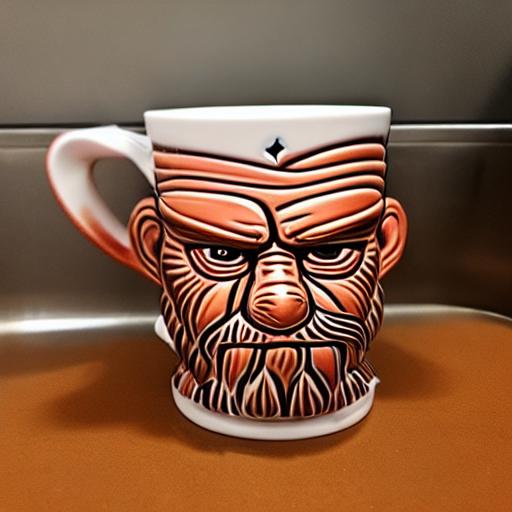} & \includegraphics[width=0.11\linewidth]{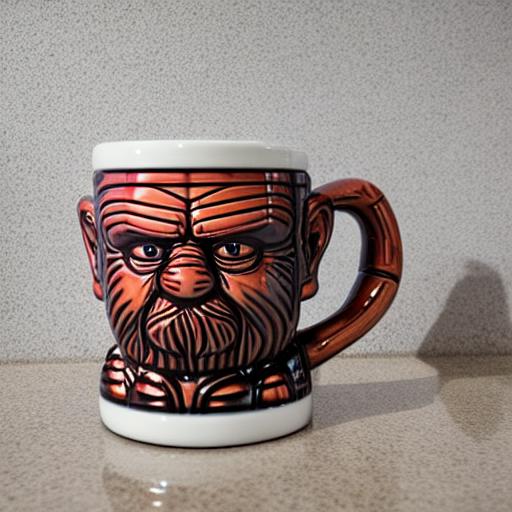} & \includegraphics[width=0.11\linewidth]{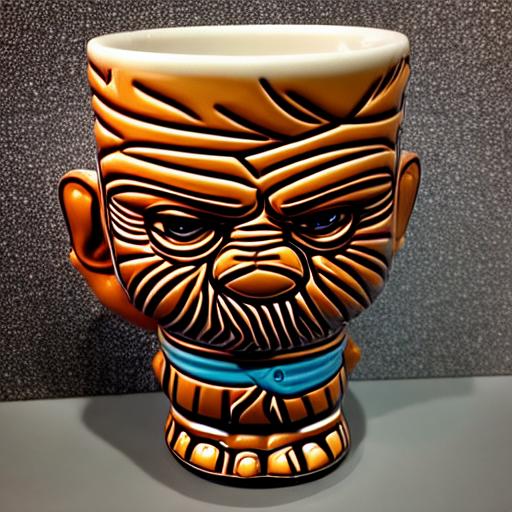} &
        \includegraphics[width=0.11\linewidth]{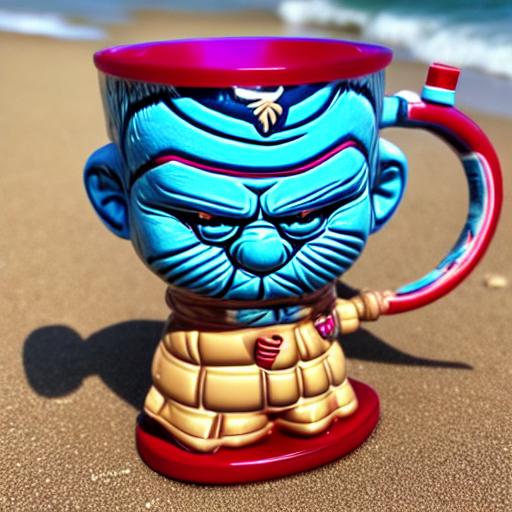} & \includegraphics[width=0.11\linewidth]{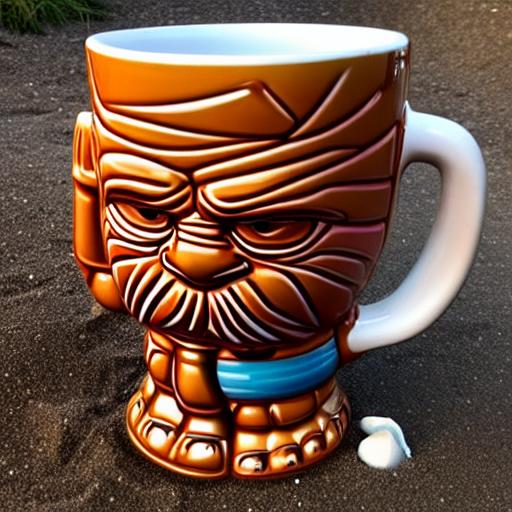} & \includegraphics[width=0.11\linewidth]{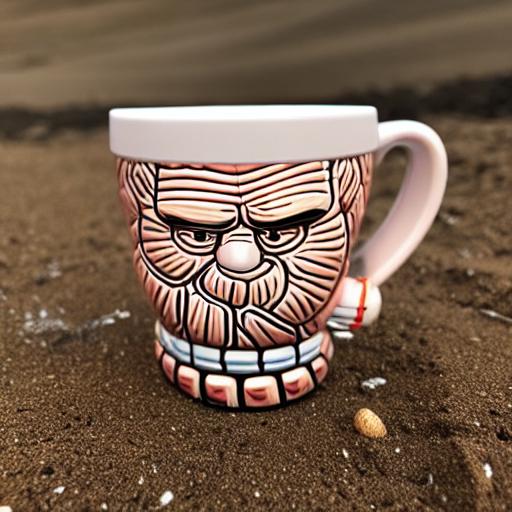}& \includegraphics[width=0.11\linewidth]{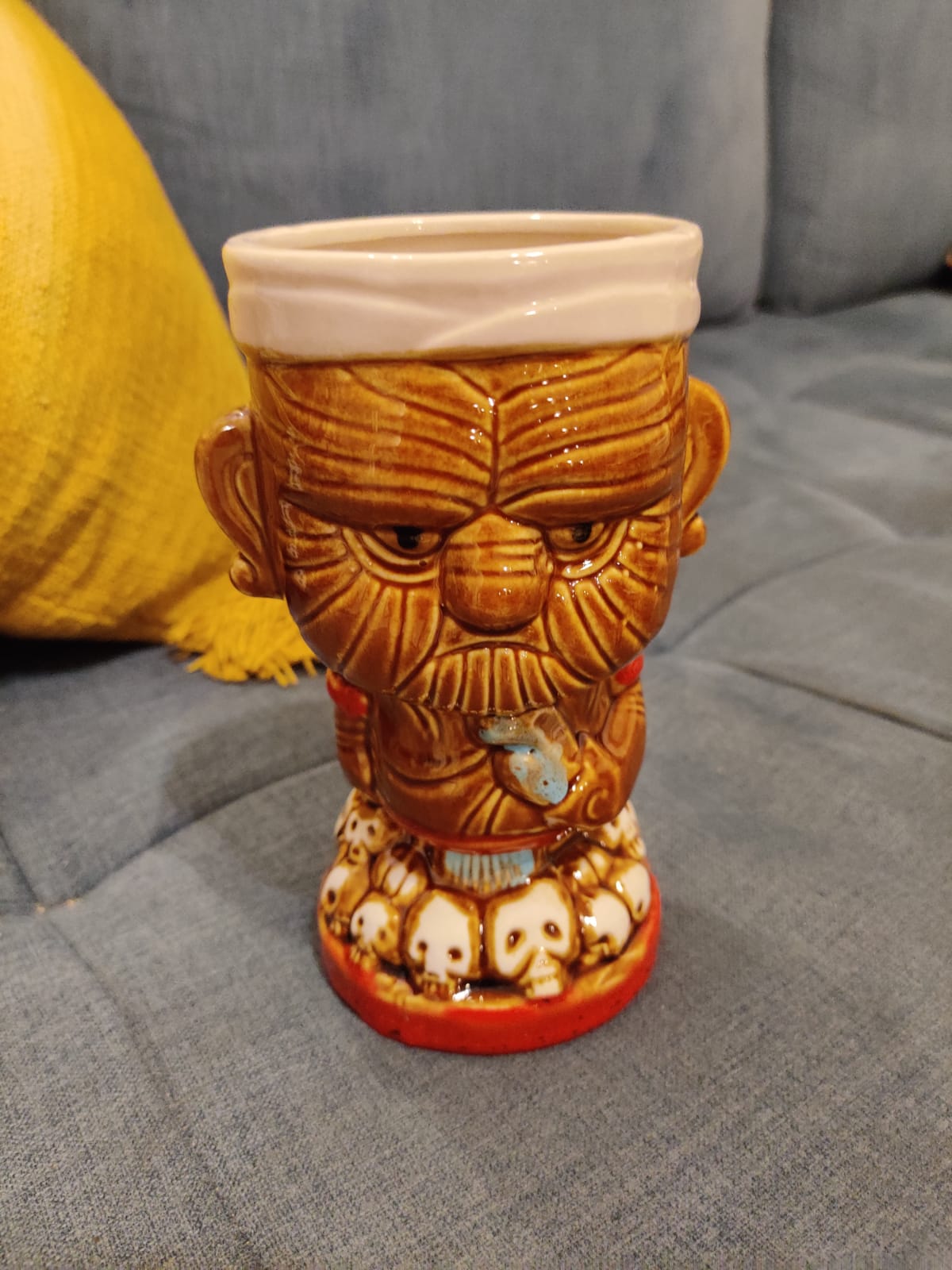} \\
        \hline
        
        TI (BERT) &
        \includegraphics[width=0.11\linewidth]{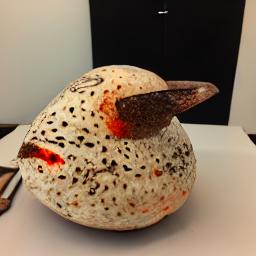} & \includegraphics[width=0.11\linewidth]{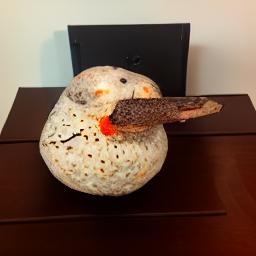} & \includegraphics[width=0.11\linewidth]{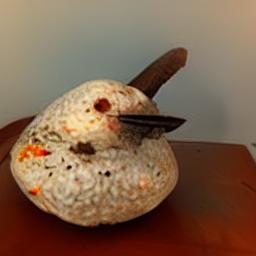} &
        \includegraphics[width=0.11\linewidth]{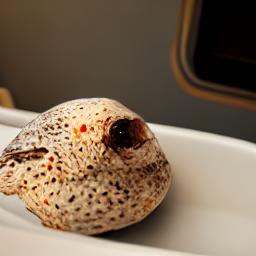} & \includegraphics[width=0.11\linewidth]{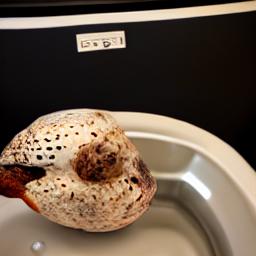} & \includegraphics[width=0.11\linewidth]{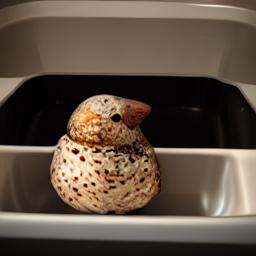} &
        \includegraphics[width=0.11\linewidth]{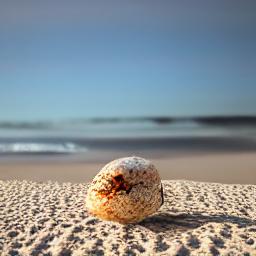} & \includegraphics[width=0.11\linewidth]{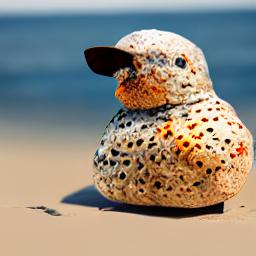} & \includegraphics[width=0.11\linewidth]{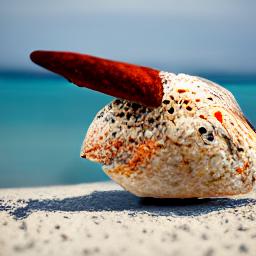}& \includegraphics[width=0.11\linewidth]{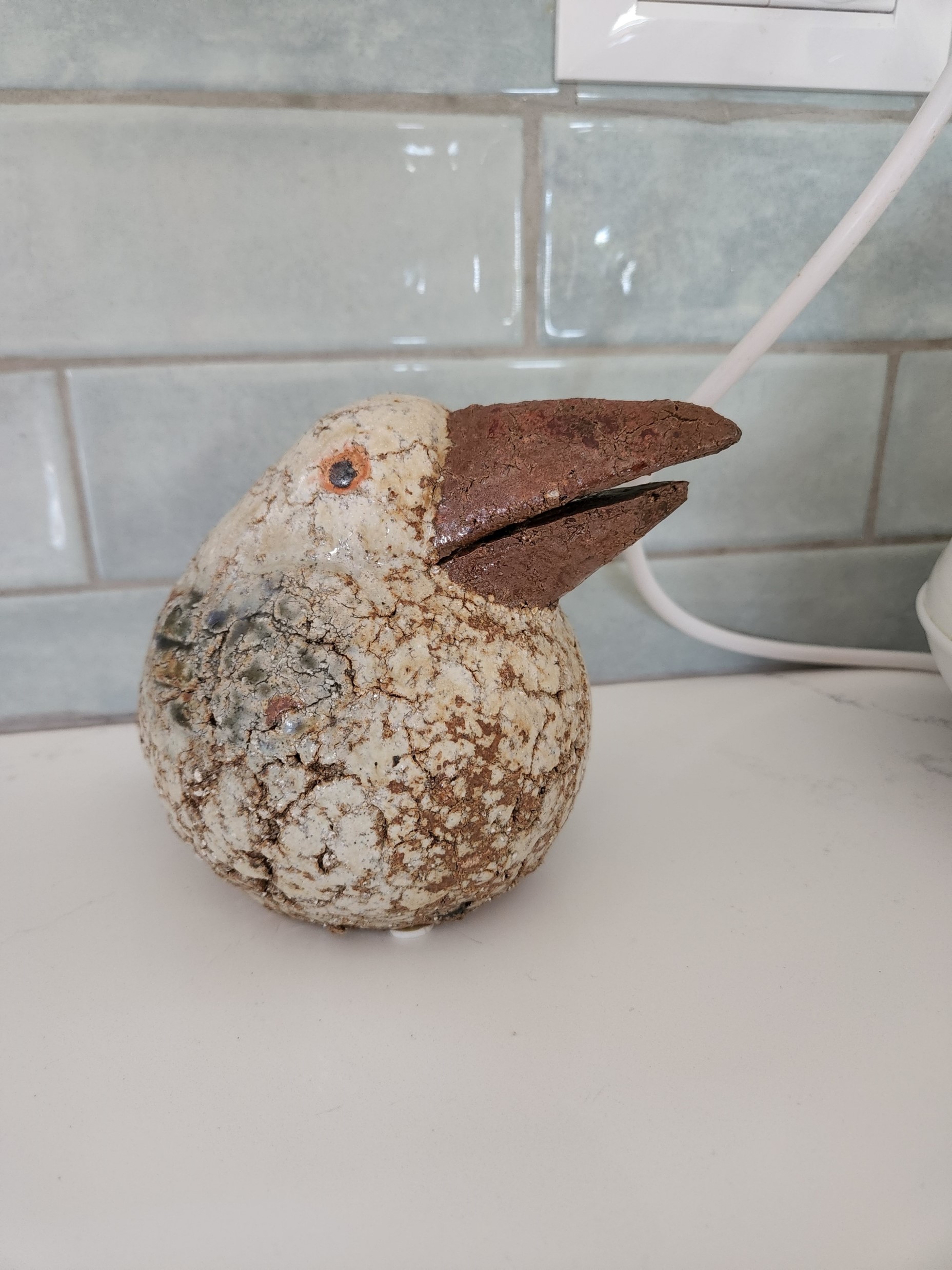} \\
        TI (CLIP) &
        \includegraphics[width=0.11\linewidth]{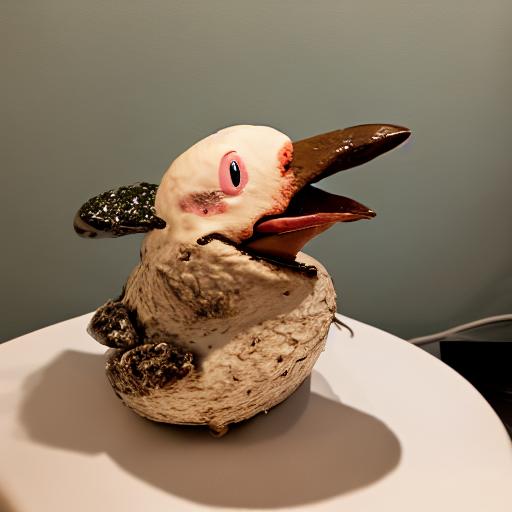} & \includegraphics[width=0.11\linewidth]{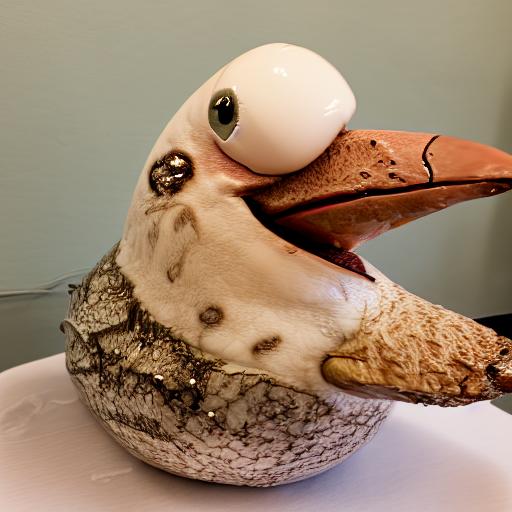} & \includegraphics[width=0.11\linewidth]{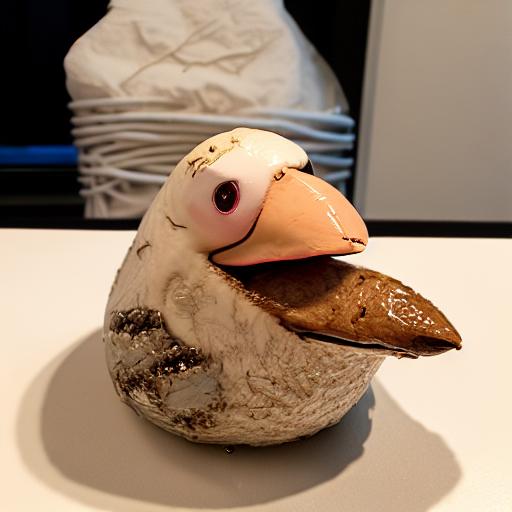} &
        \includegraphics[width=0.11\linewidth]{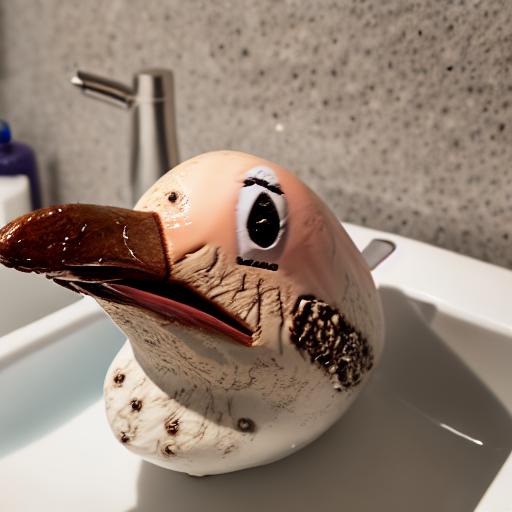} & \includegraphics[width=0.11\linewidth]{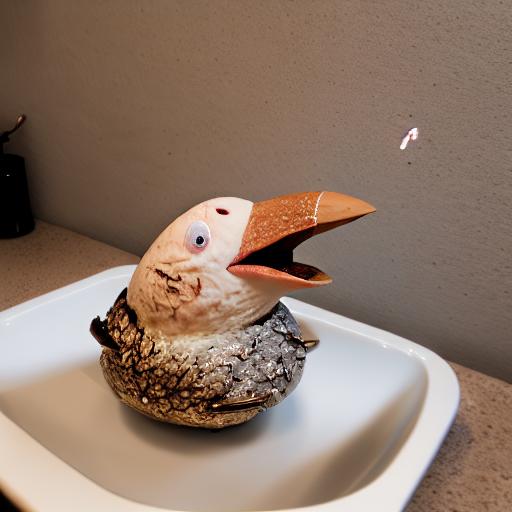} & \includegraphics[width=0.11\linewidth]{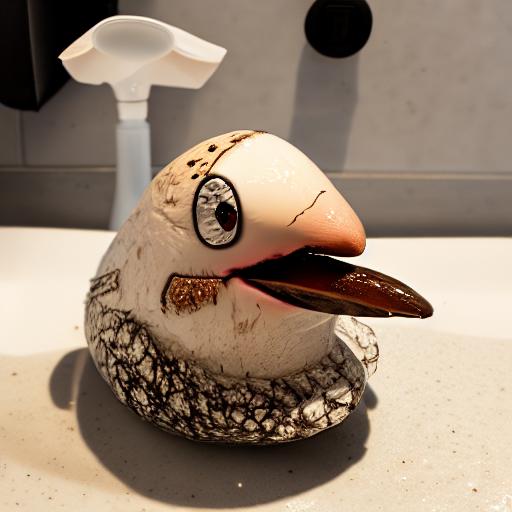} &
        \includegraphics[width=0.11\linewidth]{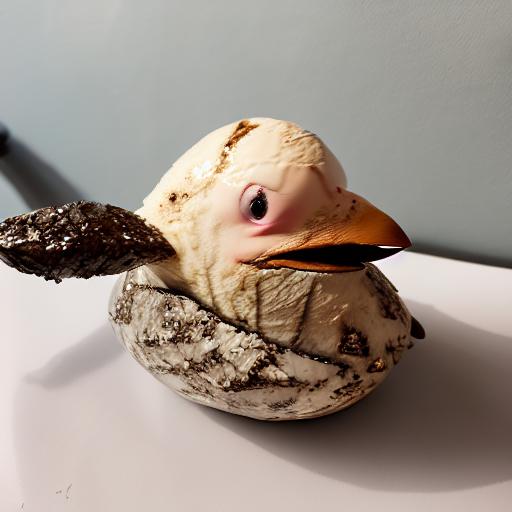} & \includegraphics[width=0.11\linewidth]{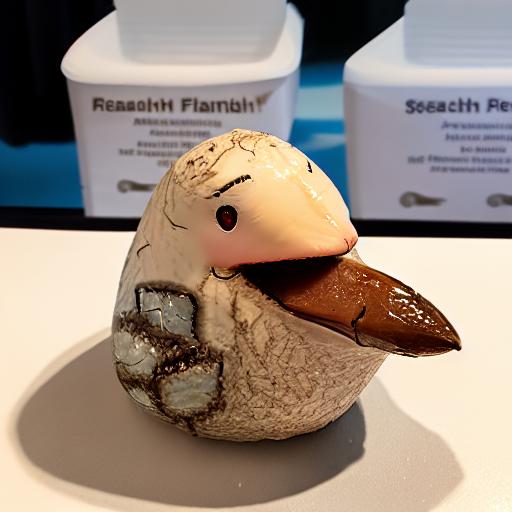} & \includegraphics[width=0.11\linewidth]{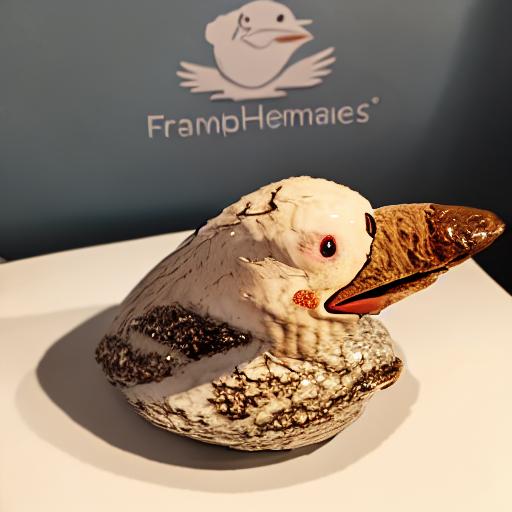}& \includegraphics[width=0.11\linewidth]{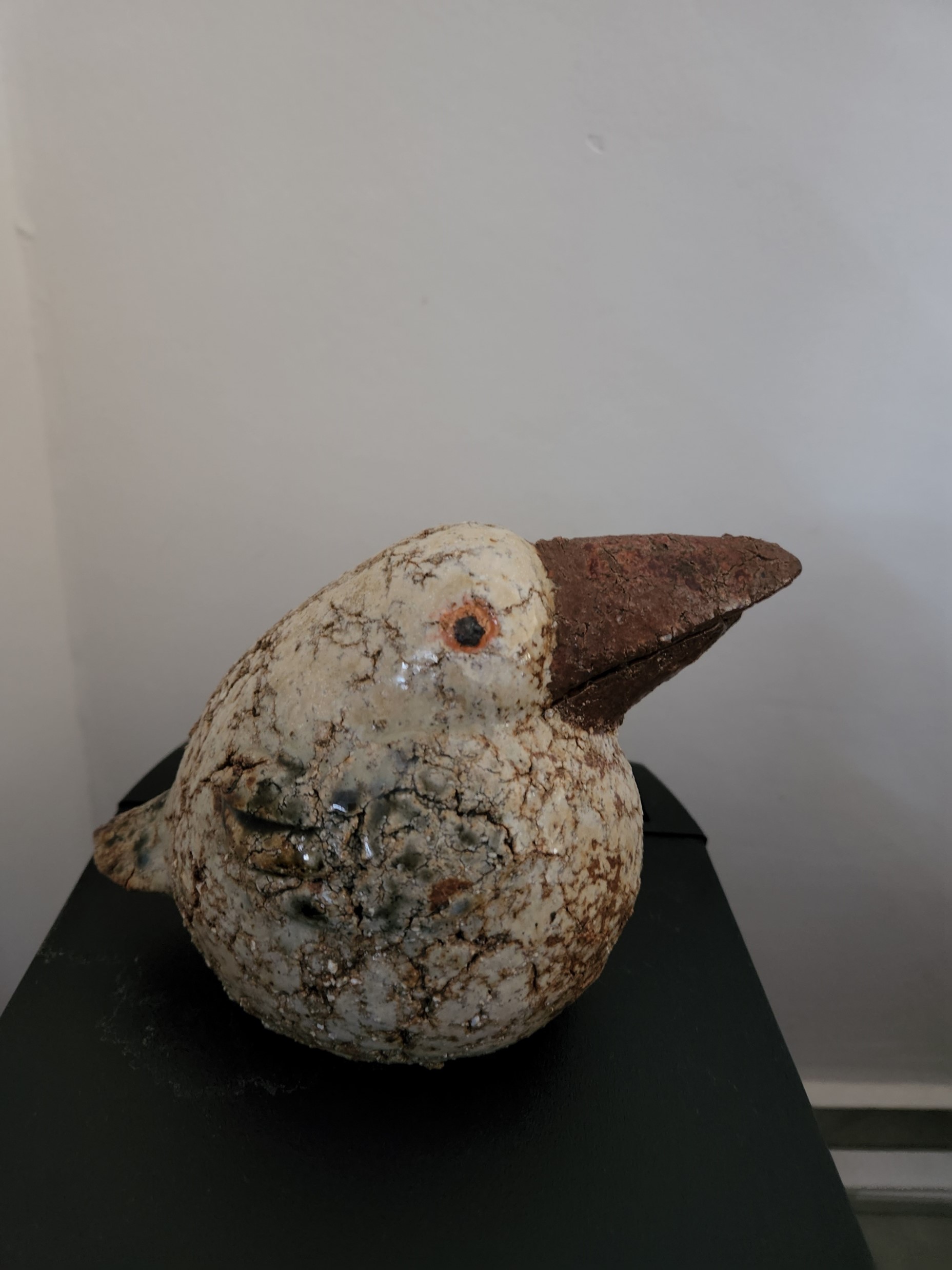} \\
        Ours (CLIP) &
        \includegraphics[width=0.11\linewidth]{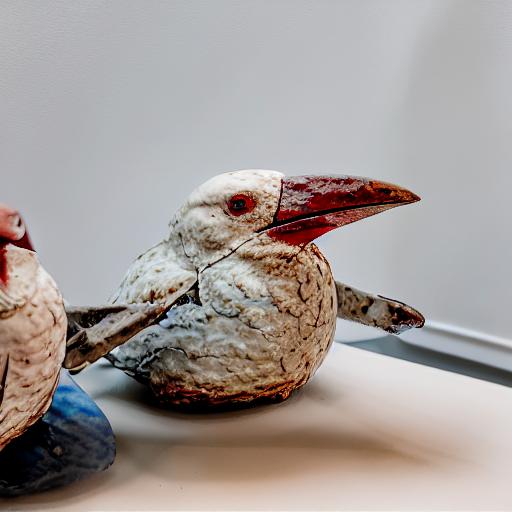} & \includegraphics[width=0.11\linewidth]{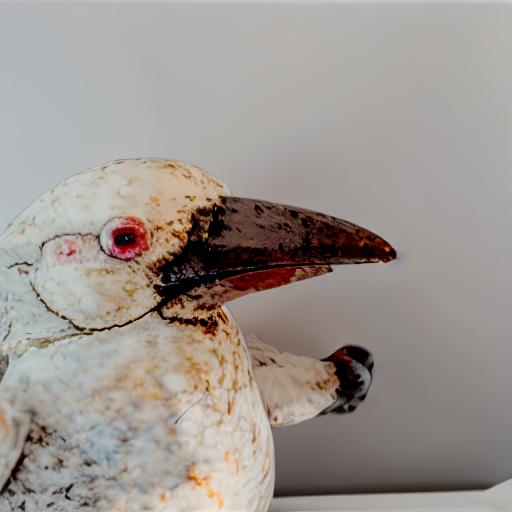} & \includegraphics[width=0.11\linewidth]{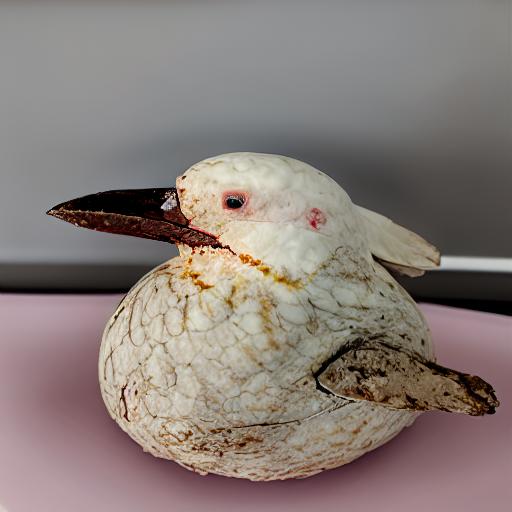} &
        \includegraphics[width=0.11\linewidth]{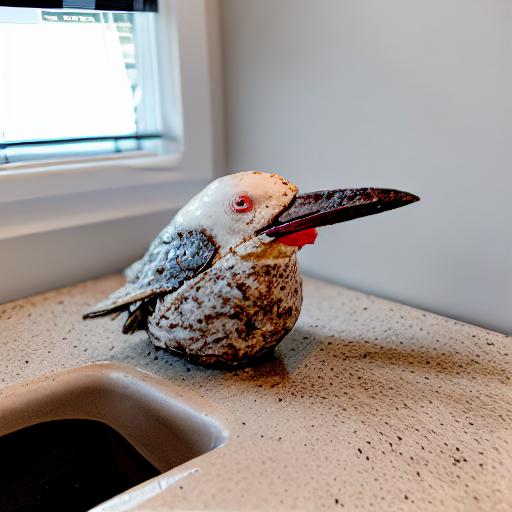} & \includegraphics[width=0.11\linewidth]{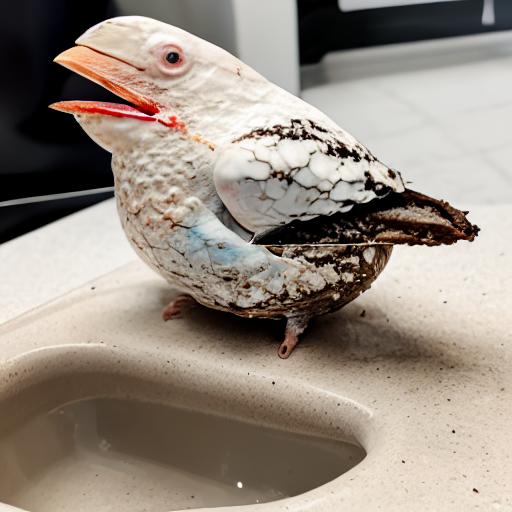} & \includegraphics[width=0.11\linewidth]{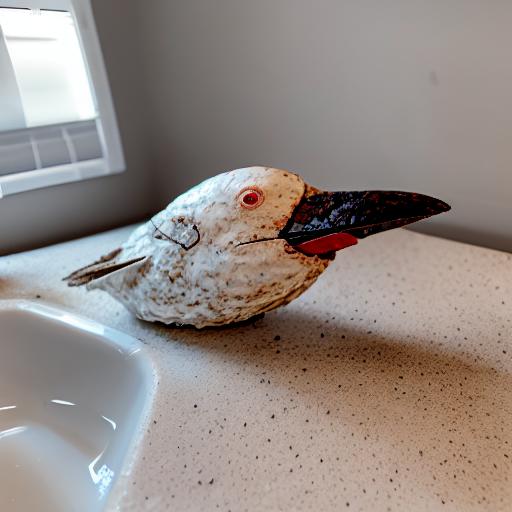} &
        \includegraphics[width=0.11\linewidth]{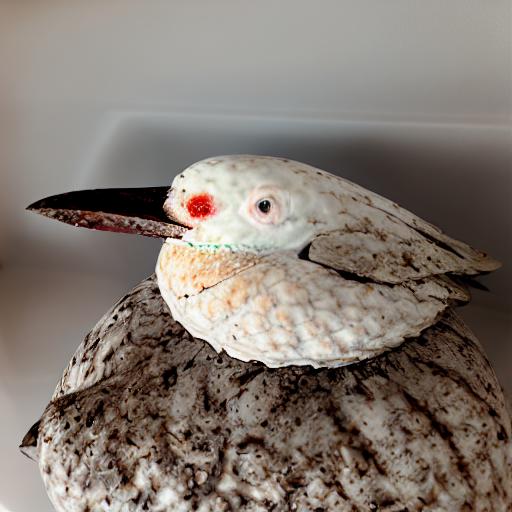} & \includegraphics[width=0.11\linewidth]{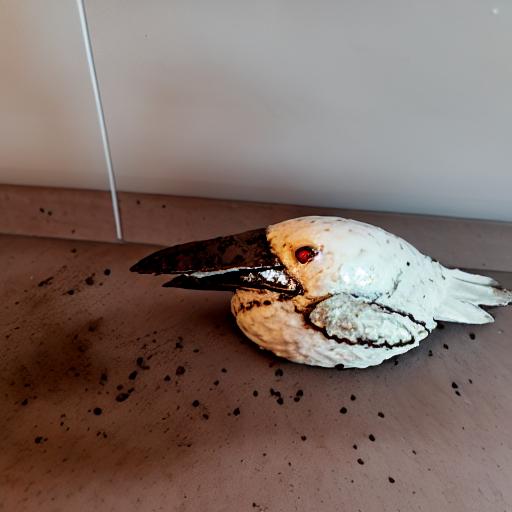} & \includegraphics[width=0.11\linewidth]{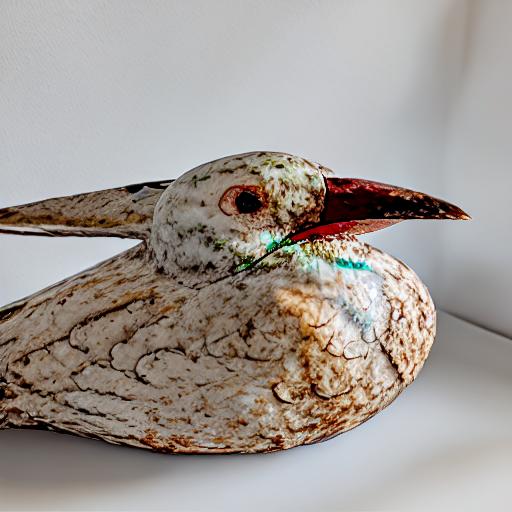}& \includegraphics[width=0.11\linewidth]{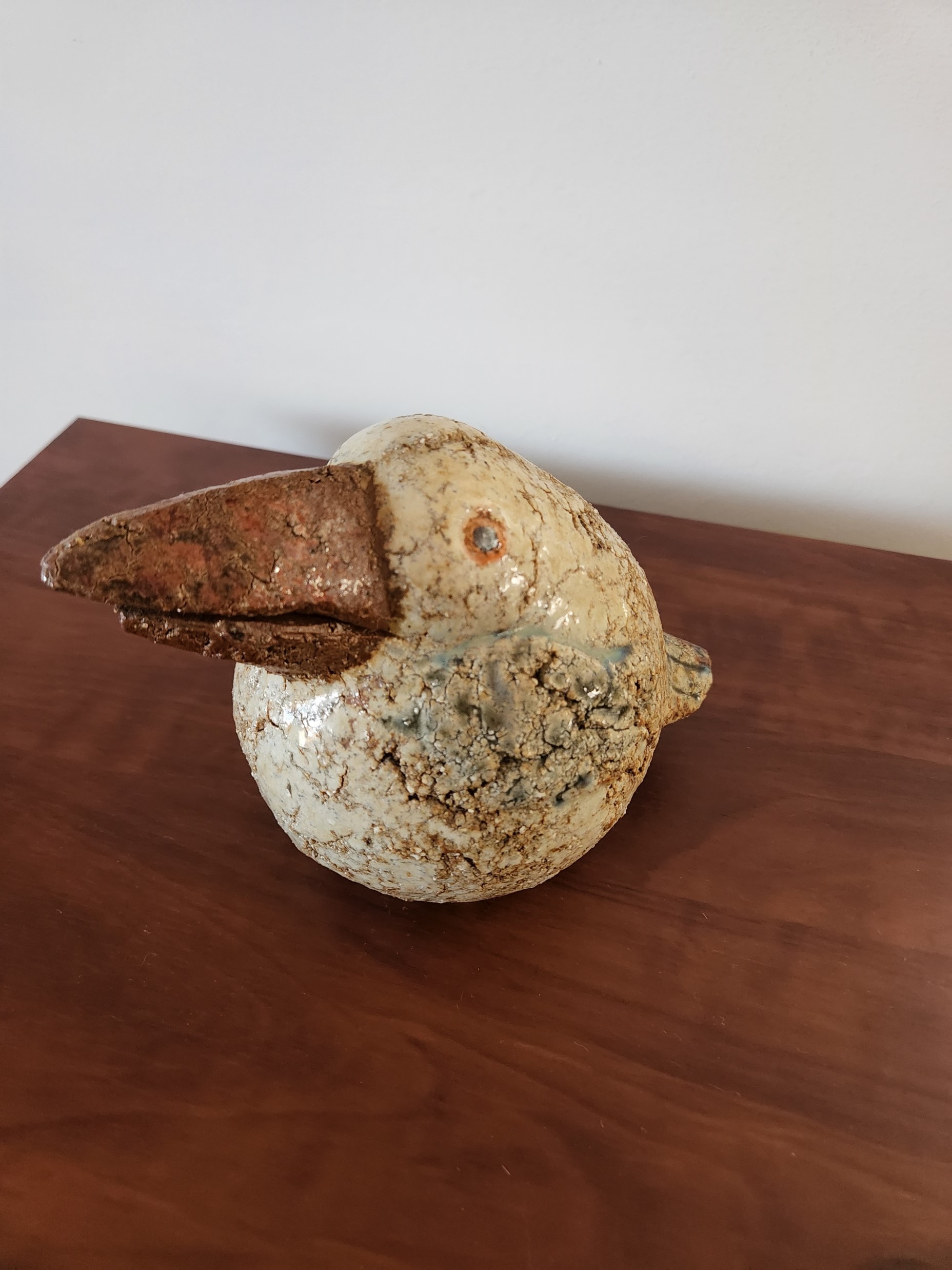} \\
        \hline
        TI (BERT) &
        \includegraphics[width=0.11\linewidth]{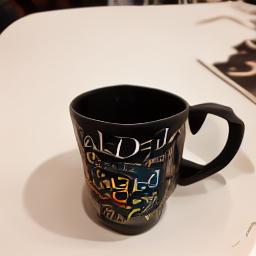} & \includegraphics[width=0.11\linewidth]{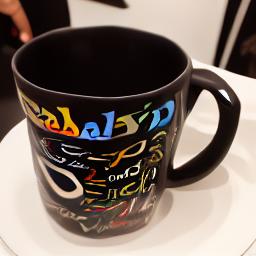} & \includegraphics[width=0.11\linewidth]{images/qual_samples/p_ld_0.jpg} &
        \includegraphics[width=0.11\linewidth]{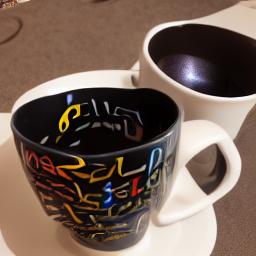} & \includegraphics[width=0.11\linewidth]{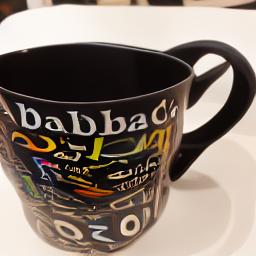} & \includegraphics[width=0.11\linewidth]{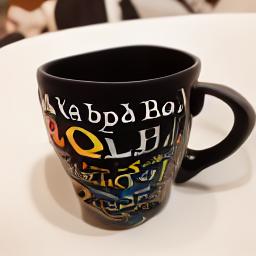} &
        \includegraphics[width=0.11\linewidth]{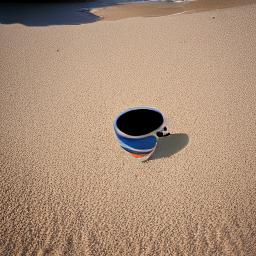} & \includegraphics[width=0.11\linewidth]{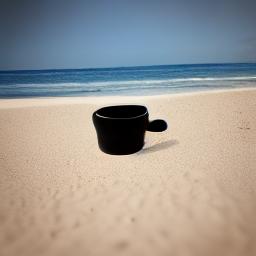} & \includegraphics[width=0.11\linewidth]{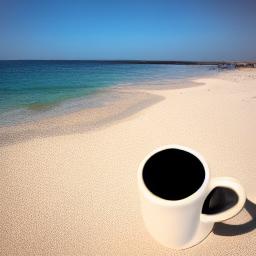}& \includegraphics[width=0.11\linewidth]{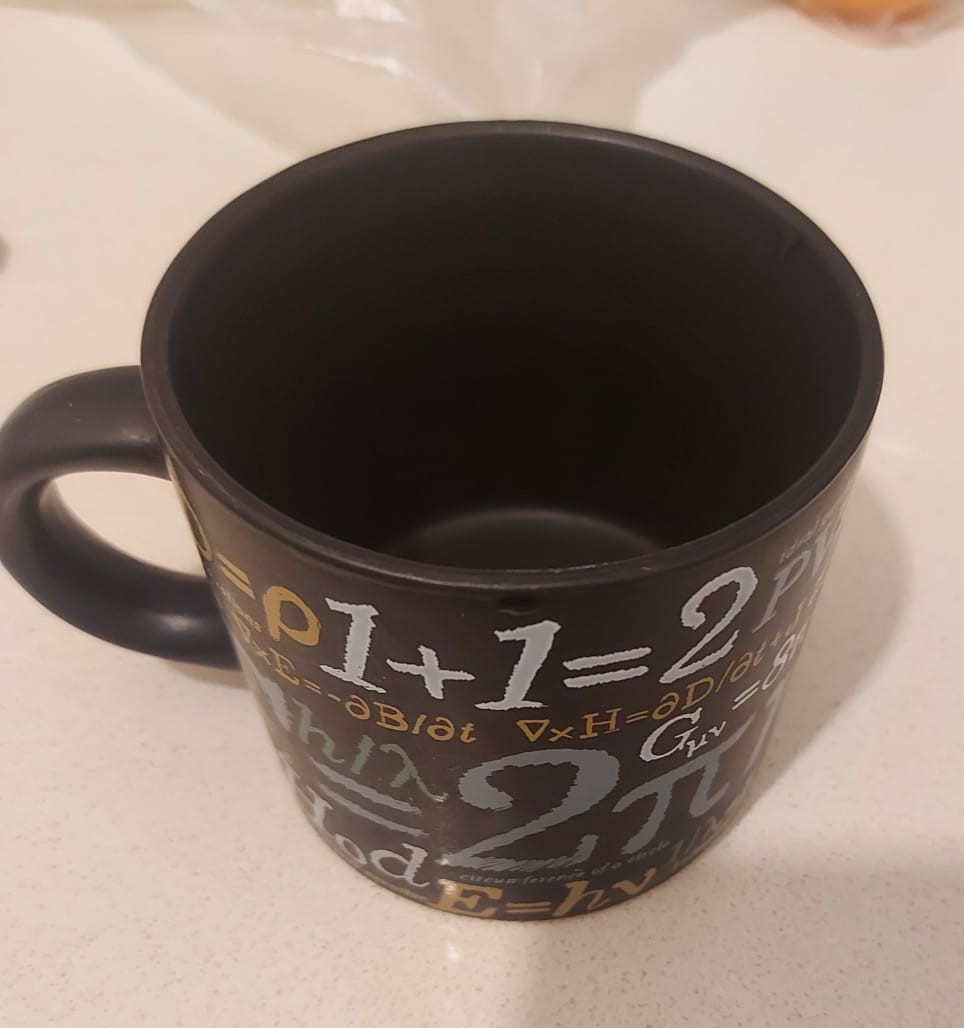} \\
        TI (CLIP) &
        \includegraphics[width=0.11\linewidth]{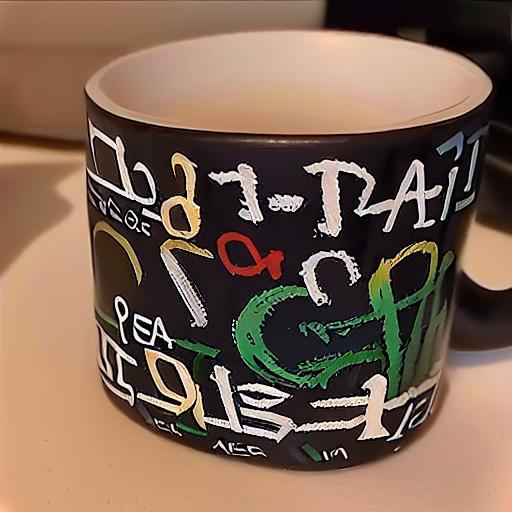} & \includegraphics[width=0.11\linewidth]{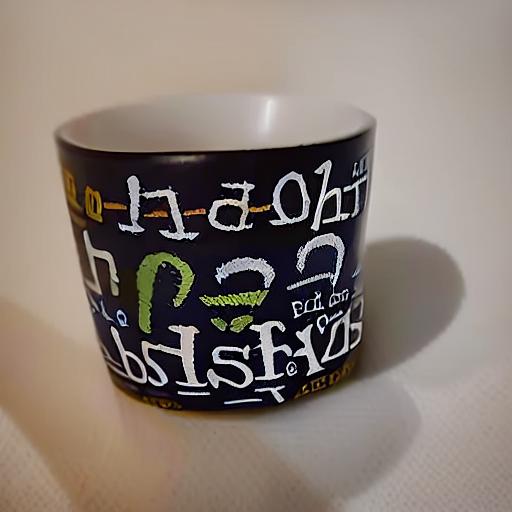} & \includegraphics[width=0.11\linewidth]{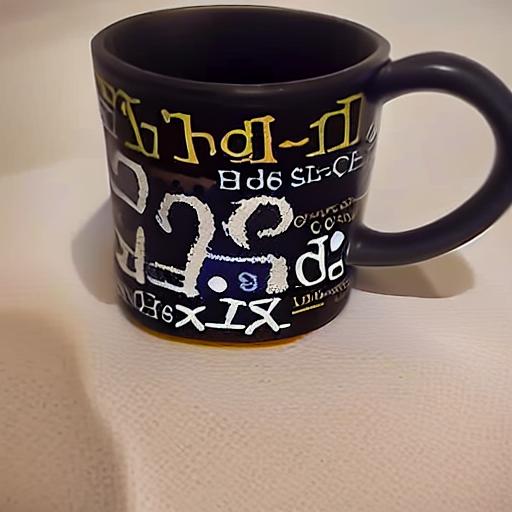} &
        \includegraphics[width=0.11\linewidth]{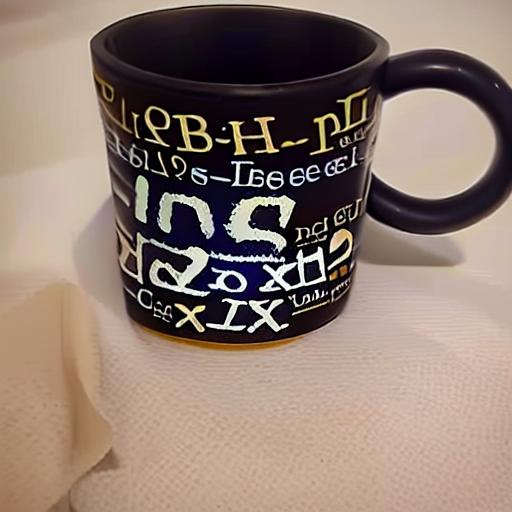} & \includegraphics[width=0.11\linewidth]{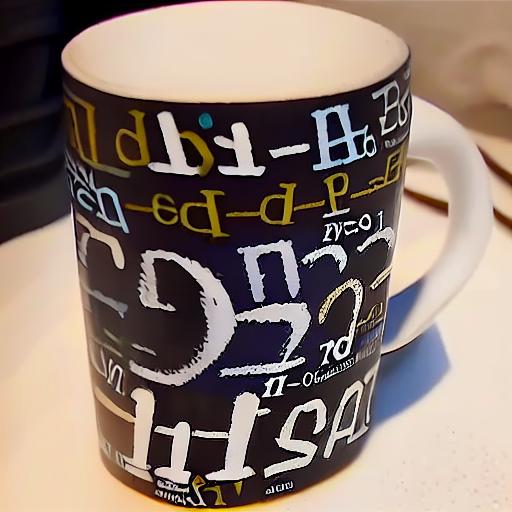} & \includegraphics[width=0.11\linewidth]{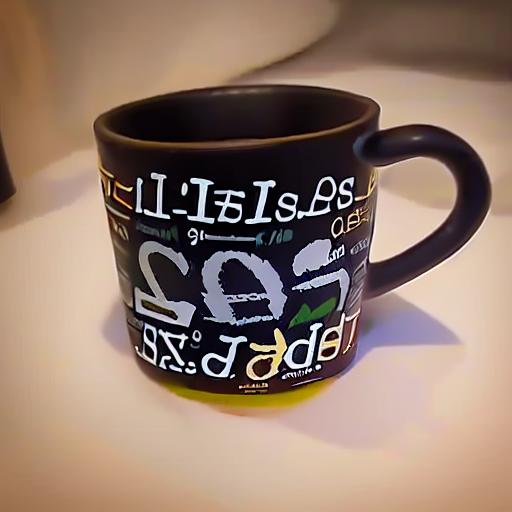} &
        \includegraphics[width=0.11\linewidth]{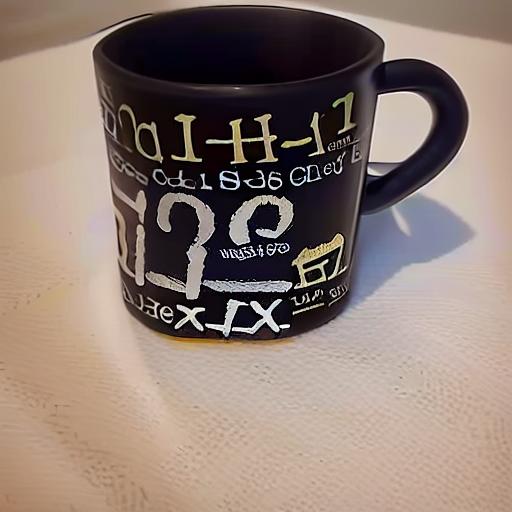} & \includegraphics[width=0.11\linewidth]{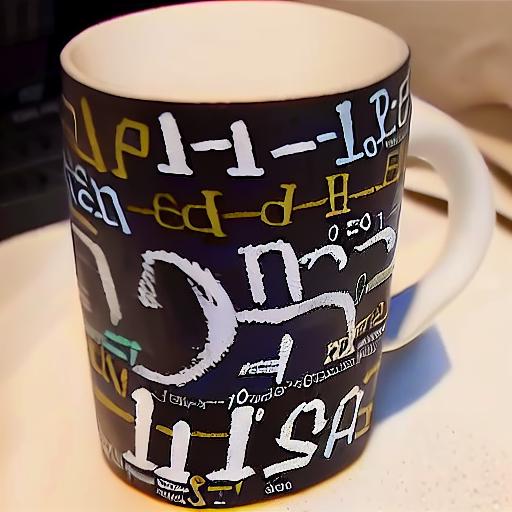} & \includegraphics[width=0.11\linewidth]{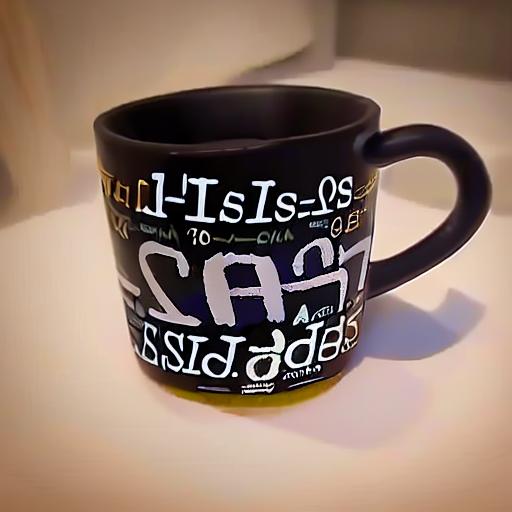}& \includegraphics[width=0.11\linewidth]{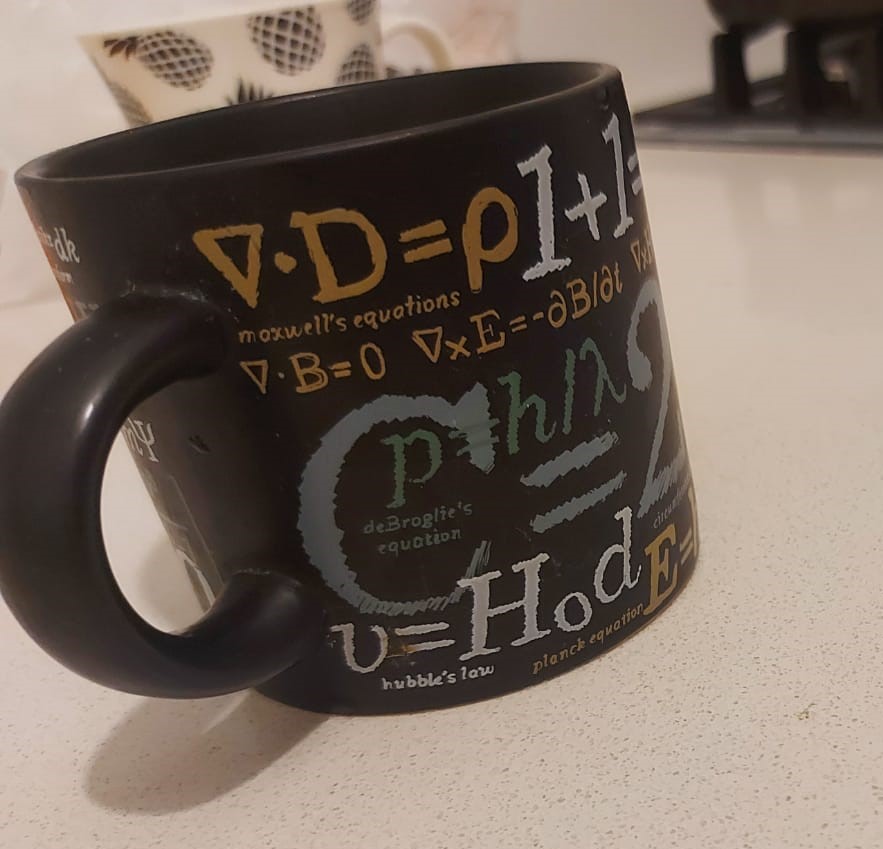} \\
        Ours (CLIP) &
        \includegraphics[width=0.11\linewidth]{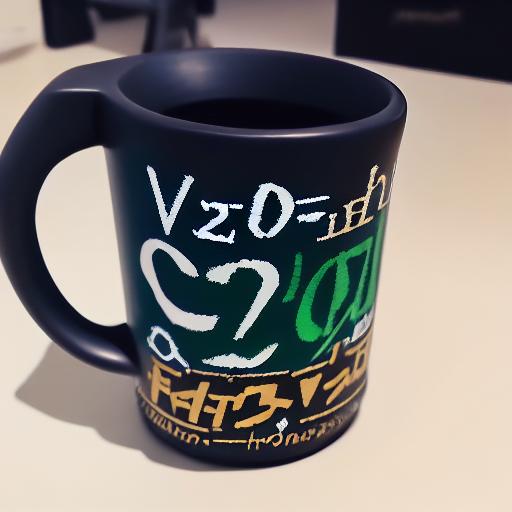} & \includegraphics[width=0.11\linewidth]{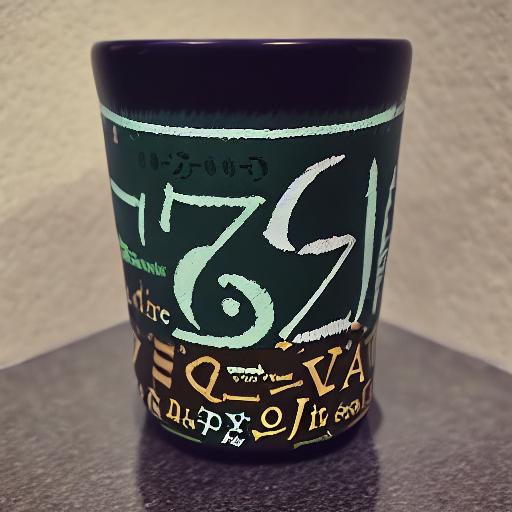} & \includegraphics[width=0.11\linewidth]{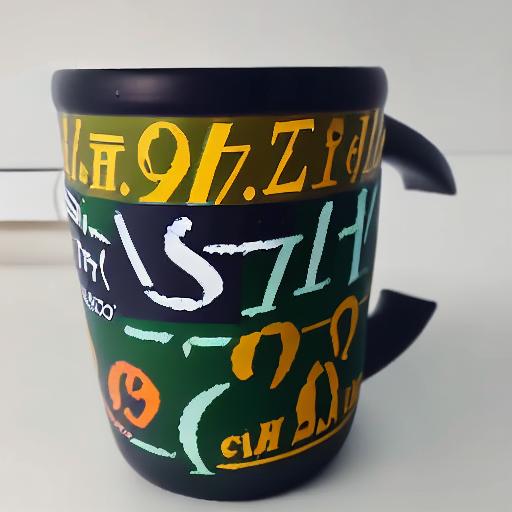} &
        \includegraphics[width=0.11\linewidth]{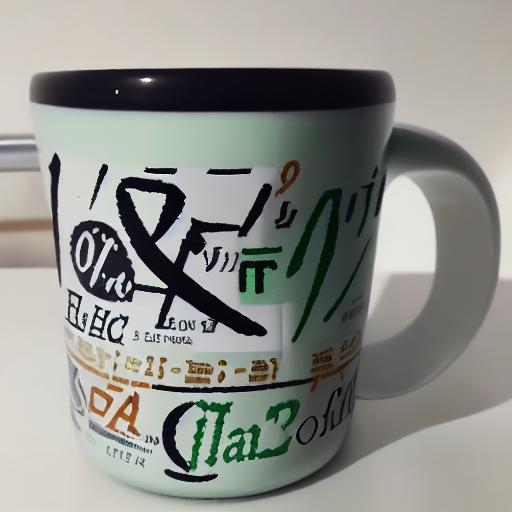} & \includegraphics[width=0.11\linewidth]{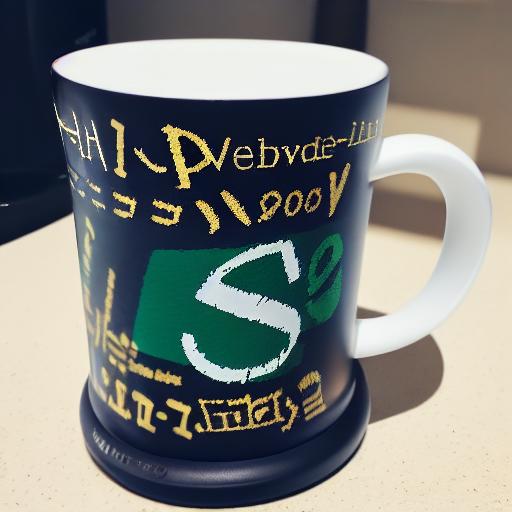} & \includegraphics[width=0.11\linewidth]{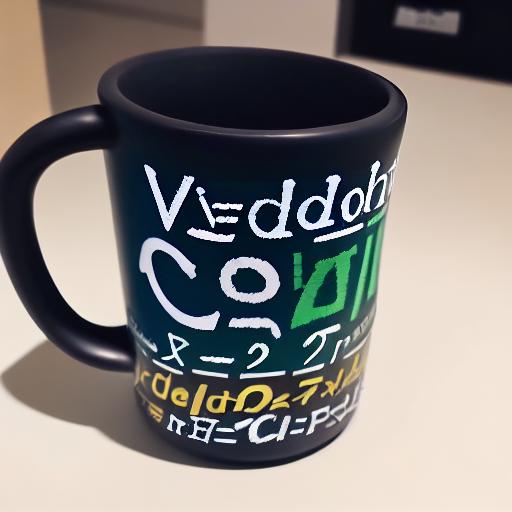} &
        \includegraphics[width=0.11\linewidth]{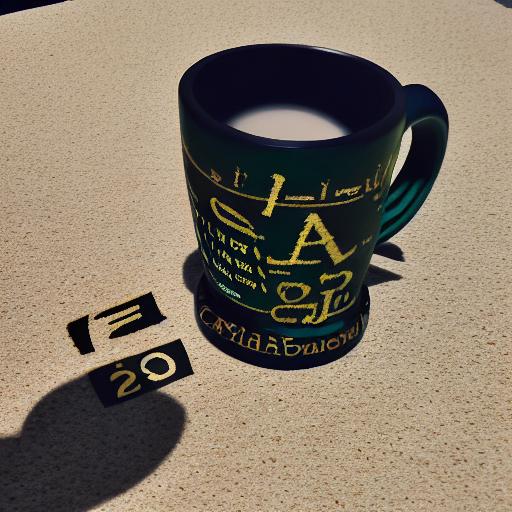} & \includegraphics[width=0.11\linewidth]{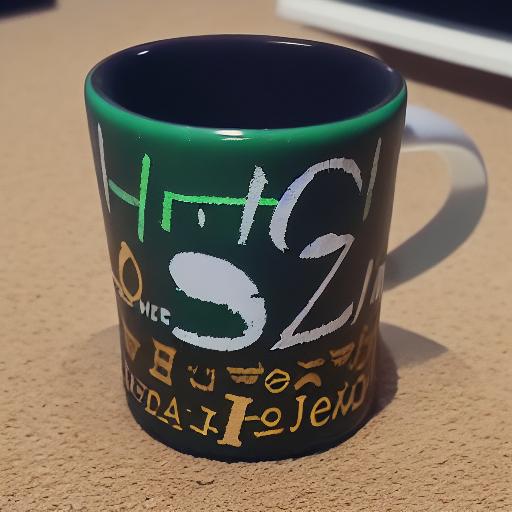} & \includegraphics[width=0.11\linewidth]{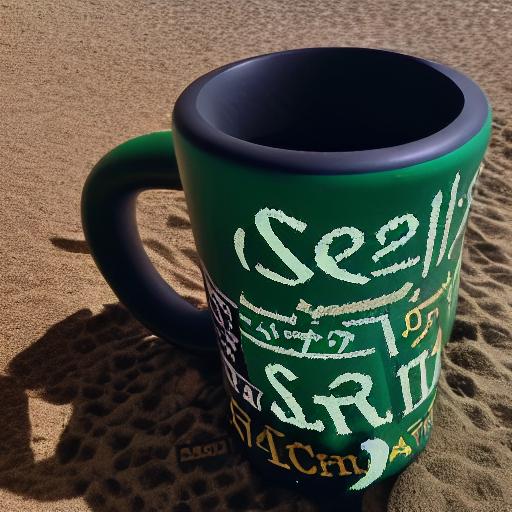}& \includegraphics[width=0.11\linewidth]{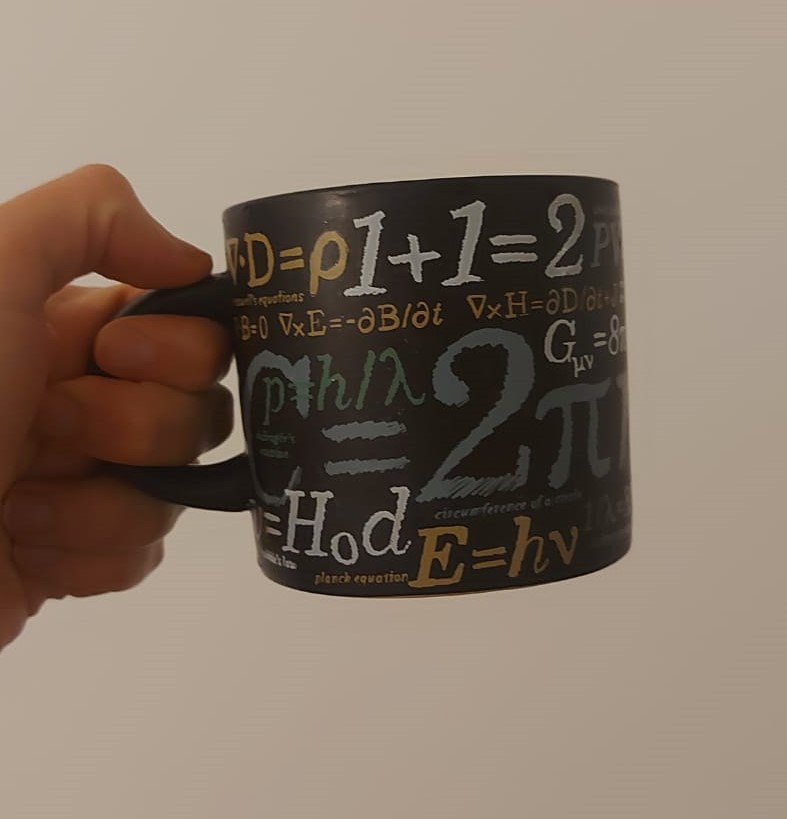} \\
        \hline
        TI (BERT) &
        \includegraphics[width=0.11\linewidth]{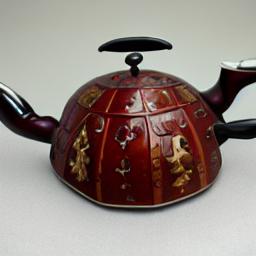} & \includegraphics[width=0.11\linewidth]{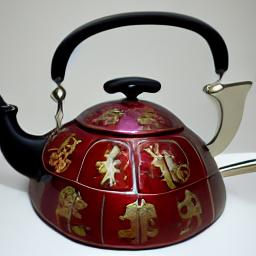} & \includegraphics[width=0.11\linewidth]{images/qual_samples/r_ld_0.jpg} &
        \includegraphics[width=0.11\linewidth]{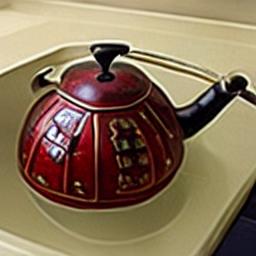} & \includegraphics[width=0.11\linewidth]{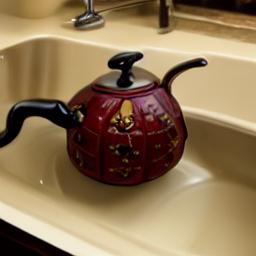} & \includegraphics[width=0.11\linewidth]{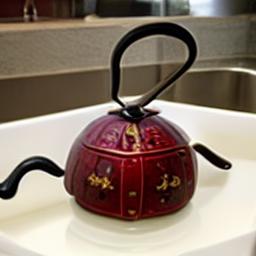} &
        \includegraphics[width=0.11\linewidth]{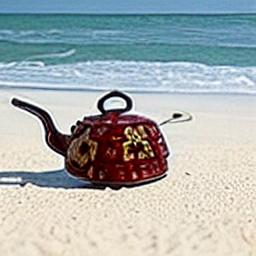} & \includegraphics[width=0.11\linewidth]{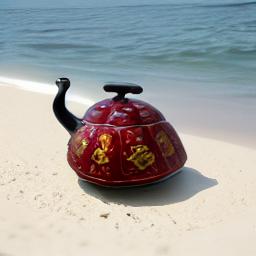} & \includegraphics[width=0.11\linewidth]{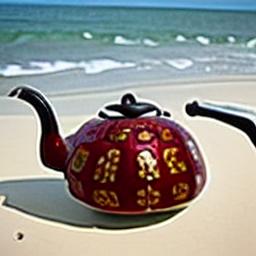} & \includegraphics[width=0.11\linewidth]{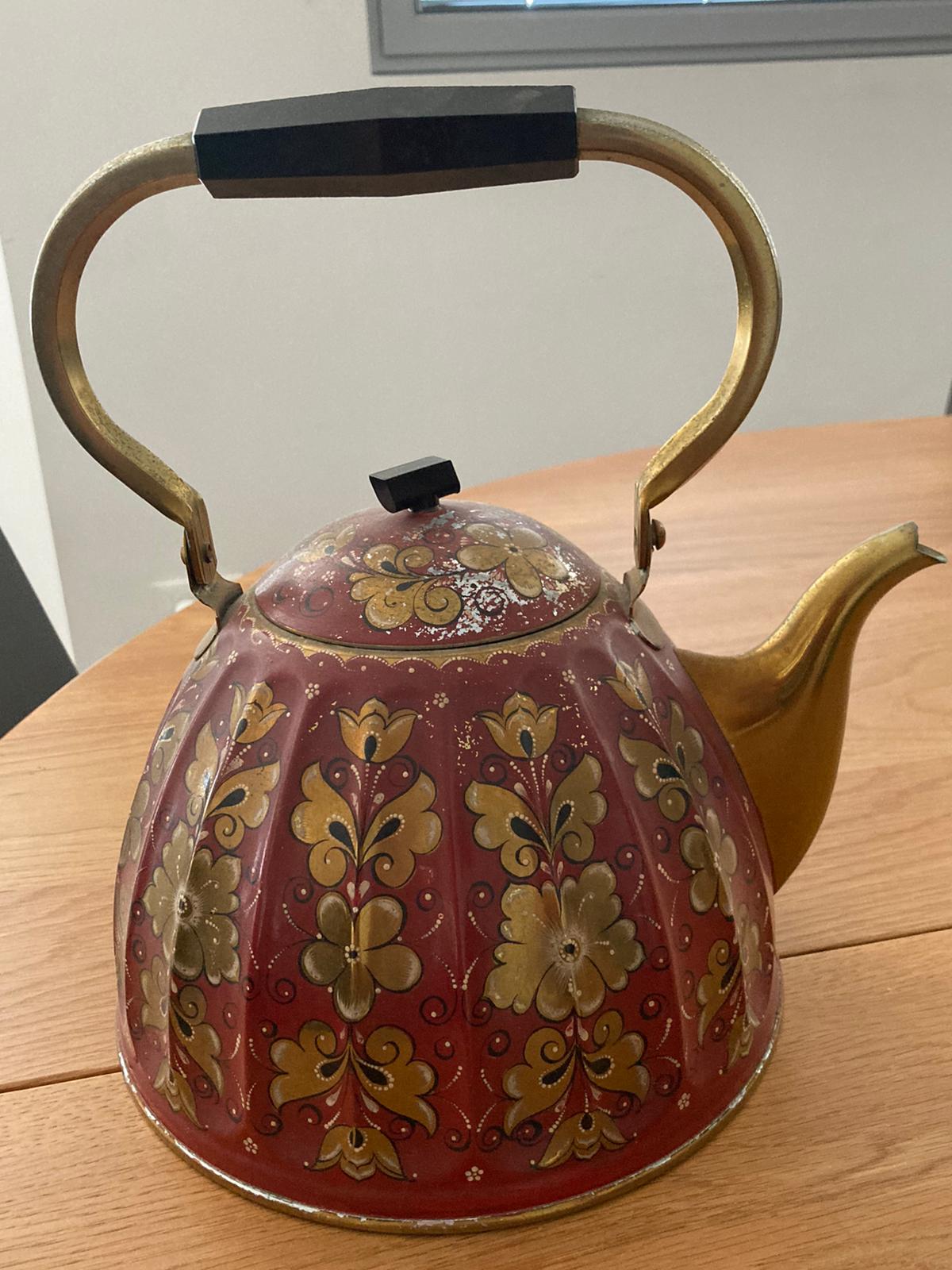}\\
        TI (CLIP) &
        \includegraphics[width=0.11\linewidth]{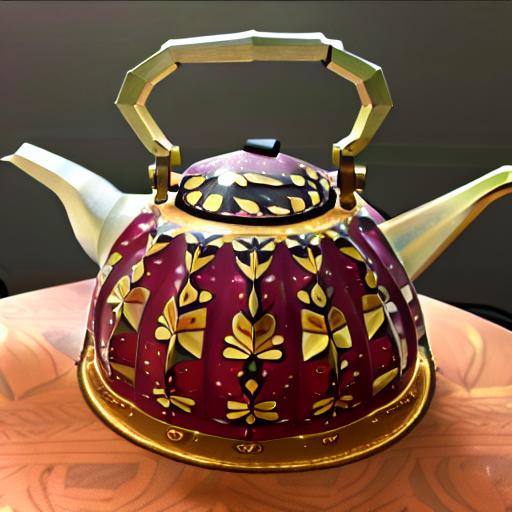} & \includegraphics[width=0.11\linewidth]{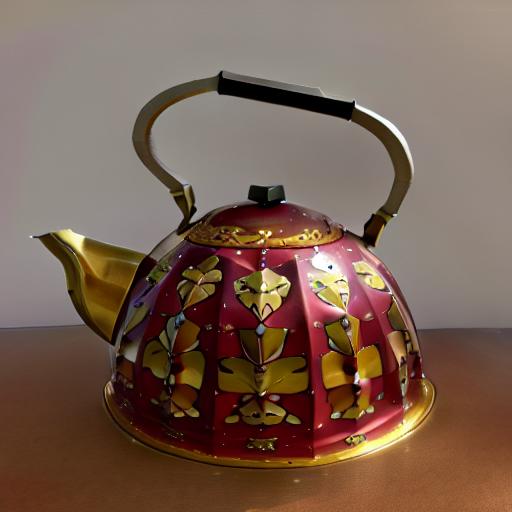} & \includegraphics[width=0.11\linewidth]{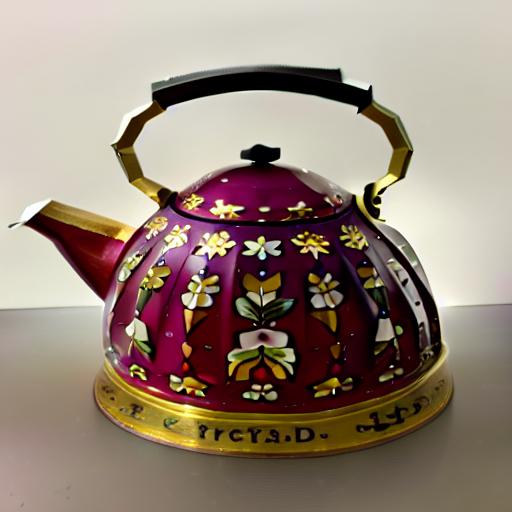} &
        \includegraphics[width=0.11\linewidth]{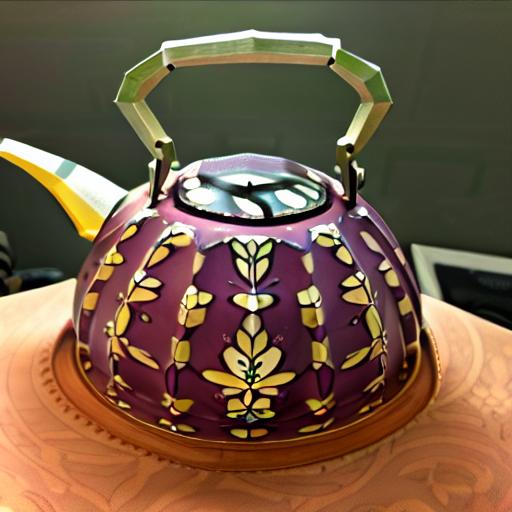} & \includegraphics[width=0.11\linewidth]{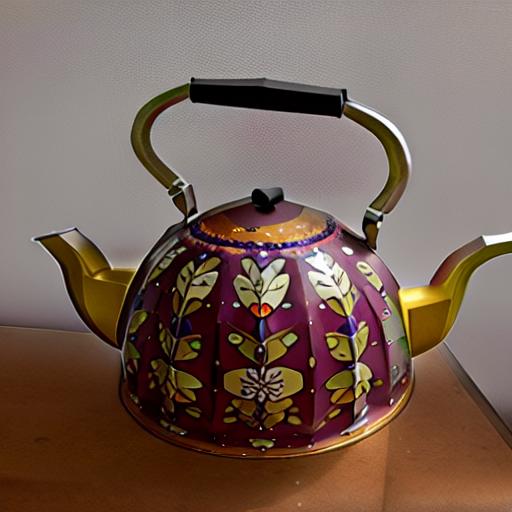} & \includegraphics[width=0.11\linewidth]{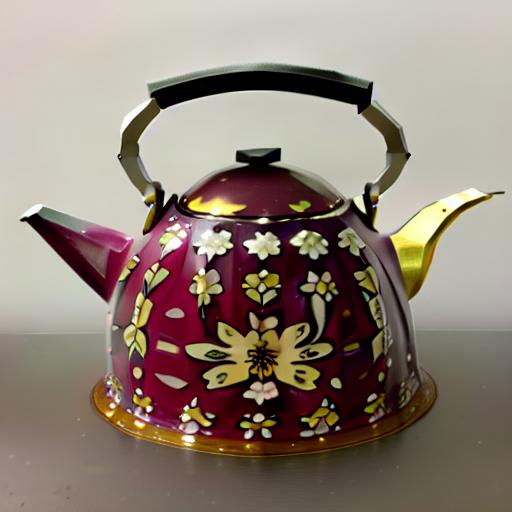} &
        \includegraphics[width=0.11\linewidth]{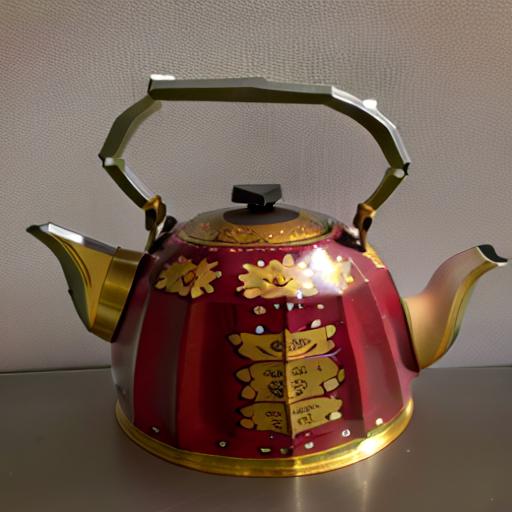} & \includegraphics[width=0.11\linewidth]{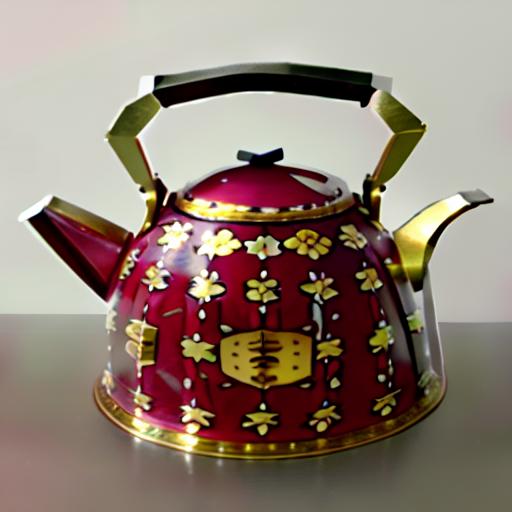} & \includegraphics[width=0.11\linewidth]{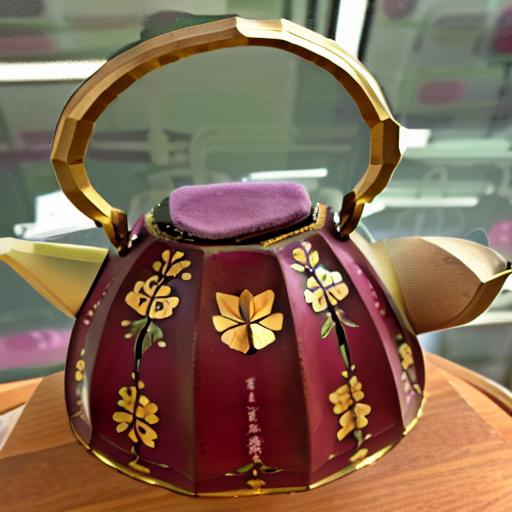}& \includegraphics[width=0.11\linewidth]{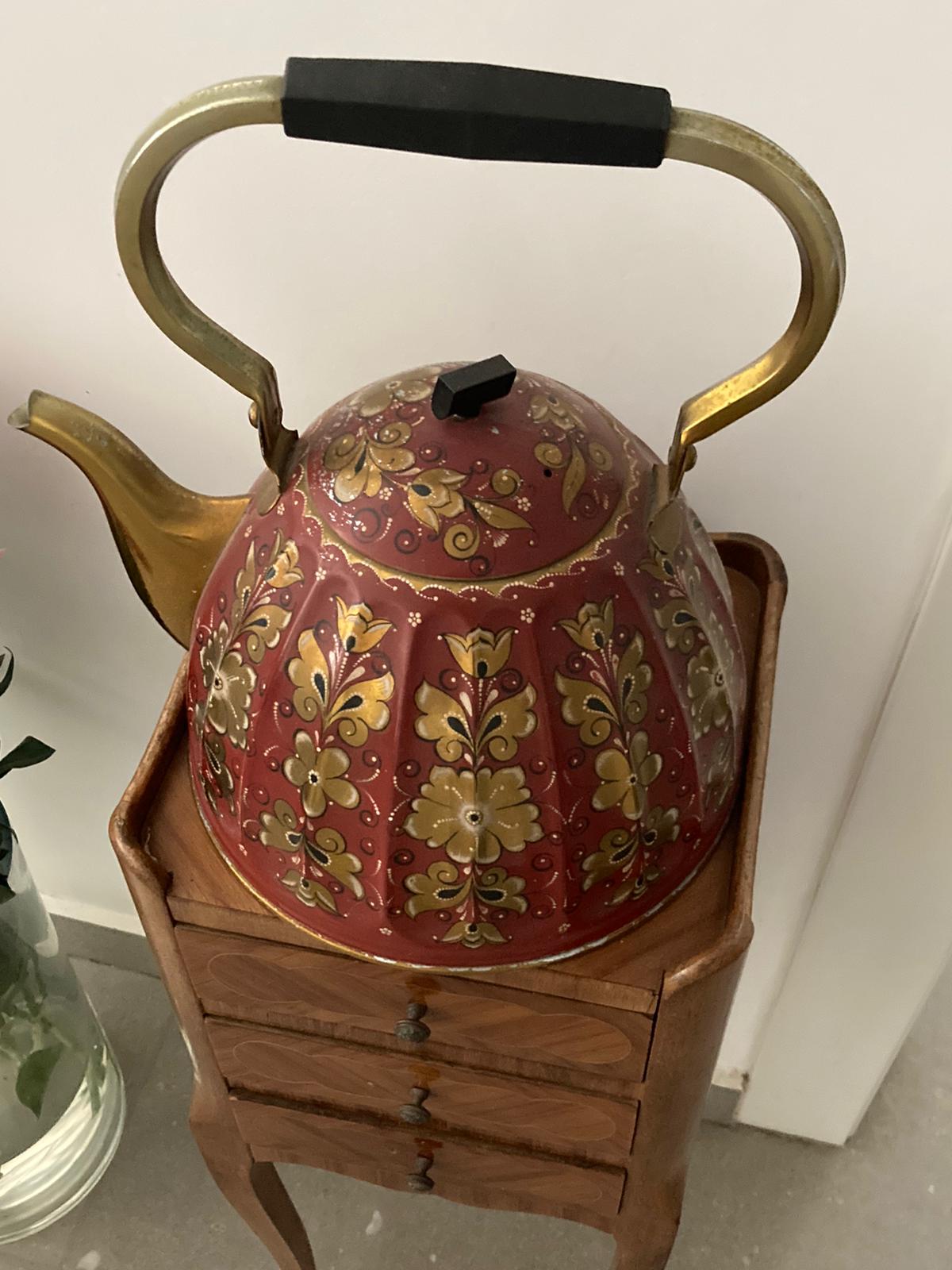} \\
        Ours (CLIP) &
        \includegraphics[width=0.11\linewidth]{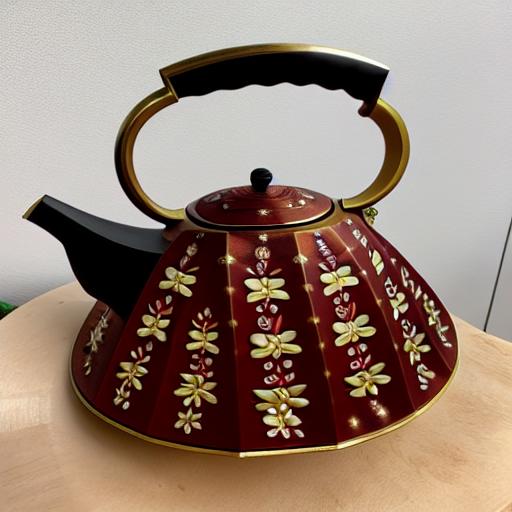} & \includegraphics[width=0.11\linewidth]{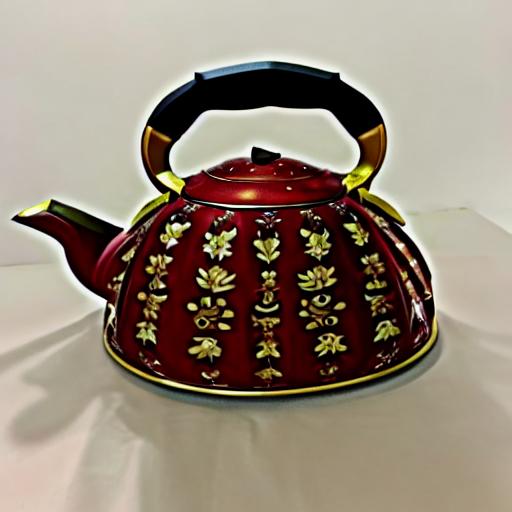} & \includegraphics[width=0.11\linewidth]{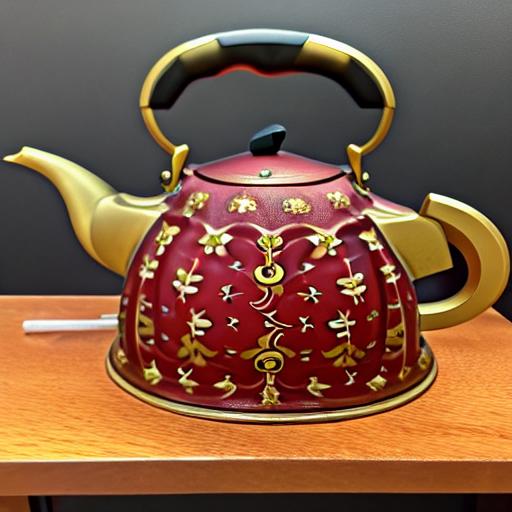} &
        \includegraphics[width=0.11\linewidth]{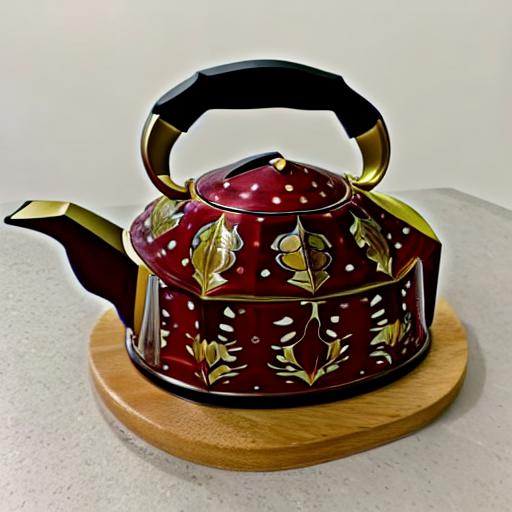} & \includegraphics[width=0.11\linewidth]{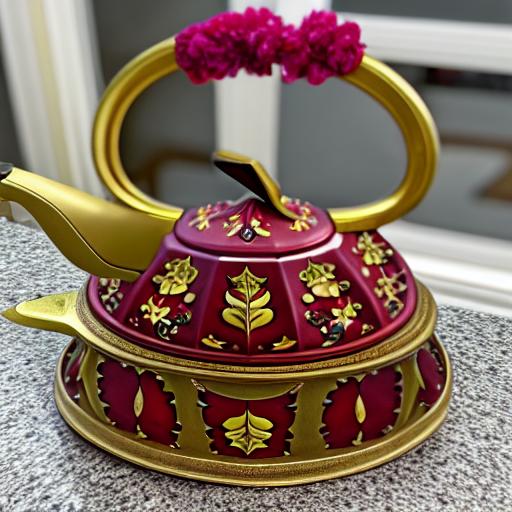} & \includegraphics[width=0.11\linewidth]{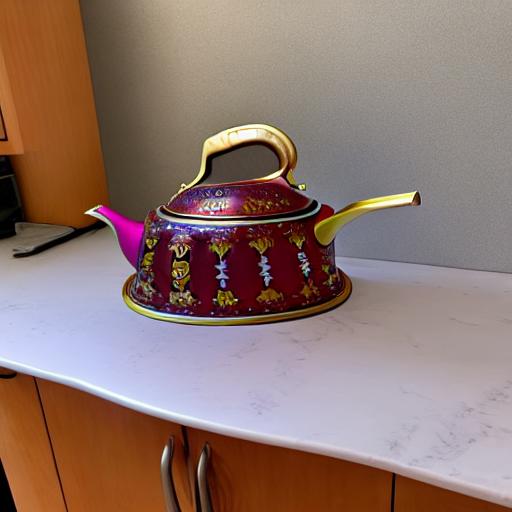} &
        \includegraphics[width=0.11\linewidth]{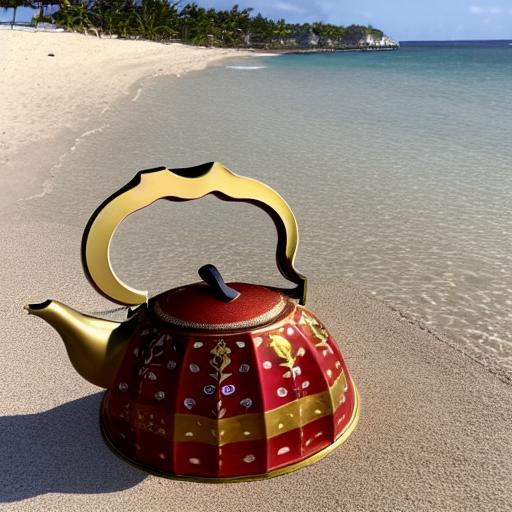} & \includegraphics[width=0.11\linewidth]{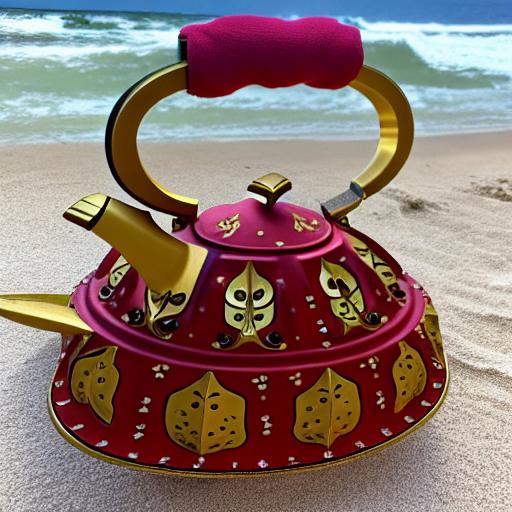} & \includegraphics[width=0.11\linewidth]{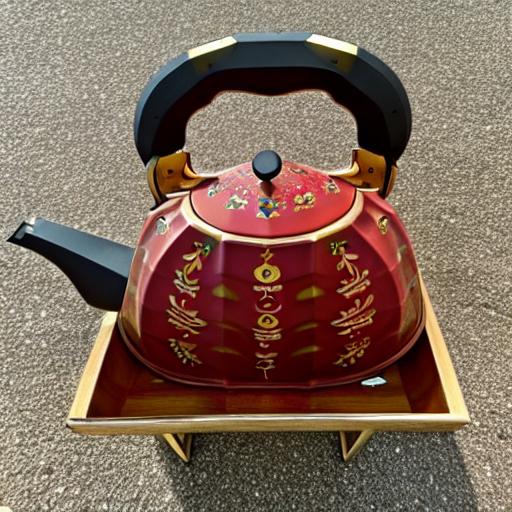}& \includegraphics[width=0.11\linewidth]{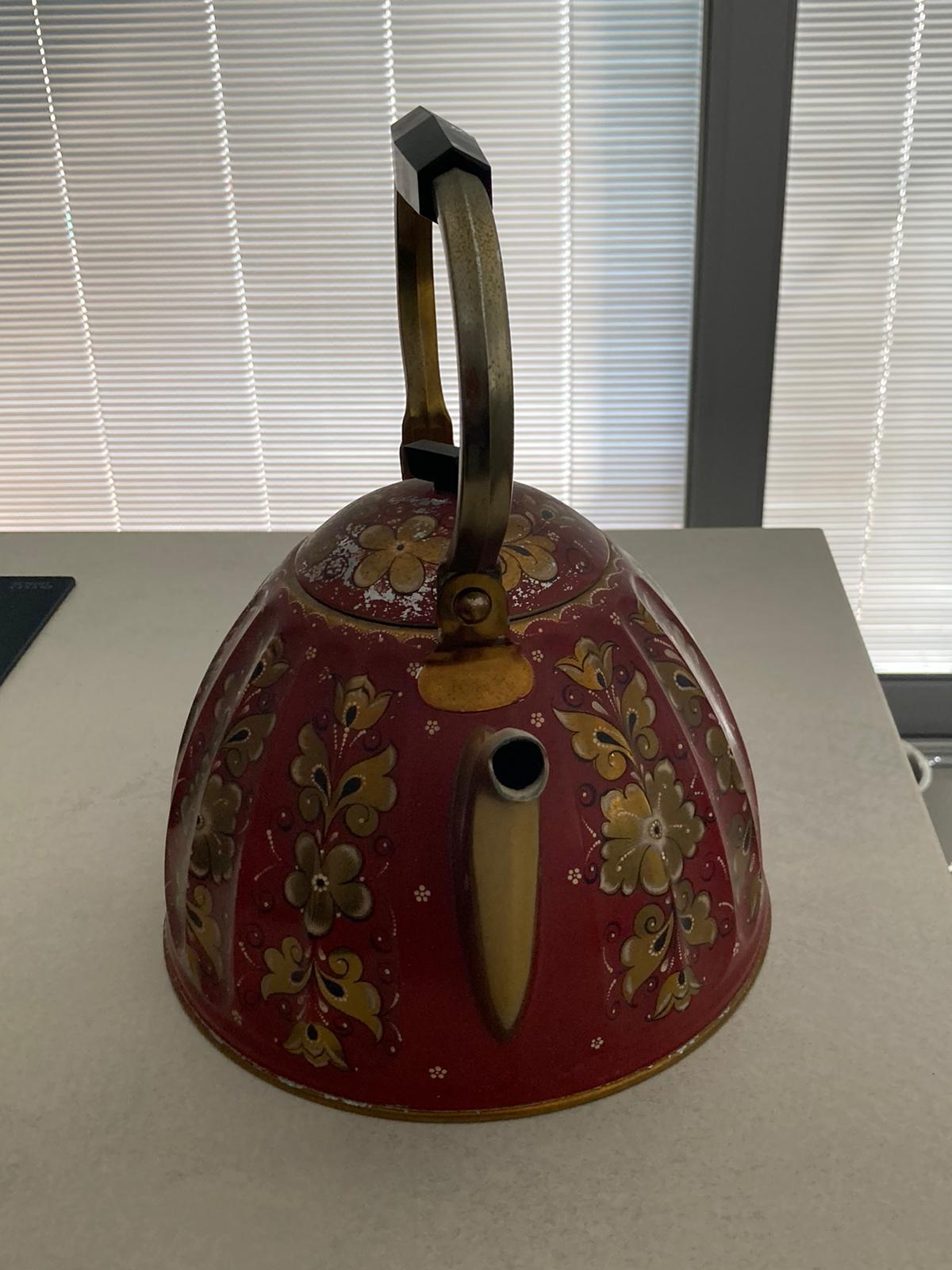} \\
      \end{tabular}} \caption{Images generated with different captions for classes from the Textual Inversion dataset.} \label{fig:results}
\end{figure*}

\noindent \textbf{Image Generation: Textual Inversion Dataset.}
In Figure~\ref{fig:ti_comp} and Figure~\ref{fig:results} we illustrate images sampled from the model for different textual prompts. These figures suggest that all three methods considered have captured the underlying concept reasonably well. This is evident as all the methods are able to produce high quality image samples when the model is prompted with `image of a $\ast$ c'. However, when textual inversion is conditioned on CLIP text features, it appears that the learned token does not compose well with natural language. This observation is consistent with Figure~\ref{fig:ti_comp}. On the other hand, when textual inversion is conditioned on BERT encoder, it better composes with natural language in half of the cases. In the remaining cases, the results tend to be very poor. For example, when the model attempts to place the colorful teapot in a sink, the model fails to generate even a trace of a teapot in the scene. In certain situations (such as the ``physics mug'' and ``red teapot'' on the beach), object fidelity is seen to be heavily affected. Even when the learned token fails to compose well with natural language, the generated image maintains details of the targeted concept. For example, in the case of the mug skull on a sink, textual inversion (BERT) does not retain the general structure of the object. On the other hand our method, while failing at the composition, preserves the structure of the object.

We quantified image generation capabilities of each method based on object/context relevance score with respect to CLIP and ALIGN features. For each concept, we considered 10 contexts  and generated 8 images per context. For context relevance, we verified whether image features of the generated image correlates most with text features of the context (out of the 10 contexts). For object relevance, we verified whether image features of the generated image correlates most with image features of the corresponding training class (compared to other classes in textual inversion dataset). In Table~\ref{tbl:clipscore}, we show accuracy for object and context classifiers when classifiers are based on CLIP and ALIGN models. On average, our proposed method leads to improved performance  of TI(CLIP) by over 15\% in both CLIP and ALIGN metrics. We note that when a better text encoder BERT is used with TI, context accuracy improves but object accuracy degrades. On average our method performs marginally better than TI(CLIP) when on CLIP metric but lags behind by nearly 6\% on ALIGN metric.

\begin{table}[]
\centering
\caption{CLIP/ALIGN-score accuracy for Textual-Inversion dataset classes considering object and context classes. Our oroposed method produces images where the concept composes better compared to the Textual Inversion baseline(CLIP).}
	\resizebox{0.7\linewidth}{!}{
\begin{tabular}{@{}lcccccc@{}}
\toprule
           & \multicolumn{3}{c}{CLIP}                                                                         & \multicolumn{3}{c}{ALIGN}                                                                        \\  \midrule
           & \multicolumn{1}{l}{Object Acc} & \multicolumn{1}{l}{Context Acc} & \multicolumn{1}{l}{Average} & \multicolumn{1}{l}{Object Acc} & \multicolumn{1}{l}{Context Acc} & \multicolumn{1}{l}{Average}  \\ \midrule
Random     & 0.166                           & 0.100                            & 0.133                       & 0.166                           & 0.100                            & 0.133                       \\
TI (BERT)  & 0.621                           & 0.743                            & 0.682                       & 0.764                           & 0.754                            & \textbf{0.759}              \\
TI(CLIP)   & 0.975                           & 0.118                            & 0.546                       & 0.983                           & 0.098                            & 0.541                       \\
Ours(CLIP) & 0.942                           & 0.456                            & \textbf{0.699}              & 0.967                           & 0.438                            & 0.702 \\ \bottomrule                      
\end{tabular}}
\label{tbl:clipscore}
\end{table}

\vspace{-.3cm}

In order to evaluate the image recognition capabilities of the learned token, we carried out benchmarking with on the Caltech256 dataset. We coupled each concept with a parent concept from Caltech256 dataset. In our evaluation we focused on three questions: 
i) Can the token distinguish the specialized concept from the parent concept?
ii) Can it distinguish specialized concept from other concepts 
iii) Can it be used to distinguish the parent concept from other concepts?
In Table~\ref{tbl:qual} we tabulate average recognition performance in terms of Area Under the Curve of ROC curve (with standard deviation in brackets). Since the prompt tuning model is trained to differentiate between target concept from its parent concept, it is not surprising that it performs well whenever the given concept is involved. However, despite the target concept being structurally similar the parent concept, the token learned by prompt tuning only obtains a AUC-ROC of 60\% when separating the parent class from other classes.  Table~\ref{tbl:qual} further suggests that textual inversion (CLIP) performs reasonably well. The proposed method, on the other hand, improves performance of textual inversion by 15\%, 19\% and 40\%. This suggests that prompt tuning overfits to local features of  objects. Since it is closely bound to the target concept, it cannot be later used to perform inference on behalf of the parent class. On  the other hand, due to the additional constraints used during training our model learns a very general representation.  Therefore, the representation of the specialized concept can use in lieu of parent concept to obtain very high performance.\\

\noindent \textbf{Text-to-Image Retrieval: DeepFashion2 Dataset.}
We trained concept tokens for each class from the train-set of the DeepFashion2 dataset from the PerVL benchmark\cite{cohenThisMyUnicorn2022a}. We used captions provided by the benchmark during training. Prior to training, we pre-processed images by manually redacting faces of all human subjects appearing in the dataset. We use this dataset to assess text-to-image retrieval performance.  We use Mean Reciprocal Rank(MRR) as the metric to compare performance between different methods. 


In Table~\ref{tbl:ret_deepfashion} we show the MRR obtained for classes in PerVL benchmark. In all methods, we compare the text feature with the image feature in the CLIP space to obtain a ordering of images. We note that, extracting text features with \pseudowords~have improved retrieval performance by 5\% compared to prompt tuning. Performance further increases by 2\% when the search query is modified using Algorithm~\ref{alg:cap}. For this dataset we considered `dress' to be the super-class name and used  ``red'', ``old'', ``worn'', ``bright'', ``dark'', ``pink'' as the list of attributes $A$. Here, we note that using Algorithm~\ref{alg:cap} with prompt tuning has deteriorated its original performance. This deterioration is expected since prompt tuning does not produce semantically meaningfully images when visualized. When this is the case, the algorithm is unable to locate the optimum weight by considering the CLIP similarity score.

\begin{figure}[ht] 
\centering
\resizebox{1\linewidth}{!}{ 
   
    \centering{{\includegraphics[width=1\linewidth] {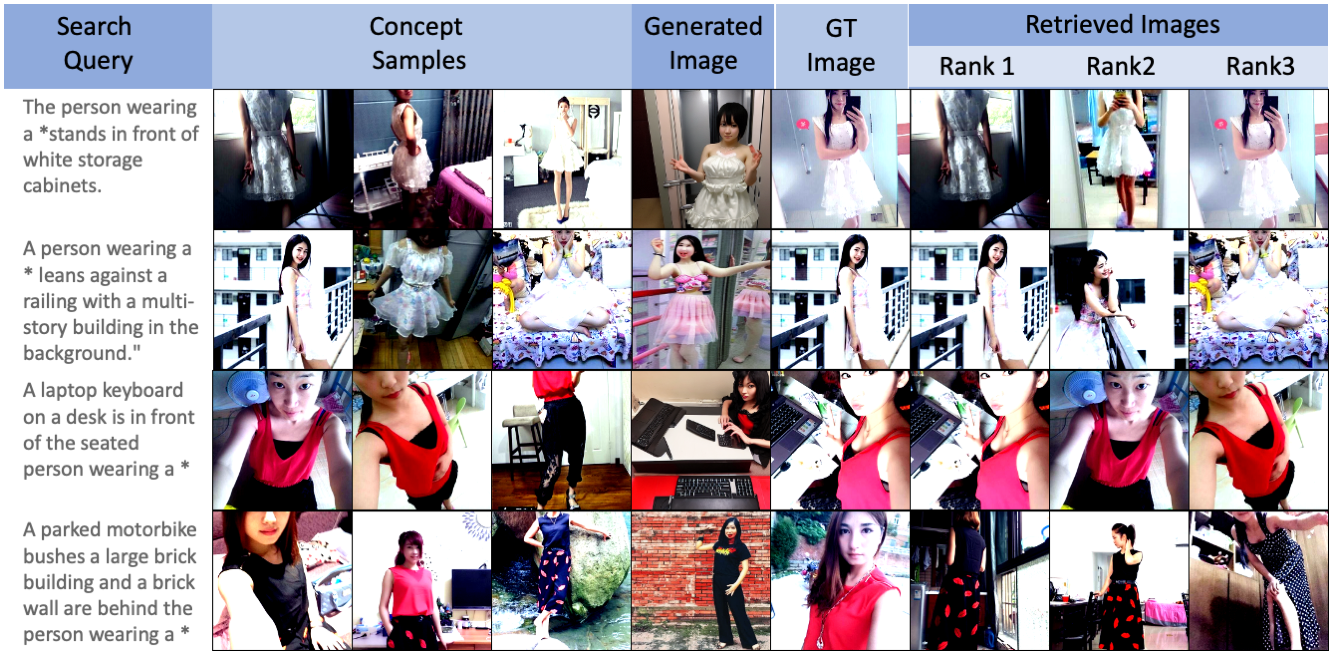} }} 
     }
    \caption{Retrieval results for DeepFashion2 dataset. For each query (row), few samples of the concept is visualized. We show a sample image generated from the query along with the GT image and top-3 ranked image retrievals. In top 3 rows, the generated image is in-line with the intended query. The GT image is amongst the top-3 image retrievals. Last row is an example of a failed image retrieval. Once visualized, the generated image shows that the search is different from the intended criteria.  }%

    \label{fig:qualDeepFashion}
\end{figure}

In Figure~\ref{fig:qualDeepFashion}, we visualize few examples of retrieval from the DeepFashion2 dataset. In Figure~\ref{fig:qualDeepFashion}, few samples of the underlying concept is visualized along the query. For each case, we have illustrated a image sample generated from the model. In the top three rows, the model has obtained a top-3 retrieval. Here we note that the generated image is inline with what is described by the text query. In the bottom row, we illustrate a failure case - where GT image is not obtained as a part of the top-3 retrieval. In this case, we show that the generated image is very different from the description of the query. It appears the model has paid more attention to the phrase `brick wall' in the query. This is an example of how our model can be used to interpret results.

\vspace{-.2cm}

\begin{table}[t]
\caption{Aggregated recognition performance measured by AUC-ROC for five classes from the Textual-Inversion dataset with standard deviation. }

\centering
	\resizebox{0.8\linewidth}{!}{
\begin{tabular}{lccc}

\toprule
                                                      & Prompt Tuning (Paragon)
                                                      & Textual Inversion
                                                      & Ours                   \\ 
\hline
Target concept vs parent concept        & \textbf{100.0 (0.0)} & 58.89 (42.12)     & 98.88 (2.21)           \\ 
Target concept vs other Caltech classes & \textbf{100.0 (0.0)} & 80.53 (30.44)     & \textbf{100.0 (0.0)}   \\ 
Parent concept vs other Caltech classes & 58.93 (25.24)        & 70.54 (15.96)     & \textbf{85.49 (11.42)} \\ 
\bottomrule
\end{tabular}} 
\label{tbl:qual}
\vspace{-.2cm}
\end{table}


\begin{table}[h]
\centering
\caption{MRR for retrieval on the DeepFashion2 Dataset (50 classes).}
\label{tab:single_col}
\resizebox{.4\linewidth}{!}{
\begin{tabular}{l c}
\toprule
Method & MRR \\
\midrule
Text Only & 17.6 (0.0) \\
Text+Avg Image & 21.7 (2.4) \\
PALAVRA \cite{cohenThisMyUnicorn2022a} & 28.4 (0.7) \\
Ideal Words \cite{Trager2023} & 37.0 (1.1) \\
Prompt Tuning & 51.4 (2.1) \\
Prompt Tuning + GAIR & 45.2 (0.8) \\
\textbf{Ours} & \textbf{56.6 (2.3)} \\
\textbf{Ours + GAIR} & \textbf{58.8 (2.3)} \\
\bottomrule
\end{tabular}}
\label{tbl:ret_deepfashion}
\end{table}

\vspace{-.5cm}

\vspace{-0.5cm}
\section{Conclusion}
We investigated the problem of learning flexible custom tokens that can be used for both generative and discriminative visual tasks. The proposed method generates custom tokens 
that can be used with natural language in search queries and lead to strong text-to-image retrieval performance, particularly when coupled with the Generation Augmented Image Retrieval algorithm. We have observed how our modifications and their parameters contribute to the tradeoff between visual fidelity and compositionality. In the future, we hope to investigate solutions that can find a good operating point that balances visual fidelity and compositionality.

{
    \small
    \bibliographystyle{splncs04}
    \bibliography{eccv}

\begin{thebibliography}{10}
\providecommand{\url}[1]{\texttt{#1}}
\providecommand{\urlprefix}{URL }
\providecommand{\doi}[1]{https://doi.org/#1}

\bibitem{abdalImage2StyleGANHowEmbed2019}
Abdal, R., Qin, Y., Wonka, P.: {Image2StyleGAN}: {How} to {Embed} {Images}
  {Into} the {StyleGAN} {Latent} {Space}? (Sep 2019),
  \url{http://arxiv.org/abs/1904.03189}, arXiv:1904.03189 [cs]

\bibitem{cohenThisMyUnicorn2022a}
Cohen, N., Gal, R., Meirom, E.A., Chechik, G., Atzmon, Y.: “{This} {Is} {My}
  {Unicorn}, {Fluffy}”: {Personalizing} {Frozen} {Vision}-{Language}
  {Representations}. In: Avidan, S., Brostow, G., Cissé, M., Farinella, G.M.,
  Hassner, T. (eds.) Computer {Vision} – {ECCV} 2022, vol. 13680, pp.
  558--577. Springer Nature Switzerland, Cham (2022).
  \doi{10.1007/978-3-031-20044-1_32},
  \url{https://link.springer.com/10.1007/978-3-031-20044-1_32}, series Title:
  Lecture Notes in Computer Science

\bibitem{creswellInvertingGeneratorGenerative2016}
Creswell, A., Bharath, A.A.: Inverting {The} {Generator} {Of} {A} {Generative}
  {Adversarial} {Network} (Nov 2016), \url{http://arxiv.org/abs/1611.05644},
  arXiv:1611.05644 [cs]

\bibitem{dongPromptTuningInversion2023}
Dong, W., Xue, S., Duan, X., Han, S.: Prompt {Tuning} {Inversion} for
  {Text}-{Driven} {Image} {Editing} {Using} {Diffusion} {Models} (May 2023),
  \url{http://arxiv.org/abs/2305.04441}, arXiv:2305.04441 [cs]

\bibitem{textinversion}
Gal, R., Alaluf, Y., Atzmon, Y., Patashnik, O., Bermano, A.H., Chechik, G.,
  Cohen-or, D.: An image is worth one word: Personalizing text-to-image
  generation using textual inversion. In: The Eleventh International Conference
  on Learning Representations (2023),
  \url{https://openreview.net/forum?id=NAQvF08TcyG}

\bibitem{ge2019deepfashion2}
Ge, Y., Zhang, R., Wu, L., Wang, X., Tang, X., Luo, P.: Deepfashion2: A
  versatile benchmark for detection, pose estimation, segmentation and
  re-identification of clothing images  (2019),
  \url{http://arxiv.org/abs/1901.07973}, cite arxiv:1901.07973

\bibitem{ddpm}
Ho, J., Jain, A., Abbeel, P.: Denoising diffusion probabilistic models. In:
  Larochelle, H., Ranzato, M., Hadsell, R., Balcan, M., Lin, H. (eds.) Advances
  in Neural Information Processing Systems. vol.~33, pp. 6840--6851. Curran
  Associates, Inc. (2020),
  \url{https://proceedings.neurips.cc/paper_files/paper/2020/file/4c5bcfec8584af0d967f1ab10179ca4b-Paper.pdf}

\bibitem{jiaVisualPromptTuning2022}
Jia, M., Tang, L., Chen, B.C., Cardie, C., Belongie, S., Hariharan, B., Lim,
  S.N.: Visual {Prompt} {Tuning} (Jul 2022),
  \url{http://arxiv.org/abs/2203.12119}, arXiv:2203.12119 [cs]

\bibitem{lester-etal-2021-power}
Lester, B., Al-Rfou, R., Constant, N.: The power of scale for
  parameter-efficient prompt tuning. In: Proceedings of the 2021 Conference on
  Empirical Methods in Natural Language Processing. pp. 3045--3059. Association
  for Computational Linguistics, Online and Punta Cana, Dominican Republic (Nov
  2021). \doi{10.18653/v1/2021.emnlp-main.243},
  \url{https://aclanthology.org/2021.emnlp-main.243}

\bibitem{liptonPreciseRecoveryLatent2017}
Lipton, Z.C., Tripathi, S.: Precise {Recovery} of {Latent} {Vectors} from
  {Generative} {Adversarial} {Networks} (Feb 2017),
  \url{http://arxiv.org/abs/1702.04782}, arXiv:1702.04782 [cs, stat]

\bibitem{liuPretrainPromptPredict2021}
Liu, P., Yuan, W., Fu, J., Jiang, Z., Hayashi, H., Neubig, G.: Pre-train,
  {Prompt}, and {Predict}: {A} {Systematic} {Survey} of {Prompting} {Methods}
  in {Natural} {Language} {Processing}. arXiv:2107.13586 [cs]  (Jul 2021),
  \url{http://arxiv.org/abs/2107.13586}, arXiv: 2107.13586

\bibitem{mokadyNulltextInversionEditing2022}
Mokady, R., Hertz, A., Aberman, K., Pritch, Y., Cohen-Or, D.: Null-text
  {Inversion} for {Editing} {Real} {Images} using {Guided} {Diffusion} {Models}
  (Nov 2022), \url{http://arxiv.org/abs/2211.09794}, arXiv:2211.09794 [cs]

\bibitem{Naeem_2021_CVPR}
Naeem, M.F., Xian, Y., Tombari, F., Akata, Z.: Learning graph embeddings for
  compositional zero-shot learning. In: Proceedings of the IEEE/CVF Conference
  on Computer Vision and Pattern Recognition (CVPR). pp. 953--962 (June 2021)

\bibitem{nayakLearningComposeSoft2022}
Nayak, N.V., Yu, P., Bach, S.H.: Learning to {Compose} {Soft} {Prompts} for
  {Compositional} {Zero}-{Shot} {Learning} (Apr 2022),
  \url{http://arxiv.org/abs/2204.03574}, number: arXiv:2204.03574
  arXiv:2204.03574 [cs]

\bibitem{radfordLearningTransferableVisual2021}
Radford, A., Kim, J.W., Hallacy, C., Ramesh, A., Goh, G., Agarwal, S., Sastry,
  G., Askell, A., Mishkin, P., Clark, J., Krueger, G., Sutskever, I.: Learning
  {Transferable} {Visual} {Models} {From} {Natural} {Language} {Supervision}.
  arXiv:2103.00020 [cs]  (Feb 2021), \url{http://arxiv.org/abs/2103.00020},
  arXiv: 2103.00020

\bibitem{rameshHierarchicalTextConditionalImage2022}
Ramesh, A., Dhariwal, P., Nichol, A., Chu, C., Chen, M.: Hierarchical
  {Text}-{Conditional} {Image} {Generation} with {CLIP} {Latents}.
  arXiv:2204.06125 [cs]  (Apr 2022), \url{http://arxiv.org/abs/2204.06125},
  arXiv: 2204.06125

\bibitem{stable_diffusion_Rombach_21}
Rombach, R., Blattmann, A., Lorenz, D., Esser, P., Ommer, B.: High-resolution
  image synthesis with latent diffusion models. CoRR  \textbf{abs/2112.10752}
  (2021), \url{https://arxiv.org/abs/2112.10752}

\bibitem{ruizDreamBoothFineTuning2022}
Ruiz, N., Li, Y., Jampani, V., Pritch, Y., Rubinstein, M., Aberman, K.:
  {DreamBooth}: {Fine} {Tuning} {Text}-to-{Image} {Diffusion} {Models} for
  {Subject}-{Driven} {Generation} (Aug 2022),
  \url{http://arxiv.org/abs/2208.12242}, arXiv:2208.12242 [cs]

\bibitem{sahariaPhotorealisticTexttoImageDiffusion2022}
Saharia, C., Chan, W., Saxena, S., Li, L., Whang, J., Denton, E., Ghasemipour,
  S.K.S., Ayan, B.K., Mahdavi, S.S., Lopes, R.G., Salimans, T., Ho, J., Fleet,
  D.J., Norouzi, M.: Photorealistic {Text}-to-{Image} {Diffusion} {Models} with
  {Deep} {Language} {Understanding} (May 2022),
  \url{http://arxiv.org/abs/2205.11487}, number: arXiv:2205.11487
  arXiv:2205.11487 [cs]

\bibitem{saitoPic2WordMappingPictures2023}
Saito, K., Sohn, K., Zhang, X., Li, C.L., Lee, C.Y., Saenko, K., Pfister, T.:
  {Pic2Word}: {Mapping} {Pictures} to {Words} for {Zero}-shot {Composed}
  {Image} {Retrieval} (Feb 2023), \url{http://arxiv.org/abs/2302.03084},
  arXiv:2302.03084 [cs]

\bibitem{ddmps-sohl-dickstein15}
Sohl-Dickstein, J., Weiss, E., Maheswaranathan, N., Ganguli, S.: Deep
  unsupervised learning using nonequilibrium thermodynamics. In: Bach, F.,
  Blei, D. (eds.) Proceedings of the 32nd International Conference on Machine
  Learning. Proceedings of Machine Learning Research, vol.~37, pp. 2256--2265.
  PMLR, Lille, France (07--09 Jul 2015),
  \url{https://proceedings.mlr.press/v37/sohl-dickstein15.html}

\bibitem{Trager2023}
Trager, M., Perera, P., Zancato, L., Achille, A., Bhatia, P., Soatto, S.:
  Linear spaces of meanings: Compositional structures in vision-language
  models. In: ICCV 2023 (2023),
  \url{https://www.amazon.science/publications/linear-spaces-of-meanings-compositional-structures-in-vision-language-models}

\bibitem{wang2022learning}
Wang, Z., Zhang, Z., Lee, C.Y., Zhang, H., Sun, R., Ren, X., Su, G., Perot, V.,
  Dy, J., Pfister, T.: Learning to prompt for continual learning. In:
  Proceedings of the IEEE/CVF Conference on Computer Vision and Pattern
  Recognition. pp. 139--149 (2022)

\bibitem{zhou2022coop}
Zhou, K., Yang, J., Loy, C.C., Liu, Z.: Learning to prompt for vision-language
  models. International Journal of Computer Vision (IJCV)  (2022)

\end{thebibliography}
}

\end{document}